\documentclass[nonblindrev,pdflatex]{informs3arxiv}

\OneAndAHalfSpacedXI
\usepackage{blkarray}
\usepackage[thinc]{esdiff}
\usepackage{bm}
\usepackage{bbm}
\usepackage{graphicx}
\usepackage{url}
\usepackage{subfig}
\usepackage{subfloat}
\usepackage{xcolor}
\usepackage{booktabs}
\usepackage[ruled]{algorithm2e} % For algorithms
\usepackage{pgfplots}
\usepackage{tikz}
\usepackage{mathtools}
\usetikzlibrary{pgfplots.groupplots}
\usepackage{thmtools}
\usepackage{thm-restate}

\pgfplotsset{soldot/.style={color=blue,only marks,mark=*}} 
\pgfplotsset{holdot/.style={color=blue,fill=white,only marks,mark=*}}

\usepgfplotslibrary{external}
\usepackage{natbib}
 \bibpunct[, ]{(}{)}{,}{a}{}{,}%
 \def\bibfont{\small}%

\newcommand{\blu}{\color{black}}
\newcommand{\bla}{\color{black}}

\TheoremsNumberedThrough     % Preferred (Theorem 1, Lemma 1, Theorem 2)
\ECRepeatTheorems

\EquationsNumberedThrough    % Default: (1), (2), ...
\MANUSCRIPTNO{MS-0001-1922.65}

\begin{document}
%%%%%%%%%%%%%%%%

%\setcounter{page}{1}
%\DOI{10.1287/xxxx.0000.0000}%


% \RUNTITLE{Loss Functions for Contextual Pricing with Observational Data: Continuous Prices and Convexity}
\RUNTITLE{Convex Surrogate Loss Functions for Contextual Pricing with Transaction Data}

% \TITLE{Bundling Information Goods of Decreasing Value}
% \TITLE{Loss Functions for Contextual Pricing with Transaction Data: Continuous Prices and Convexity}
\TITLE{Convex Surrogate Loss Functions for Contextual Pricing with Transaction Data}

\ARTICLEAUTHORS{%
\AUTHOR{Max Biggs}
\AFF{Darden School of Business, University of Virginia, Charlottesville, VA 22902, \EMAIL{biggsm@darden.virginia.edu}} %,
} % end of the block

\ABSTRACT{
We study an off-policy contextual pricing problem where the seller has access to samples of prices that customers were previously offered, whether they purchased at that price, and auxiliary features describing the customer and/or item being sold. This is in contrast to the well-studied setting in which samples of the customer's valuation (willingness to pay) are observed. In our setting, the observed data is influenced by the previous pricing policy, and we do not know how customers would have responded to alternative prices. We introduce suitable loss functions for this setting that can be directly optimized to find an effective pricing policy with expected revenue guarantees, without the need for estimation of an intermediate demand function. %We propose \textit{convex} loss functions for this setting which provide \textit{expected revenue guarantees} when optimized. 
We focus on \textit{convex} loss functions. This is particularly relevant when linear pricing policies are desired for interpretability reasons, resulting in a tractable convex revenue optimization problem. We propose generalized \textit{hinge} and \textit{quantile} pricing loss functions that price at a multiplicative factor of the conditional expected valuation or a particular quantile of the prices which sold, despite the valuation data not being observed. We prove expected revenue bounds for these pricing policies respectively when the valuation distribution is log-concave, and we provide generalization bounds for the finite sample case. Finally, we conduct simulations on both synthetic and real-world data to demonstrate that this approach is competitive with, and in some settings outperforms, state-of-the-art methods in contextual pricing.
}%

%\KEYWORDS{deterministic inventory theory; infinite linear programming duality;
%  existence of optimal policies; semi-Markov decision process; cyclic schedule}


\maketitle

%%%%%%%%%%%%%%%%%%%%%%%%%%%%%%%%%%%%%%%%%%%%%%%%%%%%%%%%%%%%%%%%%%%%%%

% Samples of sectioning (and labeling) in MNSC
% NOTE: (1) \section and \subsection do NOT end with a period
%       (2) \subsubsection and lower need end punctuation
%       (3) capitalization is as shown (title style).
%
%\section{Introduction.}\label{intro} %%1.
%\subsection{Duality and the Classical EOQ Problem.}\label{class-EOQ} %% 1.1.
%\subsection{Outline.}\label{outline1} %% 1.2.
%\subsubsection{Cyclic Schedules for the General Deterministic SMDP.}
%  \label{cyclic-schedules} %% 1.2.1
%\section{Problem Description.}\label{problemdescription} %% 2.

% Text of your paper here
\section{Introduction}
There is an increasing amount of data being collected and stored on customers and sales histories. This has led to the development of more targeted pricing algorithms, in which firms try to increase revenues by offering customers contextual prices that depend on the attributes of the customer and/or product.  %We focus on the fundamental case of single product pricing by a monopolist, with no inventory constraints. 
There has been interest in utilizing abundant historical, posted-price data on whether each customer purchased a particular item at the price they were offered. As an example, we explore offering customers personalized prices for grocery items, based only on previous transaction data %on whether each customer purchased an item based on the price they were offered on that day 
and demographic information from their participation in a loyalty program. Using such observational data can be less costly than running randomized trials \citep{dube2017scalable}, or online algorithms that balance exploration and exploitation over time \citep{den2015dynamic}, either of which can lead to a significant loss of revenue in the short term. Our observational posted-price setting has received less attention than the setting where the seller has access to customer valuation (willingness to pay) samples or distribution information \citep{mohri2014learning, dhangwatnotai2015revenue, devanur2016sample, medina2017revenue, huang2018making, babaioff2018two, daskalakis2020more, allouah2021revenue, beyhaghi2021improved}. A challenge in the posted-price setting is that we do not observe counterfactual data on whether customers would have purchased if offered different prices, an instance of the fundamental problem of causal inference \citep{holland1986statistics}. Furthermore, the observed data is influenced by the previous pricing policy, which can make it difficult to estimate how customers will respond to uncommon prices \citep{shalit2017estimating}.

% Access to valuation samples is common when setting reserve prices for a second price auction, which is a motivating application for much of this work. In this setting a valuation sample corresponds to bids customers submit, which are equivalent to valuations, by the truthful mechanism of second price auctions. 

%Furthermore, firms are increasingly requiring that pricing algorithms are interpretable \citep{bastani2017interpreting, bastani2018interpreting, amram2020optimal, biggs2021model}. This transparency enables firms to understand how the algorithm is pricing items, verify this matches their intuition, and ensure the algorithm satisfies any regulatory requirements. Within the class of interpretable policies, we primarily focus on linear pricing policies, whereby the price is a linear function of customer and/or product features. 
\bla

% \blu Can I put something in about black box being bad? Maybe borrow some interpretabilty language from the other paper or Wei's, find exploration citation. Say something somewhere about demand being unknown \bla

 A popular approach in practice is the predict then optimize, or direct method, whereby an intermediate contextual demand function is estimated to predict the probability of a customer purchasing at a given price, and then optimized to maximize revenue \citep{chen2015statistical, ferreira2016analytics, dube2017scalable, alley2019pricing, baardman2020detecting, biggs2021model}. In practice, often sellers do not know the functional form of demand, leading to a model selection problem. Although advances in machine learning have introduced new models that can capture the complexities of contextual demand more accurately, these models can result in complex revenue maximization problems. Note that this is a consequence of the choice of the demand \textit{model} used, rather than the inherent properties of the \textit{true} demand. For example, although tree ensemble or neural network models can result in accurate demand estimation \citep{ferreira2016analytics,mivsic2017optimization,chen2021assortment, feldman2021customer}, both are highly nonlinear functions of their inputs, resulting in a nonlinear estimated revenue surface when the price is optimized. This holds even when considering simple pricing policies, such as linear functions. Furthermore, it is unclear if there exist bounds on expected revenue from optimizing such complex demand functions.

A more direct approach is to formulate a loss function for the pricing setting and optimize such a loss directly through empirical risk minimization without estimating demand first. At a high level, a loss function provides a way to measure how well a policy performs directly from data. 
Unlike in typical supervised learning settings, we do not observe the ideal price to charge each customer (their valuation or willingness to pay), which could potentially be considered the labels we are trying to learn. As such, it is not clear how to use the distance from this label as our criteria, as is often done in regression. In the classification setting, a surrogate loss function is often justified through desirable properties of the prediction function that minimizes it, such as Bayes consistency (predict 1 if $P[Y|X]>0.5, 0$ otherwise, which is the optimal classification policy \citep{bartlett2006convexity}). It is less clear how to define and determine desirable properties of a loss function in our posted price setting. We propose convex pricing loss functions where the pricing policy obtained from optimizing the loss function has attractive expected revenue bounds. However, unlike the classification setting, we show that there is no convex loss function that can always find the optimal pricing policy, so there is always some gap in  expected revenue.

\section{Contributions}
We propose loss functions for contextual pricing with observational posted price data, whereby customers are offered a price as a function of the customer and/or product attributes, which address the aforementioned issues.
\begin{itemize}
\item \textbf{Computational efficiency:} We propose loss functions that are \textit{convex}, so they can be optimized in a computationally efficient manner. This is particularly relevant when implementing linear pricing policies, which are often desirable for interpretability and generalizability, as this results in a convex revenue optimization problem. The importance of interpretable pricing is documented in \cite{amram2020optimal} and \cite{biggs2021model}. Transparency lets sellers understand how the algorithm is pricing items, verify this matches their intuition, and ensure the algorithm satisfies regulatory requirements. 

% put in something about how we can characterterize the pricing rule.

\item \textbf{Characterization of pricing policies:}  We propose two loss functions for the pricing setting, where the behavior of the pricing policy that results from optimizing the loss function can be characterized and understood. This is in contrast to existing loss functions for pricing where the behavior of the optimal policy is unclear (e.g., \cite{ye2018customized}). The first function we propose is a \textit{hinge pricing loss function}, which shares similarities with the classification hinge loss function but is adapted to the posted-price setting. We show that despite not observing valuation data, the resulting pricing policy prices at the scaled conditional expected value of the valuation distribution. The second is a \textit{quantile pricing loss function}, which prescribes prices at a chosen quantile of prices at which similar customers purchased. %The second is a \textit{hinge pricing loss function}, which has similarities with the classification hinge loss function but is adapted to the posted-price setting. This function captures the intuition that if the item sells at the posted price, a higher price should be charged, while if it doesn't sell, the price needs to be lowered. We derive optimality conditions for these loss functions and show that despite not observing valuation data, the resulting pricing policies price at the scaled conditional expected value and at a quantile of the valuation distribution, respectively. 
Since both of these approaches are dependent on a scaling parameter, we show how this can be chosen in a robust manner to maximize revenue against an adversarial valuation distribution. 
%and prove bounds on the expected revenue from resulting pricing policies relative to the optimal revenue achievable from any contextual pricing policy, 

\item \textbf{Expected revenue guarantees:} The loss functions we propose have bounds on the expected revenue relative to the optimal revenue achievable from the optimal contextual pricing policy that has access to the customer valuation distribution. Assuming the complementary cumulative distribution function (survival function) of customer valuations is log-concave, we prove revenue bounds of 0.772 for the hinge pricing policy and 0.749 for the quantile pricing policy against an adversarially chosen valuation function. These bounds are stronger than 0.5 bounds proved in \cite{chen2021model} in a setting that is closely related but restricted to uniform historical pricing policies.  %This analysis also provides insight into why the loss function in \cite{ye2018customized} performs well in practice.

\item \textbf{Finite-sample bounds:} In addition to the expected performance, we also show generalization bounds for a finite sample of observations for a class of linear policies that satisfies the computational efficiency requirements discussed above.

\item \textbf{Competitive  empirical performance:} Finally, we provide simulations on both synthetic and real-world data to demonstrate that this approach is competitive with, and in some settings outperforms, state-of-the-art methods in contextual pricing, which may work well in practice but do not have known expected revenue guarantees. In particular, we empirically show that using machine learning functions such as gradient-boosted trees to estimate demand often results in poor pricing policies due to the difficulty of optimizing a non-convex function, while simpler functions such as logistic regression often do not capture the non-linear interactions in demand, resulting in sub-optimal performance relative to the convex surrogates pricing loss functions.
\end{itemize}

\section{Other related literature}

A significant body of literature focuses on the online pricing setting, where the seller chooses prices to balance learning and earning over time \citep{kleinberg2003value, gallego2006dynamic, feng2010integrating, broder2012dynamic, harrison2012bayesian, cheung2017dynamic, besbes2018dynamic, den2020discontinuous, calmon2021revenue,keskin2021data}. Within this area, there has also been a focus on incorporating contextual data to offer more targeted prices \citep{javanmard2016dynamic, qiang2016dynamic, bertsimas2017data, cohen2018dynamic, nambiar2019dynamic, cohen2020feature,ban2020personalized, zhang2021data, chen2021nonparametric, liu2021optimal}. \blu Of particular relevance are \cite{amin2014repeated}, and \cite{cohen2020feature}, who study online contextual pricing with access to similar data where they observe whether a sale occurred at the price offered. \cite{amin2014repeated} study a setting where the customers' valuations are deterministic (no noise) around an unknown linear function. \cite{cohen2020feature} consider a noisy setting where a linear valuation with an unknown mean is perturbed by an error term that has a known distribution. The distribution of the error term is used in optimizing prices. In contrast, we study a noisy setting where the distribution of the error is unknown, but comes from a log-concave distribution. \bla  A comprehensive review of earlier work in dynamic pricing can be found in \cite{den2015dynamic}. %There has also been work in the Sequential Posted Price (SPP) setting where customers are offered sequential prices are offered and over time and data observed on whether customer purchased \citep{calinescu2011maximizing, yan2011mechanism, agrawal2012price, chawla2015power, azar2018prophet, beyhaghi2021improved}, similar to the setting we study but in an online environment and without covariates. 
As experimentation can be costly, we focus on a setting where we want to maximize revenue in the short term, using only existing observational data.

% \cite{chen2019distribution} provide bounds when only mean and variance of valuation distribution is known.
%\blu maybe should also mention the data-rich regime\bla 

An important differentiating feature in the pricing literature is how much information the seller has access to. There is a substantial body of work that focuses on a seller with limited samples of valuation data \citep{dhangwatnotai2015revenue, huang2018making, babaioff2018two, daskalakis2020more, derakhshan2020data, allouah2021revenue}, including contextual side information \citep{mohri2014learning, devanur2016sample, medina2017revenue}. In the setting with a single valuation sample and no contextual data, assuming the valuation distribution is regular, \cite{dhangwatnotai2015revenue} prove expected revenue bounds of $0.5$ by setting the next customer's price equal to the valuation of the first customer. \cite{huang2018making} improve the revenue bound to $0.589$ by pricing at a $0.85$ multiplicative factor of the valuation observed, provided the valuation distribution is log-concave. \cite{daskalakis2020more} study the case with two samples and prove bounds of $0.558$ for regular distributions, while \cite{allouah2021revenue} improve to $0.615$ and also provide upper and lower bounds for any number of samples in the regular and log-concave valuation distribution setting, using a dynamic programming approach. \cite{huang2018making} also study a \textit{data-rich} regime, showing that in order to find a $(1-\epsilon)$ optimal price from valuation data, the sample complexity scales polynomially in $1/ \epsilon$. 

Algorithms also exist under alternative knowledge about the valuation distribution.  \cite{cohen2015pricing} provide bounds when only the support of the valuation distribution is known; \cite{azar2013optimal} and \cite{chen2019distribution} use the mean and variance of the valuation distribution; \cite{elmachtoub2020decision} uses the coefficient of deviation of the valuation distribution; \cite{bergemann2011robust} use a neighborhood containing the true valuation distribution. In contrast to all this work, we have posted-price samples on whether the item sells or not at the price consumers were given, rather than samples or other knowledge of the valuation distribution. Hence, the observed posted-price sales samples are affected by the previous pricing policy, while the valuation samples are not. This needs to be accounted for in any pricing algorithm. Some of this work also focuses on the ``value of price discrimination" rather than practical algorithms for pricing that can be solved tractably.

% maybe remove that last sentence 

Closest to our work are efforts to formulate contextual pricing loss functions that can be optimized directly to find pricing policies. \cite{mohri2014learning} propose a loss function when setting reserve prices for a second price auction, while assuming valuation samples (the maximum each customer is willing to pay) are available. The surrogate functions they propose are continuous but non-convex, hence still challenging to optimize. \blu \cite{huchette2020contextual} provide strong Mixed Integer Programming (MIP) formulations for this loss function, which allows the global optimal solution for small problem instances to be found. They also provide linear relaxations. \bla Similar to our setting, \cite{biggs2021loss} propose pricing loss functions for the observational posted-price setting when prices are restricted to a discrete price ladder. Here we focus on continuous prices. In the same setting we study, \cite{ye2018customized} propose a customized $\epsilon$-insensitive loss, used for contextual pricing at Airbnb, which is related to the functions we study. We further discuss this function in detail in Section \ref{sec:hinge_loss}. Another very related paper is \cite{chen2021model}, which offers model-free pricing for assortments. While \cite{chen2021model} do not consider contextual pricing for the single-item case they observe the price that the product is purchased. As such, the setting for their revenue bounds is very similar to ours; these settings are further compared in section \ref{sec:exp_rev_bounds}.   %This captures the intuition that if the item sold at the price it was offered, it is desirable to offer a higher price, while if it didn't sell, a lower price should likely be offered. 
%While this approach is shown to perform well in simulations, there are no guarantees on what the revenue will be from following a pricing policy obtained by optimizing this loss function.

% put in Joey's paper neurips paper, also negin's paper on LP based relaxations

% \blu Need to be specific about the "bounds" referred to here. M

A potential approach to pricing using observational data is to use methods from the off-policy learning literature such as inverse propensity scoring (IPS)  \citep{rosenbaum1983central,beygelzimer2009offset,li2011unbiased}. Although  IPS methods are typically used when the treatment is binary, there have been extensions to continuous treatments \citep{austin2015moving, kallus2018policy}, which better model pricing. In \cite{kallus2018policy}, the reward is estimated using a weighted average according to how far the given treatment in the historical data is from the proposed policy, as evaluated by a kernel. The approach of \cite{kallus2018policy}, however, leads to non-convex optimization problems for most practical choices of the kernel. There is a recent stream of literature that applies some ideas from off-policy learning to pricing problems with censored demand, which occurs due to a lack of inventory \citep{ban2020confidence, bu2022offline, qi2022offline}. In contrast, our model doesn't incorporate inventory effects but focuses on the different setting where the observed purchase outcomes are binary, as may occur in a personalized pricing setting.

\section{Model}
\label{sec:model}

We study a fundamental contextual pricing problem of a monopolist selling a single product with no inventory constraints. The monopolist wants to set prices based on historical data to maximize sales in the short term. Each customer is described by features $X \in \mathcal{X}$ and has an unobserved valuation $V \in \mathbb{R}_+$, which is the maximum amount they are willing to pay, both drawn from a joint distribution. The customer is offered the realization $P \in \mathbb{R}_+$ of a stochastic price from a historical pricing policy that follows a known conditional distribution with density $\phi(P|X)$. This assumption supposes that the practitioners know or have recorded the pricing policy used in the past. One setting where this is likely satisfied is when the firm is already doing algorithmic pricing. We observe whether the customer purchases $Y \in \{0,1\}$, depending on whether the price is above or below their valuation. Specifically, 

\begin{equation}
Y(P) =    \begin{cases}
      1   & \text{if} ~~ P \leq V \\
      0   & \text{if} ~~  P > V   \\
    \end{cases}  \label{y_def} 
\end{equation}

  We do not observe the counterfactual outcomes associated with the customer being given a different price from what was assigned by the previous pricing policy, nor do we observe the valuation of each customer. As such, we have access to an i.i.d. dataset of samples $ S_n = \{(Y_i,P_i,X_i)\}_{i=1}^n$. For identifiability \citep{swaminathan2015counterfactual}, we assume overlap and ignorability.

%\blu Need to put in that we look at the single item, no inventory, not dynamic etc\bla

\begin{assumption}(Overlap)
$ \phi(p|X) > 0, \quad\forall p\in \mathbb{R}_+$
\end{assumption}

\begin{assumption}(Ignorability)
$Y(p) \perp   P | X, \quad \forall p ~\in \mathbb{R}_+$
\end{assumption}

The overlap assumption requires that each price has a nonzero probability of being offered to each customer. The known impossibility result of counterfactual evaluation also applies when it is not satisfied \citep{langford2008exploration}. The ignorability assumption requires that there be no hidden confounding variables influencing the pricing decision and the customers' purchasing decision. It is commonly satisfied as long as the factors that drove historical pricing decisions are available in the observed data \citep{bertsimas2020predictive}. For clarity of exposition, we also assume that all customers have a positive valuation, meaning all customers will purchase if given the product for free, $\mathbb{P}(V > 0| X)= \bar{F}_V(0)=1$, and that the complementary CDF (survival function) $\bar{F}_V(v)= \mathbb{P}(V > v| X)$, is log-concave. Recall that a function is log-concave if its domain is a convex set and it satisfies 
\begin{equation*}
f(\theta x + (1-\theta)y) \geq    f(x)^{\theta} f(y)^{1-\theta} \qquad \forall ~ x,y \in \text{dom} ~f, ~0<\theta<1
\end{equation*}

 %\blu Maybe we need to put an upper bound here \bla
%  \blu Comment on how this relates to the PDF, can we find other pricing papers which use a log-concave function? \bla

% \begin{assumption}(Will buy for free)
% $\bar{F}_V(0)=1$
% \end{assumption}
% maybe say something about the comparing this to supervised loss functions. Maybe also talk about how we can use valuation loss function form mohri and medina.
The log-concavity assumption encompasses a broad range of valuation distributions including commonly used distributions such as normal, exponential, and uniform \citep{bagnoli2005log}. Furthermore, it is known that if the density function is log-concave, then so is the complementary CDF \citep{bagnoli2005log}. Log-concavity also implies that the hazard rate is monotone, a common assumption in pricing \citep{cole2015sample,huang2018making,allouah2021revenue}. Without log-concavity, there are simple examples showing that the revenue gap may be unbounded \citep{cole2015sample}. 

% Our goal is to find a randomized pricing policy, $\pi \in \Pi: \mathcal{X} \rightarrow \Delta, \Delta :\{q(p): q(p)\geq 0 , \int_{\mathbb{R}_+}  q(p) dp =1 \} $ which gives us a distribution of prices  $\pi(p|X)$ to offer a customer once  their features have been observed. The expected revenue obtained from a policy conditioned on $X$ is:

Our goal is to find a pricing policy, $\pi \in \Pi: \mathcal{X} \rightarrow \mathbb{R}_+$, which prescribes a price for each customer. The expected revenue obtained from a policy conditioned on $X$ is:

%Note maybe we should define as a randomized pricing policy
% $$
% \mathcal{R}(\pi(X)) = \mathbb{E}_{V}[\pi(X) \mathbbm{1}\{\pi(X) \geq V\}|X], ~~~ \pi^*(X) = \argmax \mathcal{R}(\pi) 

% $$
\begin{equation}
\label{eq:valuation_loss_fn}
% \mathcal{R}(\pi(X)) = \mathbb{E}_{V}\left[\int_{\mathbb{R}_+} \pi(p|X) p \mathbbm{1}\{p \leq V \} dp|X \right]
\mathcal{R}(\pi(X)) = \mathbb{E}_{V}[\pi(X) \mathbbm{1}\{\pi(X) \leq V\}|X]
\end{equation}

% \blu Should put in cost? assume that it is constant (or 0)?\bla

% \begin{figure}  
% \centering  
%             \begin{tikzpicture}[scale=0.8] 
%             \begin{axis}[ axis lines = left,
%                 xlabel = \(\pi(X)\),
%                 ylabel = {\(\pi(X) \mathbbm{1}\{\pi(X) \leq V\}\)},
%                 ymin=0, 
%                 ymax=1.1,
%                 xtick={1},
%                 xticklabels={$V$},
%                 ytick={1},
%                 yticklabels={$V$},
%                 x label style={at={(axis description cs:1,0.1)},anchor=north},
%                 y label style={at={(axis description cs:0.07,0.8)},anchor=south}]
%             \addplot[domain=0:1,black,ultra thick,dash dot] {x};
%             \addplot[domain=1:2,black,ultra thick,dash dot] {0.01};
%             \addplot[black, mark=*, only marks] coordinates {(1,1)};
%             \draw[black,dash dot] (axis cs:1.01,1) -- (axis cs:1.01,0);
%             \end{axis}
%             \end{tikzpicture} 
% \caption{Example of the function with valuation samples optimized in \cite{mohri2014learning} }
% \label{fig:valuation_loss}
% \end{figure}  

% \begin{figure}[]
% 	\centering
% 	\includegraphics[width=0.8\textwidth]{fig/rev_bounds_quantile_better.pdf}
% 	\caption{Revenue bounds from Theorem \ref{quantile_revenue} as a function of $\tau$ with upper bound from Lemma \ref{lemma_upper_bound}}
% 	\label{fig:rev_bounds_quantile}
% \end{figure}

\begin{figure}  
\centering  
	\includegraphics[]{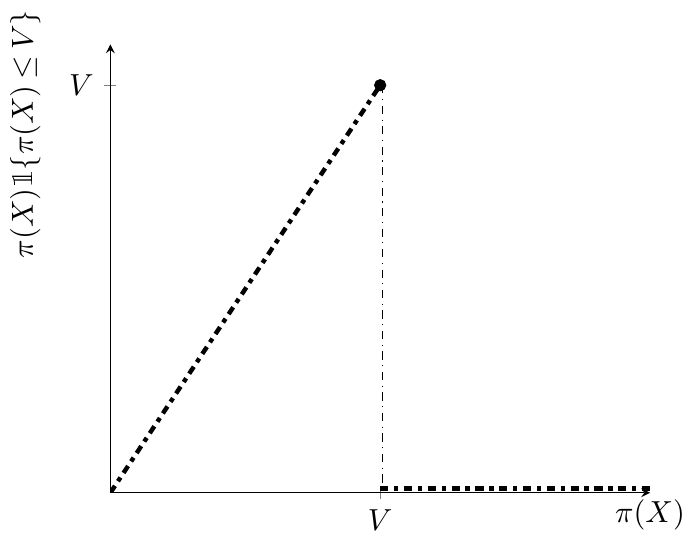}
\caption{Example of the function with valuation samples optimized in \cite{mohri2014learning} }
\label{fig:valuation_loss}
\end{figure}  

That is, if the price offered $\pi(X)$ is less than the customer's valuation $V$, the item sells and the revenue obtained is the price offered, while if the price is above the customer's valuation the revenue is zero.  %This can also be modified into profit by introducing a per unit cost, but for simplicity, we will focus on maximizing revenue. 
Unfortunately, it is not possible to directly optimize $\mathcal{R}(\pi(X))$  because the distribution of customer valuations is not known and samples are not observed. If valuation samples are observed, a pricing policy could be found by optimizing an empirical version \citep{mohri2014learning}: 

$$ \argmax_{\pi \in \Pi} \frac{1}{n} \sum_{i=1}^n \pi(X_i) \mathbbm{1}\{\pi(X_i) \leq V_i\}$$

This is visualized in Figure \ref{fig:valuation_loss}. However, optimizing this function, which is non-convex and discontinuous, is challenging \citep{mohri2014learning, huchette2020contextual}.
%In the same setting, \citet{medina2017revenue} propose polynomial-time algorithms based on access to an estimate of the valuation, such as that obtained by regressing on the valuation samples.
% Mohri a Medina talk about how this is quasi-convex, but the sum of non-convex functions (for example empirical risk minimization) is, in general, not quasi-convex.
Instead, we need to find a loss function that provides good prices when optimized using the samples we observe $Y, P$, rather than valuations $V$.

\subsection{Convex relaxations}

In particular, we aim to find loss functions that are convex, so they can be optimized efficiently, yet still have provable guarantees on expected revenue, relative to the optimal contextual pricing policy which has access to the valuation distribution, $p^*=\argmax_{\pi} \mathcal{R}(\pi(X))$, where the dependence on $X$ is henceforth omitted for notational clarity. We start by showing an impossibility result, which states that there is no loss function with access only to posted price data that is able to recover the optimal price for all valuation distributions. This Proposition is a minor extension of Theorem 2 of \citet{mohri2014learning}, which proves that there is no convex surrogate for which the global minimum price is obtained in the case with access to valuation data. We denote a loss function using observational posted-price data as $L(\pi(X),Y, P): \mathbb{R}_+  \times \{0,1\} \times \mathbb{R}_+ \rightarrow \mathbb{R}$. % belonging to the class of functions which are convex in their first argument $\mathcal{L}$.

% \begin{lemma}
% There is no non-constant function $L_O$ convex with respect to its first argument such that for any distribution $D$ on $\{0,1\} \times \mathbb{R}_+$, there exists a non-negative minimizer $p^* \in \mathcal{R}(\pi(X))$, such that $ \min_p \mathbb{E}_{Y,P}  L_O (p, Y,P) = \mathbb{E}_{Y,P}  L_O (p^*, Y,P)$.
% \end{lemma}

% \begin{lemma}
% \label{lemma:no_opt_convex_surrogate}
% There is no non-constant function $\mathbb{E}_{Y,P}[L(\cdot, Y,P)|V]$, with $L (\cdot, Y,P)$ convex with respect to its first argument and $\lim_{v \rightarrow v_0^-} \mathbb{E}_{Y,P}[L(v_0^-, Y,P)|V=v]=\mathbb{E}_{Y,P}[L(v_0, Y,P)|V=v_0]$ such that for any distribution $(Y,P,V) \sim D$ on $\{0,1\} \times \mathbb{R}_+ \times \mathbb{R}_+$ satisfying Equation \ref{y_def}, there exists a non-negative optimal price $p^* \in \argmax_{p} \mathcal{R}(p)$, such that $ \min_p \mathbb{E}_{Y,P} [ L (p, Y,P)] = \mathbb{E}_{Y,P} [ L (p^*, Y,P)]$.
% \end{lemma}

\begin{proposition}
\label{prop:no_opt_convex_surrogate}
There is no nonconstant function $\mathbb{E}_{Y,P}[L(\cdot, Y,P)|V=v]$, left continuous in $v$ and with $L (\cdot, Y,P)$ convex with respect to its first argument, such that for any distribution $(Y,P,V) \sim D$ on $\{0,1\} \times \mathbb{R}_+ \times \mathbb{R}_+$ satisfying Equation \ref{y_def}, there exists a nonnegative optimal price $p^* \in \argmax_{p} \mathcal{R}(p)$, satisfying $ \min_p \mathbb{E}_{Y,P} [ L (p, Y,P)] = \mathbb{E}_{Y,P} [ L (p^*, Y,P)]$.
\end{proposition}

%\blu Need to comment on some of these assumptions \bla
The requirement of left continuity of the expected loss is a minor technical requirement that follows from a left continuity assumption in \citet{mohri2014learning}. 
This impossibility result contrasts with the well-studied classification setting, where there are convex surrogates for the $0/1$ loss that can recover the optimal classification policy, a property known as Bayes consistency or classification calibration \citep{bartlett2006convexity}. \citet{huchette2020contextual} also show that optimizing the function $\max_{\theta} \sum_i \langle \theta, X_i \rangle  \mathbbm{1}\{\langle \theta, X_i \rangle \leq V_i \}$ is NP-Hard in the setting where valuation samples are available with a linear policy class. %Our problem is more difficult, since if the historical pricing was perfect (each customer was offered exactly their valuation), our problem would reduce to this. -- eeehhhh not quite true when we have stochastic data.
For convenience, we provide commonly used notations in Table \ref{notations}.

\begin{table}[]
\centering
% \setlength\extrarowheight{-8pt}
%\begin{tabular}{>{\raggedright\arraybackslash}m{20mm}m{145mm}}
\begin{tabular}{l l}
\toprule
% \multicolumn{1}{>{\arraybackslash}m{20mm}}{\textbf{Notation}} 
%     & \multicolumn{1}{>{\centering\arraybackslash}m{145mm}}{\textbf{Description}}\\
Notation & Description   \\
\hline
$V$         &   Valuation of the customer (\textit{unobserved})    \\ 
$Y$         &   Whether the customer purchased the item   \\ 
$P$         &   Price posted in the past  \\ 
$X$         &   Contextual features describing the customer and or product  \\ 
$\phi(P|X)$ &   Conditional historical probability density of a price being offered to a customer \\
$\pi(X)$ &  Contextual pricing policy (decision) \\
$\mathcal{R}(\cdot)$ &   Expected conditional revenue \\
$p^*$         &   Optimal price \\
$c$         &   Parameter the seller can set for hinge pricing loss \\
$\tau$         &   Parameter the seller can set for quantile pricing loss \\
$p_h$         &   Price set by the hinge pricing loss function \\
$p_q$         &   Price set by the quantile pricing loss function \\
\toprule
\end{tabular}
\caption{A summary of frequently used notation}
\label{notations}
\end{table}

\begin{figure}[]
	\centering
	\subfloat[$Y=1$]{\includegraphics[]{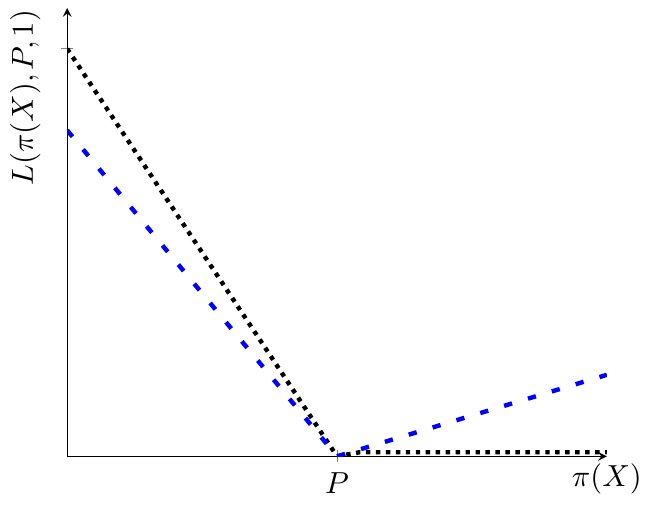}}
	\subfloat[$Y=0$]{\includegraphics[]{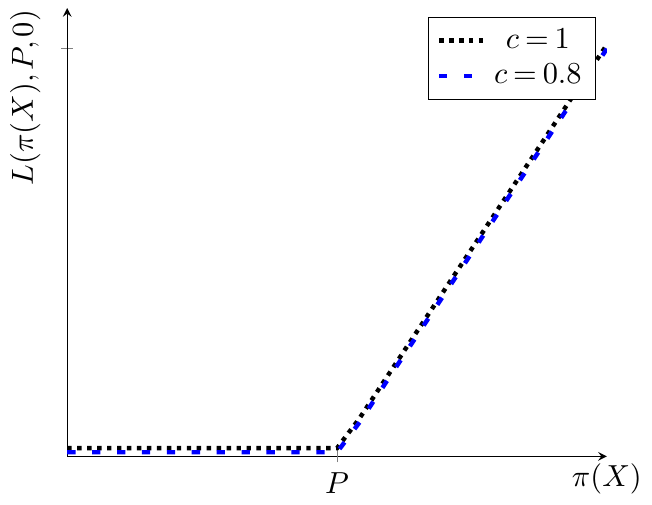}}
\caption{Example of the pricing hinge loss function}
\label{fig:hinge_loss}
\end{figure}

% \begin{axis}[
%     extra y ticks={2.18},
%     extra y tick style={
%         grid=major,
%         yticklabel={$P_c$},
%         yticklabel style={yshift=0.7ex, anchor=east}}]
% \addplot+[samples=100,domain=-2:2] {x^2};
% \end{axis}

\section{Pricing hinge loss function}
\label{sec:hinge_loss}

% \begin{figure}[]
% 	\centering
% 	\includegraphics[width=0.9\textwidth]{fig/hinge_loss.pdf}
% 	\caption{Example of pricing hinge loss function}
% 	\label{fig:hinge_loss}
% \end{figure}

Since it is not possible to find a convex loss function that recovers the optimal pricing policy, we focus on loss functions that are capable of achieving a high proportion of the optimal revenue.  We propose the hinge pricing loss function, which has similarities to the hinge loss function used for classification tasks. %While we can prove theoretical revenue bounds using this loss function, they aren't as strong as the quantile loss function in the worst case. However, it does perform well in practice and is intuitive. 
 This loss function is defined in Definition \ref{pricing_hinge_loss} and visualized in Figure \ref{fig:hinge_loss}.  

\begin{definition} The hinge pricing loss function is given by
\label{pricing_hinge_loss}
$$L^h_{c}(\pi(X),Y,P)= \frac{1}{\phi(P|X)}[c Y (P-\pi(X))^+ + (1-cY)(\pi(X)-P)^+]$$
\end{definition}

Where $c$ is a parameter chosen by the seller, which we will discuss how to choose later in this section. When $c=1$, this loss function penalizes prices that are below the listed price when the item was sold and penalizes prices that are above the listed price when the item wasn't sold, giving the loss function the characteristic ``hinge" shape. This is reasonable since if an item sold, the customer's valuation is likely higher than the listed price, so pricing above the listed price is more attractive. Conversely, if an item did not sell, then the customer's valuation is likely below the listed price, so pricing below this price should be encouraged. The parameter $c$ controls how aggressive the pricing policy is. For $c<1$, when the item sells, there is a reduced penalty for pricing below the sale price and even a small penalty for pricing above the sale price. The penalty remains the same for pricing above the price offered when there is no sale, as shown in Figure \ref{fig:hinge_loss}.

Similar to inverse propensity scoring methods \citep{rosenbaum1983central}, each observation is scaled by a weight that is inversely proportional to the probability of receiving the price, $\phi(P|X)$, to counteract the imbalance due to the historical pricing policy. In practice, a pricing policy from a sample can be found using empirical loss minimization, which is further discussed in Section \ref{sec:generalization_bounds}:

$$ \argmin_{\pi \in \Pi} \frac{1}{n} \sum_{i=1}^n \frac{1}{\phi(P_i|X_i)}[c Y_i (P_i-\pi(X_i))^+ + (1-cY_i)(\pi(X_i)-P_i)^+]$$

% $$ \argmin_{\pi \in \Pi} \frac{1}{n} \sum_{i=1}^n L^h_{c}(\pi(X_i),Y_i,P_i)$$

% \begin{remark}
\subsection{Comparison with \cite{ye2018customized}}
\label{sec:airbnb_loss}
The hinge pricing loss shares a similar motivation to the loss function from \cite{ye2018customized}, and in certain settings for extreme parameter choices, the loss functions are the same. \cite{ye2018customized} proposes a customized $\epsilon$-insensitive loss  used for contextual pricing at Airbnb:
\begin{equation*}
    L(\pi(X),Y,P) = Y[(P-\pi(X))^+ + (\pi(X)-c_1P)^+] + (1-Y)[(\pi(X) - P)^+ + (c_2P - \pi(X))^+]
\end{equation*}

\begin{figure}[]
	\centering
	\subfloat[$Y=1$]{\includegraphics[]{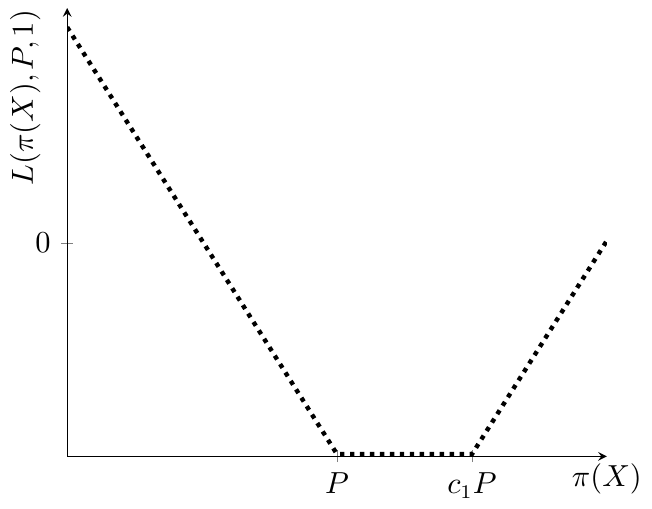}}
	\subfloat[$Y=0$]{\includegraphics[]{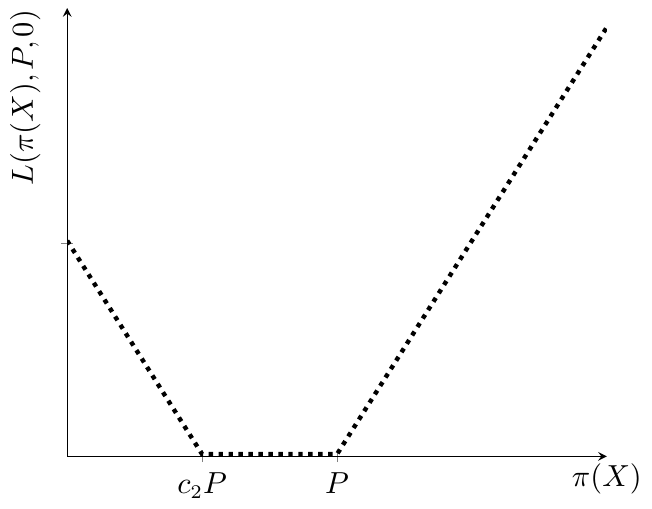}}
\caption{Example of pricing loss function in \citet{ye2018customized}}
\label{fig:airbnb_loss}
\end{figure}

This loss function is shown in Figure \ref{fig:airbnb_loss}. Similar to the pricing hinge loss, this captures the intuition that if the item sold, then the customer's valuation is likely higher than the price the customer was offered, and the future price offered should be raised. However, the loss function from \cite{ye2018customized} additionally introduces the idea that the price shouldn't be raised too much, since eventually the customer will not purchase if the price gets too high. This is reflected in the function where there is no loss between $P$ and $c_1P$, but an increasing loss incurred above $c_1P$, where $c_1 > 1$. This gives the loss function a characteristic ``U" shape (the authors highlight the similarity with $\epsilon$-insensitive regression), compared to the hinge shape of the loss function we propose. Likewise, for the items that don't sell, both loss functions incur a loss for pricing above the listed price at which no sale occurred, pressuring the price downward. However, the loss function from \cite{ye2018customized} additionally incurs a loss for prices that are below  $c_2P$, where $c_2 < 1$. The intuition is that if the price is far below the customers' valuation, there will be potentially unrealized revenue. 

While this approach from \cite{ye2018customized} is intuitive and has been shown to perform well in practice, there are no guarantees on what the expected revenue is for the resulting policy. Furthermore, the behavior of the pricing policy is unclear, as the policy obtained by optimizing this loss function is not characterized. Finally, there is little theoretical justification for the choices on $c_1$ and $c_2$, which are left to the user to choose. Due to these drawbacks, it is unclear whether the additional arms that differentiate the \cite{ye2018customized} loss and hinge loss are justified. %We will address all these issues for the hinge loss function and give insights into why it is effective.  

Another key distinction is that the loss from \cite{ye2018customized} does not take into account the historical pricing policy, whereas the hinge loss divides by the propensity score $\phi(P|X)$ to counteract the imbalance. As such, unless the historical pricing policy is uniform (such as would arise from a randomized trial), this is likely to affect the perfromance of this policy. 

However, due to the similarities between the two loss functions, we are able to provide revenue guarantees for the loss function from \cite{ye2018customized}, for a restricted class of settings. In particular, when the parameters in \citet{ye2018customized} are set to $c_1= \infty$ and $c_2= -\infty$, and the historical pricing policy is uniform, then the loss proposed in \citet{ye2018customized} is the same as the hinge pricing loss function for $c=1$. In this case, our bounds and insights apply to \citet{ye2018customized}. 
% \end{remark}

\subsection{Characterization of the pricing policy}

% Therefore, pricing below the price the customer was given ($\pi(X) \leq P$) results in an opportunity cost as a higher price could have been charged with the customer still purchasing. To increase revenue, the customer should be given a price higher than the price they were offered, although it is not clear how much higher, since eventually the customer will not purchase if the price gets too high. This is reflected in in there being no loss between $P$ and $c_1P$, but an increasing loss being incurred above $c_1P$. Similarly, if the item doesn't sell for a given customer, then it is likely the price should be lowered, but again, it is not clear by how much. 

% One drawback of this approach is that it is not clear how to choose constants $c_1$ and $c_2$. An advantage, of this approach relative to the predict then optimize approaches, is that the loss function is convex so it is easier to optimize. While this is desirable, it is of little use if the policy it converges to is not justified. 

% While this approach is shown to perform well in simulations and is intuitive, there are no guarantees on what the revenue will be from following a pricing policy obtained by optimizing this loss function.
% Furthermore, there isn't any analysis of how this approach performs for the pricing setting in particular.

 We can characterize the behavior of the policy that results from using the hinge pricing loss function. When we minimize the conditional expectation of this loss function, we end up pricing at the conditional expected valuation for each customer, scaled by a parameter $c$, despite not observing the valuation data. This is proved in Lemma \ref{lemma_opt_cond_hinge}. As mentioned previously, the parameter $c$ controls how aggressive the pricing policy is, with $c<1$ resulting in prices less than the expected valuation and $c>1$ resulting in prices above the expected valuation. We focus on $c<1$, as this results in stronger revenue bounds. We provide guidance on how to choose $c$ robustly following our analysis. If we condition on $X$ and for notational simplicity denote $p_h=\argmin_{\pi(X)} \mathbb{E}_{Y,P}[L^h_{c}(\pi(X),Y,P)|X]$, we can prove Lemma \ref{lemma_opt_cond_hinge}.

\begin{lemma}
\label{lemma_opt_cond_hinge} (Hinge pricing loss optimality condition): Let $p_h$ be the minimizer of the hinge pricing loss function, then $$p_h= c \mathbb{E}[V|X]$$ 
\end{lemma}

\proof{Proof of Lemma \ref{lemma_opt_cond_hinge}:}
% \begin{align*}
% \mathbb{E}_{Y,P}&[L^h_c(p_h,Y,P)|X]  \\ 
% & = \int_p \left( \frac{c \mathbb{P}(Y=1|p,X)}{\phi(p|X)}  (p-p_h)^+ + \frac{1-c \mathbb{P}(Y=1|p,X)}{\phi(p|X)}(p_h-p)^+ \right) \phi(p|X) dp \\
% & = \int_{p_h}^{\infty} c \mathbb{P}(Y=1|p,X) (p-p_h) dp +  \int_0^{p_h} (1 - c \mathbb{P}(Y=1|p,X)) (p_h-p) dp
% \end{align*}

\begin{align*}
\mathbb{E}_{Y,P}[L^h_c(p_h,Y,P)|X]  & = \int_p \left( \frac{c \mathbb{P}(Y=1|p,X)}{\phi(p|X)}  (p-p_h)^+ + \frac{1-c \mathbb{P}(Y=1|p,X)}{\phi(p|X)}(p_h-p)^+ \right) \phi(p|X) dp \\
& = \int_{p_h}^{\infty} c \mathbb{P}(Y=1|p,X) (p-p_h) dp +  \int_0^{p_h} (1 - c \mathbb{P}(Y=1|p,X)) (p_h-p) dp
\end{align*}

% Where the second equality follows since $\mathbb{E}[Y|p,X]=\mathbb{P}(V>p|X)$ from Equation \ref{y_def}. 
Applying the Leibniz rule to find the optimality conditions:
\begin{align*}
\diffp{}{{p_h}}\mathbb{E}_{Y,P}[L^h_c(p_h,Y,P)|X]  & = \int_{p_h}^{\infty} - c \mathbb{P}(Y=1|p,X) dp +  \int_0^{p_h} 1 - c \mathbb{P}(Y=1|p,X) dp \\
& = \int_0^{p_h} 1 dp - c \int_{0}^{\infty} \mathbb{P}(Y=1|p,X) dp \\
& = \int_0^{p_h} 1 dp - c \int_{0}^{\infty} \bar{F}_V(p) dp  \\
& = p_h - c \mathbb{E}[V|X]  =0 
\end{align*}

Where the third equality follows since $\mathbb{E}[Y|p,X]=\mathbb{P}(V>p|X)$ from Equation \ref{y_def}. Therefore, $p_h= c \mathbb{E}[V|X]$ at optimality.  \Halmos

\endproof

This simple result is important since it provides clarity about the behavior of the pricing policy that results from minimizing the pricing hinge loss function. This is not clear in previous loss functions used for pricing, such as \cite{ye2018customized}.  We also note that this is a different minimizer than that of the standard hinge loss function used in binary classification, which predicts 1 if $P[Y|X]>0.5$ and $0$ if $P[Y|X]<0.5$ \citep{bartlett2006convexity}.

This result implicitly assumes that there are no model misspecification issues that may prevent this minimizer from being reached. In this respect, the result is similar to Bayes consistency, which depends on the underlying distributions %which is used to justify surrogate loss functions in the classification setting based on the conditional expectation of the minimizer 
rather than model-specific bounds for the policy class considered. Model-specific bounds are studied for the case of linear policies in Section \ref{sec:generalization_bounds}. In the next section, we will show the expected revenue bounds that can be achieved by following this  pricing policy.

%The importance is analogous to Bayes consistency in the classification setting, whereby surrogate loss functions are justified through desirable properties of the policy which minimizes them. %Bayes consistent policies predict 1 if $P[Y|X]>0.5, 0$ otherwise, which is the optimal classification policy \citep{bartlett2006convexity})
 
% This result is analogous to the  . In the next section, we will show the expected revenue bounds which can be achieved from following this policy.

\subsection{Expected revenue bounds}
\label{sec:exp_rev_bounds}
For the expected hinge pricing policy, we can find bounds on the optimal expected revenue, relative to the best pricing policy that has access to the conditional valuation distribution and solves  $p^*=\argmax_{p} \mathcal{R}(p)$. Furthermore, these bounds are robust in the sense that this holds across \textit{all} valuation distributions that satisfy the log-concave assumption, including an adversarially chosen distribution. Note these bounds hold in expectation and do not address what happens for a finite sample or a particular policy class, which is addressed in Section \ref{sec:generalization_bounds}.

% \blu Say something about this being close to optimal if there is no uncertainty if $c=1$. \bla

\begin{theorem}
\label{hinge_revenue}
For $0 < c \leq 1$:
\begin{equation*} 
\frac{\mathcal{R}(p_h)}{\mathcal{R}(p^*)} \geq \min \left\{ \min_{z\leq -2c} \frac{cze^{-z(\frac{1}{c}-1)-1}}{z+c} ,~ \min_{0 < f < 1}  \frac{c(f-1)e^{c(f-1)}}{f \ln(f) }  \right\}
\end{equation*}
\end{theorem}

% \begin{theorem}
% \label{hinge_revenue}
% For $0 < c \leq 1$:
% \begin{equation*} 
% \frac{\mathcal{R}(p_h)}{\mathcal{R}(p^*)} \geq \min \left\{  \frac{ct(c)e^{-t(c)(\frac{1}{c}-1)-1}}{t(c)+c} ,~ \min_{0 < f < 1}  \frac{c(f-1)e^{c(f-1)}}{f \ln(f) }  \right\},~ \text{where}~ t(c)= \dfrac{c}{2}\left(\sqrt{\dfrac{c-5}{c-1}} +1 \right)
% \end{equation*}
% % Where $t(c)= \dfrac{1-c+\sqrt{(c-5)(c-1)}}{2(c-1)}$.
% \end{theorem}

To prove this bound we need to examine two cases: the case where the prescribed price is below the optimal price for the adversarial distribution ($p_h < p^*$), and the case where it is above ($p_h > p^*$). Which case is relevant depends on the chosen parameter of $c$. If a lower value of $c$ is chosen (i.e., much below the expected valuation), the adversary will choose a valuation distribution where $p_h < p^*$, which is characterized by many high-valued customers that the lower price $p_h$ doesn't fully exploit. Conversely, if a higher value of $c$ is chosen, the adversary will choose a distribution where $p_h > p^*$, which will be characterized by a long tail where the selling probability at $p_h$ is low, relative to a significantly higher selling probability at the slightly lower price $p^*$. This is explored more in Section \ref{sec:tightness_of_bound}, and the worst-case valuation distributions are characterized in Figure \ref{fig:worst_case_dist}. We prove Lemmas that give revenue bounds in each case, and prove they are tight. %These results rely on the valuation distribution being log-concave to restrict how quickly the valuation distribution can diminish. 
We start with the case where $p_h < p^*$.

\begin{restatable}{lemma}{lemmaexpgreat}
\label{revenue_case_1}
Consider a pricing policy that prescribes a price $p_h$ such that $p_h \geq c \int_0^{\infty} \bar{F}_V(p) dp$ and $p_h < p^*$, then:
$$ \frac{\mathcal{R}(p_h)}{\mathcal{R}(p^*)} \geq  \min_{0 < f < 1} \frac{c(f-1)e^{c(f-1)}}{f \ln(f) }.$$
\end{restatable}

The proof can be found in Appendix \ref{sec:proofs} and follows from manipulations of log-concave distributions. We note that the condition in Lemma \ref{revenue_case_1}, $p_h \geq c \int_0^{\infty} \bar{F}_V(p) dp$, is satisfied by the optimality conditions for the hinge pricing loss function with equality. While the bound is relatively opaque, it simplifies significantly for lower values of $c$. In this case, the hinge pricing policy receives a constant fraction $c$ of the revenue.

\begin{restatable}{corollary}{lemmaminc}
\label{minimum_at_c}
$  \dfrac{\mathcal{R}(p_h)}{\mathcal{R}(p^*)} \geq \min_{0 < f < 1}  \dfrac{c(f-1)e^{c(f-1)}}{f \ln(f) }= c$ for $0 \leq c \leq 0.5$
\end{restatable}

The proof is in Appendix \ref{sec:proofs}. For $0 \leq c \leq 0.5$, the adversarial distribution is simple and puts a point mass at $p^*$, so $\mathbb{E}[V|X]=p^*$ and the hinge price gets a fraction $c$ of the revenue. This distribution is shown in Figure \ref{flat_dist}. The worst-case distribution is more involved for $c>0.5$.

We prove an expected revenue bound for the case where the prescribed price is greater than the optimal price $p_h \geq p^*$ in Lemma \ref{lemma_hinge_revenue_case_2}.

%\blu This requires differentiability, put in assumption \bla Differentiability is implied by log-concavity?

\begin{restatable}{lemma}{lemmahingeother}
% \begin{lemma}
\label{lemma_hinge_revenue_case_2}
If there exists a pricing rule $p_h$ satisfying optimality conditions from Lemma \ref{lemma_opt_cond_hinge} and $p_h > p^*$, then:
% \begin{equation} \frac{\mathcal{R}(p_h)}{\mathcal{R}(p^*)} \geq   
%     \begin{dcases}
%           \min_{z\leq -2c} \frac{cze^{-z(\frac{1}{c}-1)-1}}{z+c}   & \text{if} ~~  c \geq \frac{p_h  \bar{F}_V(p_h)'}{\bar{F}_V(p_h)(\ln(\bar{F}_V(p_h)) -1) + p_h \bar{F}_V(p_h)'}  \\
%       (c + 1)e^{-c}   &  \text{if} ~~ c \leq \frac{p_h  \bar{F}_V(p_h)'}{\bar{F}_V(p_h)(\ln(\bar{F}_V(p_h)) -1) + p_h \bar{F}_V(p_h)'} 
%     \end{dcases}  
\begin{equation} 
\frac{\mathcal{R}(p_h)}{\mathcal{R}(p^*)} \geq   
          \min \left\{ \min_{z\leq -2c} \frac{cze^{-z(\frac{1}{c}-1)-1}}{z+c}, 
      (c + 1)e^{-c} \right\} 
\end{equation}

% For $0.5\leq c <1$,
%      $$ \argmin_{z\leq -2c}  \dfrac{cz  e^{-z(\frac{1}{c}-1)-1}}{z + c}=- \dfrac{c}{2}\left(\sqrt{\dfrac{c-5}{c-1}} +1 \right)$$
\end{restatable}

%\blu This requires differentiability, put in assumption \bla Differentiability is implied by log-concavity?

An important special case of this occurs when  $c=1$ when the hinge pricing function prices at the conditional expected valuation $p_h=\mathbb{E}[V|X]$. In this case, the bound from Lemma \ref{lemma_hinge_revenue_case_2} simplifies significantly.

\begin{restatable}{corollary}{hingeexpone}
% \begin{corollary}
$  \dfrac{\mathcal{R}(p_h)}{\mathcal{R}(p^*)} = \min_{z\leq -2c} \dfrac{cze^{-z(\frac{1}{c}-1)-1}}{z+c} = e^{-1}$ for $ c = 1$.
\end{restatable}

Lemma \ref{lemma_hinge_revenue_case_2} is a generalization of Lemma 5.4 in \citet{lovasz2007geometry}, who proved a similar result for the case when $c=1$. \blu Furthermore, the case where $c=1$ is also an application of Gr\"unbaum's Theorem \citep{grunbaum1960partitions}. \bla We also note that for $c<1$, a closed-form solution can also be found for the minimizer of the bound.

\begin{restatable}{corollary}{minhingeclosedexpr} 
\label{min_hinge_closed_expr}
For $0<c<1$,
     $$ \argmin_{z\leq -2c}  \dfrac{cz  e^{-z(\frac{1}{c}-1)-1}}{z + c}=- \dfrac{c}{2}\left(\sqrt{\dfrac{c-5}{c-1}} +1 \right)$$
\end{restatable}

  % \begin{lemma}
  % \label{min_hinge_closed_expr} For $c>0.5$,
  %    % $$\dfrac{c(1-c+\sqrt{(c-5)(c-1)})}{2(c-1)}  = \argmin_{z\leq -2c}  \dfrac{cz  e^{-z(\frac{1}{c}-1)-1}}{z + c}$$
  %    $$- \dfrac{c}{2}\left(\sqrt{\dfrac{c-5}{c-1}} +1 \right)= \argmin_{z\leq -2c}  \dfrac{cz  e^{-z(\frac{1}{c}-1)-1}}{z + c}$$
  % \end{lemma}

Theorem \ref{hinge_revenue} follows from applying the optimality conditions in Lemma \ref{lemma_opt_cond_hinge}, and combining the worst-case from bounds in Lemma \ref{revenue_case_1} and \ref{lemma_hinge_revenue_case_2}. These expressions are difficult to evaluate, but it is possible to do so using simulation. It can be shown that for the range $c> 0.792$, $\min_{z\leq -2c} \frac{cze^{-z(\frac{1}{c}-1)-1}}{z+c} \leq  (c + 1)e^{-c} $. Furthermore, for $c<0.823$, $ \min_{0 < f < 1}  \frac{c(f-1)e^{c(f-1)}}{f \ln(f) } \leq \min_{z\leq -2c} \frac{cze^{-z(\frac{1}{c}-1)-1}}{z+c}$ and $\min_{0 < f < 1}  \frac{c(f-1)e^{c(f-1)}}{f \ln(f) } \leq (c + 1)e^{-c}$. As a result, the bound $(c + 1)e^{-c}$ is never the minimum for any value for $c$, leading to its exclusion in the bound presented in Theorem \ref{hinge_revenue}. Furthermore, simulated evaluation of the expressions in Theorem \ref{hinge_revenue} leads to the following characterization of the worst-case revenue:

\begin{corollary}
\label{simulated_hinge_bound}
$\max_{0 \leq c \leq 1} \frac{\mathcal{R}(p_h)}{\mathcal{R}(p^*)} \geq 0.7715$, which occurs at $c^*=0.8234$. 
\end{corollary}

These bounds can be compared to the bounds in \cite{chen2021model}, who study a similar setting. In their single item setting, \cite{chen2021model} study pricing from transaction data with observations of the price of purchases. In the more restricted case of a uniform historical pricing policy, they are able to show that their pricing policy achieves a 0.5 fraction of the optimal revenue, under a similar but not identical assumption on the failure rate of the valuation distribution. The comparison with \cite{chen2021model} is explored in more detail in Section \ref{sec:comp_chen}, as their approach relates closely to the next loss function we propose.

\begin{figure}[]
	\centering
	\includegraphics[width=0.75\textwidth]{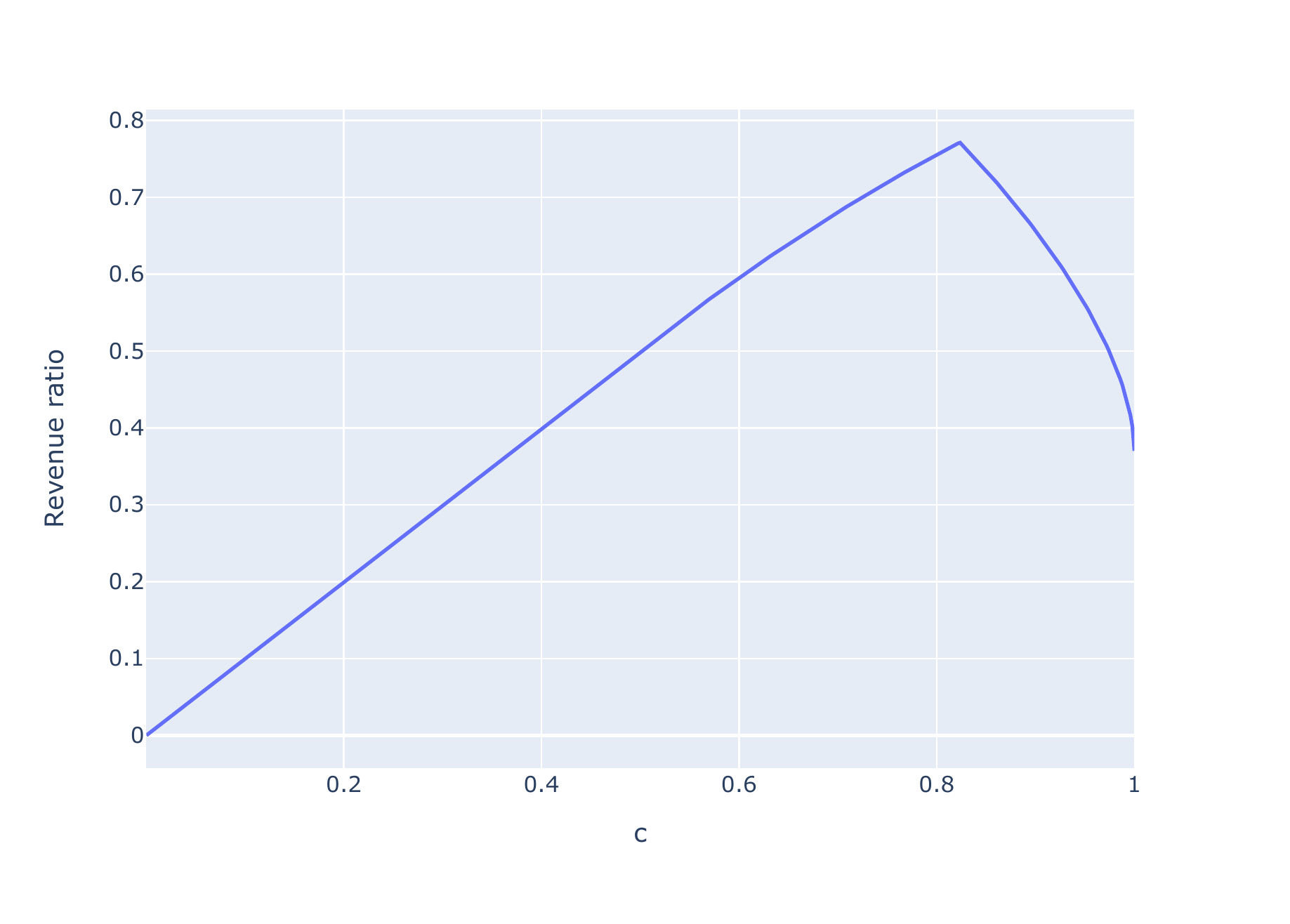}
	\caption{Revenue bounds from Theorem \ref{hinge_revenue} as a function of $c$}
	\label{fig:rev_bounds_hinge}
\end{figure}

This result provides the pricing practitioner with guidance on how to select $c$ in a robust manner by balancing the risk that the valuation distribution is concentrated at high or low values. In particular, this suggests pricing at $c=0.8234$, which is the point at which the two bounds are equal and the adversary is ambivalent about choosing a distribution with an optimal price above or below $p_h$. The revenue bounds as a function of $c$ are shown in Figure \ref{fig:rev_bounds_hinge}. We note that the revenue bounds decrease more rapidly when  $c>0.8234$ rather than if $c<0.8234$, which suggests selecting $c$ above this point might be a riskier choice. We also note that the decrease in revenue for $c<0.8234$ is approximately linear, which is an extension of Corollary \ref{minimum_at_c}. The fact that the optimal parameter $c$ is less than 1 captures the intuition that it is better to price below the customer's expected valuation than above it. Indeed, if the price is above the customer's valuation, there will not be a sale and no revenue will be gained, whereas if the price is below the customer's valuation, there will still be a sale and some revenue gained, albeit less than could optimally be achieved. 

% Since Theorem \ref{quantile_revenue} only has parameters $\tau$, $f$, $z$ it is possible to find the best revenue bound across all valuation distributions with simulation. We note that through simulation it can shown that range $c> 0.792$, $\frac{cze^{-z(\frac{1}{c}-1)-1}}{z+c} \leq  (c + 1)e^{-c} $ .

% We can combine these results to prove Theorem \ref{hinge_revenue}. 
% % refresh these results and add some some colour/insights here.
% % blu is this Grub
% \proof{Proof of Theorem \ref{hinge_revenue}} The Theorem follows from applying the optimality conditions in Lemma \ref{lemma_opt_cond_hinge}, and  combining the bounds in Lemma \ref{revenue_case_1} and \ref{lemma_hinge_revenue_case_2}.  
% \endproof

%This conservative approach is particularly necessary in this bound where the distribution is adversarial. 

%We also note that the seller can use their intuition or knowledge of the problem to guide the choice of $c$ and achieve substantially stronger bounds. For example, in the case where the selling outcomes are deterministic, i.e., when the valuation is constant (or a deterministic function of $X$), then pricing slightly less than the expected valuation is close to optimal. This can be achieved by setting $c$ slightly less than 1, due to the asymmetry of revenue noted above. %More variation in the valuation distribution requires a more conservative choice of $c$. 

 The hinge pricing loss can also be used to adapt the results from \citet{medina2017revenue} to this setting. \citet{medina2017revenue} provide pricing algorithms based on access to an estimate of the valuation, such as that obtained by regressing on valuation samples. Lemma \ref{lemma_opt_cond_hinge} shows that using the hinge pricing loss function, for example using $c=1$, will provide an estimate of the expected value of the valuation distribution that can be used in their algorithm. However, their approach is based on forming clusters of customers with similar valuations and setting a price for each, so it does not result in the interpretable linear policies we desire.

% they also provide upper bounds rather thna lower bounds?

% The hinge pricing loss function and analysis can also be used to provide some insight to the pricing loss function proposed in \citet{ye2018customized}, and described in more detail in Appendix \ref{sec:airbnb_loss}. In particular, when the parameters in \citet{ye2018customized} are set to $c_1= \infty$ and $c_2= -\infty$, and the historic pricing policy is uniform, then the loss proposed in \citet{ye2018customized} is the same as the hinge pricing loss function. In this case, our bounds apply to \citet{ye2018customized}. %When $c_1=c_2=1$, the loss proposed in \citet{ye2018customized} is equal to quantile (median) regression on the historical posted-prices. Note this is different from our proposed quantile loss function which only uses sold items and adapts for the historical pricing policy. Therefore, an intermediate $c_1, c_2$ can be informally viewed as a combination of these two cases.

\section{Quantile pricing loss function}
\label{sec:quantile_loss}

\begin{figure}[]
	\centering
	\subfloat[$Y=1$]{\includegraphics[]{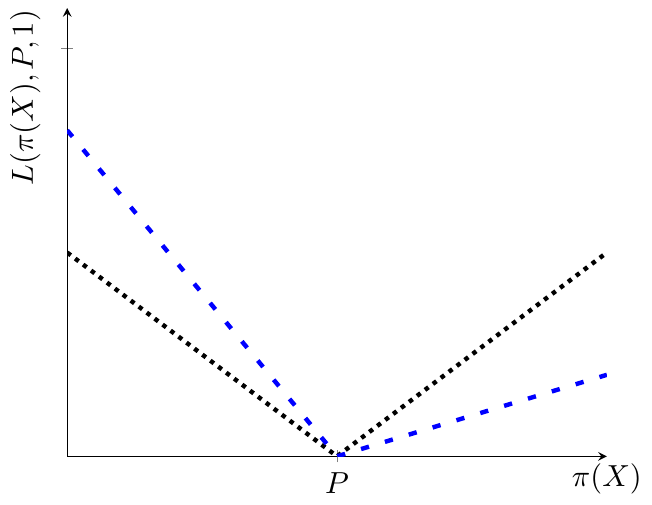}}
	\subfloat[$Y=0$]{\includegraphics[]{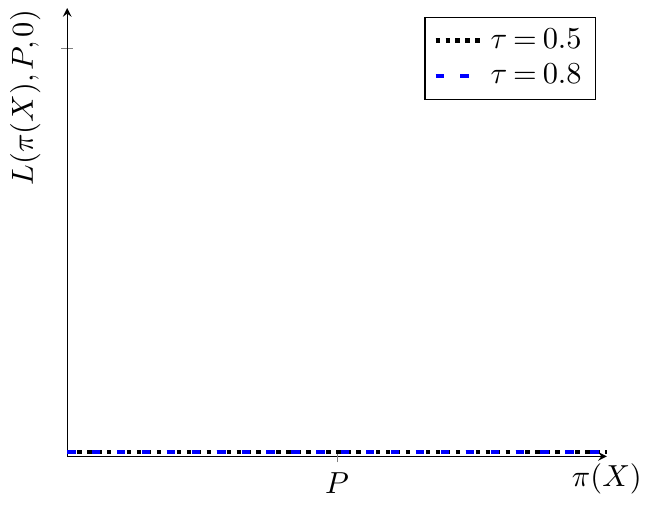}}
\caption{Example of pricing quantile loss function}
\label{fig:quant_loss}
\end{figure}

We also propose an alternative loss function that is able to achieve similar, albeit slightly lower, worst-case pricing bounds but outperforms the pricing hinge loss in some experimental situations. Furthermore, it can be used in settings where the no-purchase decision is not observed, which is often the case with transaction data (e.g., \cite{chen2021model}). The quantile loss is visualized in Figure \ref{fig:quant_loss}.

%While the hinge pricing loss function is intuitive and works well in many practical settings, we also provide an alternative convex loss function which has better expected revenue guarantees:
% \begin{figure}[]
% 	\centering
% 	\includegraphics[width=0.9\textwidth]{fig/quant_loss.pdf}
% 	\caption{Example of pricing quantile loss function}
% 	\label{fig:quant_loss}
% \end{figure}

\begin{definition} The quantile pricing loss function is given by
$$L^q_{\tau}(\pi(X),Y,P)= \frac{Y}{\phi(P|X)}[(1-\tau)(P-\pi(X))^+ + \tau(\pi(X)-P)^+]$$
% $$
% \pi^q_{\tau}(X)=\argmin \mathbb{E}_{Y,P}[L^q_{\tau}(\pi(X),Y,P)|X] 
% $$
\end{definition}

 This can be considered as a weighted quantile loss function for price, using only data for which the product sold. As such, when this loss function is minimized, prices are prescribed at a given quantile of the historical prices of items that sold. The quantile priced at depends on $\tau$, and the strategies for picking a suitable $\tau$ for pricing are discussed later in this section. While to the best of our knowledge, this loss function is not currently used in practice, it does bear resemblance to some commonly used ad hoc contextual pricing strategies, whereby the price is set to be close to similar products that have recently sold. If the product doesn't sell, there is no contribution to the loss. Similar to the hinge pricing loss, the quantile pricing loss is weighted by a weight inversely proportional to the probability of the product receiving each price $\phi(P|X)$, to counteract the imbalance due to the historical pricing policy. 

% A significant advantage of the quantile pricing loss function is that it only requires purchase observations. In particular, this loss function can still be applied to the setting with transaction data where the non-purchase decisions are not observed, as long as the historical pricing policy is still known. This is a commonly studied setting (e.g., \cite{chen2021model}).

% but the quantile  pricing loss function does not price at a conditional quantile of the valuation distribution. Rather it prices at a given fraction of the area under the complementary CDF of the valuation function: 

% $\frac{\int_{p_q}^{\infty} \bar{F}_V(p) dp}{\int_{0}^{\infty} \bar{F}_V(p) dp}= \tau$ 

% This follows immediately from Lemma 6.2. This is in comparison to a traditional quantile loss function on the (unobserved) valuation data, which would give an estimator of the fraction of the area under the pdf:

% $\frac{\int_{p_q}^{\infty} f_V(p) dp}{\int_{0}^{\infty} f_V(p) dp}= \tau$ So  $\bar{F}_V(p_q) = \tau$

\subsection{Comparison with \cite{chen2021model}}
\label{sec:comp_chen}

\begin{figure}[]
	\centering
	\subfloat[$Y=1$, Chen et al.]{\includegraphics[]{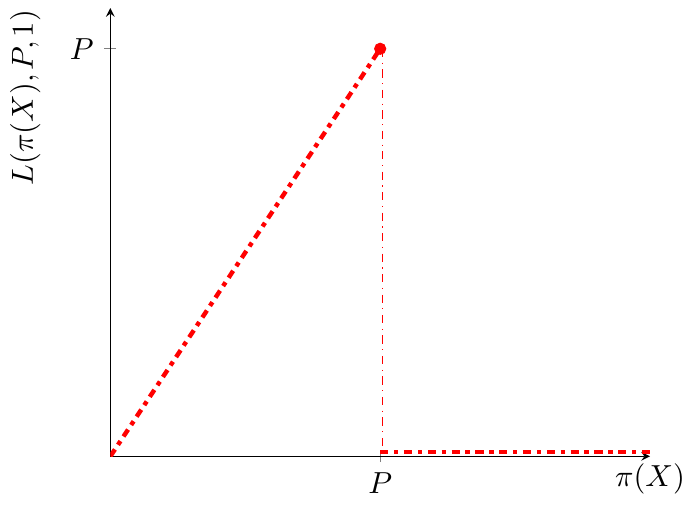}\label{fig:chen_1}}
	\subfloat[$Y=0$, Chen et al.]{\includegraphics[]{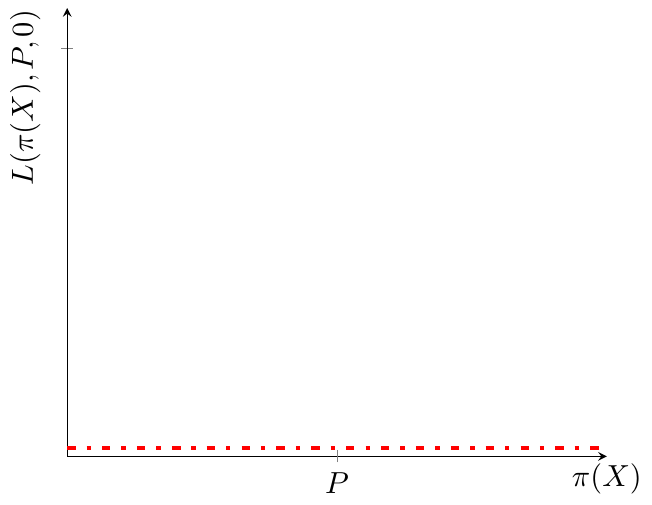}\label{fig:chen_0}}\\
 	\subfloat[$Y=1$, IPS]{\includegraphics[]{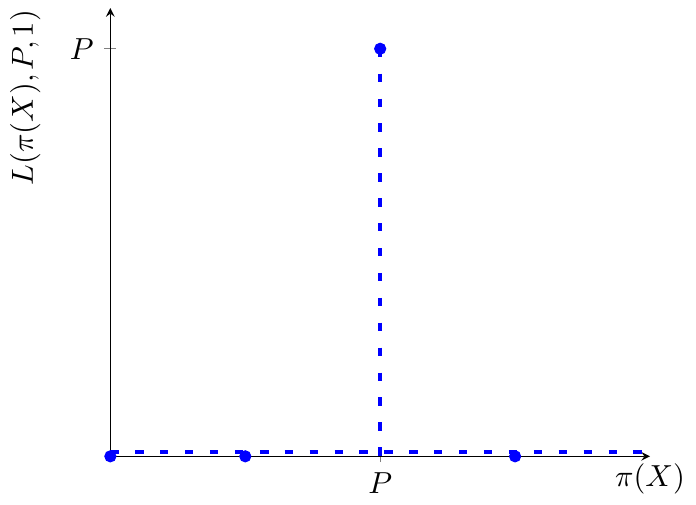}\label{fig:IPS_1}}
	\subfloat[$Y=0$, IPS]{\includegraphics[]{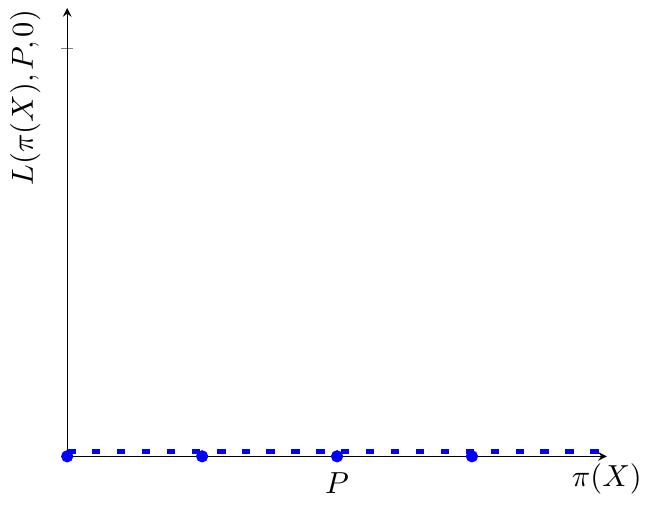}\label{fig:IPS_2}}\\
 	\subfloat[$Y=1$, all transformed]{\includegraphics[]{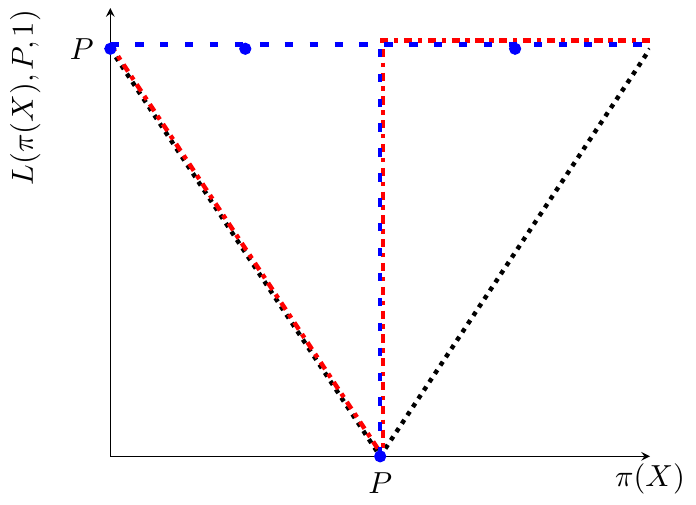}\label{fig:transformed}}
	\subfloat[$Y=0$]{\includegraphics[]{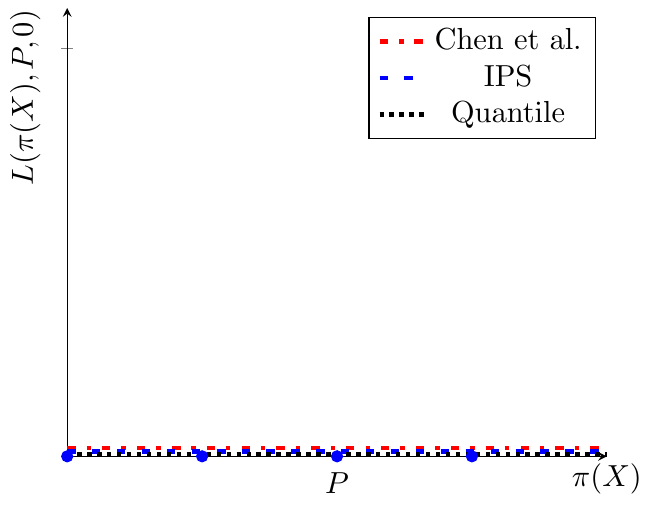}}
\caption{Comparison of \cite{chen2021model}, IPS  and pricing quantile loss function for a uniform historical pricing policy and $\tau=0.5$}
\label{fig:comaprison_with_chen}
\end{figure}

The quantile loss has similarities to the loss function from \cite{chen2021model}. While \cite{chen2021model} focuses on multi-product assortments and does not focus on convexity or contextual information, they study a similar problem setting of pricing from transaction data with observations of the price of purchases. In particular, for the single product setting, they propose maximizing the following function to obtain a pricing policy:\footnote{Since \cite{chen2021model} do not study contextual setting, there is no dependence on $X$ for the pricing policy $\pi(X)$, which we introduce. We also note that Figure \ref{fig:comaprison_with_chen} assumes the historical pricing policy is uniform.}
\begin{equation*}
    L(\pi(X),Y,P) = \pi(X)Y\mathbbm{1}\{P>\pi(X)\}
\end{equation*}

 This is visualized in Figures \ref{fig:chen_1} and \ref{fig:chen_0}. The rationale is that if the customer purchased at price $P$, they will purchase at any price below $P$, but \cite{chen2021model} take a conservative approach and assume there will be no sale above $P$. They study a setting with transaction data where only purchases are observed, but not no-purchase outcomes. As such, they implicitly assume there is no contribution from the no-purchase decisions as shown in Figure \ref{fig:chen_0}.

The loss function from \cite{chen2021model} can be transformed from a maximization to a minimization through multiplication by -1, and translated by a constant $P$, to result in the loss function in Figure \ref{fig:transformed}. Neither of these transformations will affect the minimizer of the function. It is clear that this loss function is non-convex, and therefore potentially difficult to optimize in the contextual setting. The quantile pricing loss function can be thought of as a convex relaxation of this function, in the sense that rather than having a discontinuity at $P$, there is a linear penalty associated with increasing the price above this price, as shown in Figure \ref{fig:transformed}. This maintains the convexity of the loss function. This relaxation does not assume all revenue is lost above $P$, so it can be considered less conservative than \cite{chen2021model}, although this also depends on the choice of $\tau$. In addition, the loss function in \cite{chen2021model} does not adjust for the historical pricing policy, and as such, revenue guarantees are only able to be obtained only for the case of uniform pricing policies.

\subsection{Comparison with Inverse Propensity Scoring}

% \begin{figure}[]
% 	\centering
% 	\includegraphics[width=0.9\textwidth]{fig/IPS_comparison.pdf}
% 	\caption{ IPS comparison with quantile loss function}
% 	\label{fig:comaprison_with_IPS}
% \end{figure}

We also note the similarity between the quantile loss and an Inverse Propensity Scoring (IPS) approach \citep{rosenbaum1983central,beygelzimer2009offset,li2011unbiased}. In a setting where the price is discrete, the IPS estimator is:

$$L_{IPS}(\pi(X),Y,P) = \dfrac{P Y \mathbbm{1}\{P=\pi(X)\}}{\phi(P|X)}$$

Here there is a contribution to the reward proportional to $\frac{PY}{\phi(P|X)}$ only if the proposed price $\pi(X)$ is equal to the price $P$ that was given to the customer in the observed data. Thus the reward obtained from a customer can be considered as a point mass at $\pi(X)=P$. This is shown in Figures \ref{fig:IPS_1} and \ref{fig:IPS_2}. Similarly, this can be transformed by multiplying by -1, and shifting by $\frac{PY}{\phi(P|X)}$ as shown in Figure \ref{fig:transformed}. We note that rather than having a point mass, the quantile reward function has a linear penalty for prices above or below the point $\pi(X)=P$. 

% what about the shift?

%at the $(1-\tau)^{th}$ percentile of the customers' valuations. This is despite not having access to valuation data. Let  $p_q=\argmin \mathbb{E}_{Y,P}[L^q_{\tau}(\pi(X),Y,P)|X]$.

\subsection{Characterization of the pricing policy}

We can characterize the pricing policy that results from minimizing the expected conditional quantile pricing loss function in Lemma \ref{lemma_opt_cond_quant}. This shows that when this loss function is optimized, the price is set such that there is a fixed ratio between the total area under the valuation complementary CDF and the area below the complementary CDF and to the left of the price. 
% his shows that when this loss function is optimized, the price is set such that there is a fixed ratio between the integral of the valuation complementary CDF below the price relative to the total area. 
This is useful for proving revenue bounds, since the total area under the complementary CDF is the maximum expected revenue achievable from a clairvoyant seller who knows the customers' valuations exactly in advance. As we will show, pricing using  the quantile pricing loss function ensures we get a constant fraction of this revenue. If we condition on $X$ and for notational simplicity denote $p_q=\argmin_{\pi(X)} \mathbb{E}_{Y,P}[L^q_{c}(\pi(X),Y,P)|X]$ we can prove Lemma \ref{lemma_opt_cond_quant}.

\begin{restatable}{lemma}{quantoptcond}
\label{lemma_opt_cond_quant} (Quantile optimality condition): Let $p_q$ be the minimizer of the expected quantile pricing loss function, then $$ \frac{\int_{0}^{p_q} \bar{F}_V(p) dp}{\int_{0}^{\infty} \bar{F}_V(p) dp}= 1-\tau $$ % $\tau  \int_{0}^{p_q} \bar{F}_V(p) dp  = (1-\tau)  \int_{p_q}^{\infty} \bar{F}_V(p) dp.$  %or equivalently $\int_{0}^{p_q} \bar{F}_V(p) dp  = (1-\tau) \int_{0}^{\infty} \bar{F}_V(p) dp$.
\end{restatable}

%\blu Remove the second optimality condition and just prove the second when used \bla

\proof{Lemma \ref{lemma_opt_cond_quant}:}
\begin{align*}
    \mathbb{E}_{Y,P}[L^q_{\tau}(p_q,Y,P)|X] =& \int_p  \frac{\mathbb{E}[Y|p,X]\phi(p|X)}{\phi(p|X)}[(1-\tau)(p-p_q)^+ + \tau(p_q-p)^+] dp \\
    =& \int_{p_q}^{\infty} (1-\tau)(p-p_q) \bar{F}_V(p)  dp + \int_{0}^{p_q}  \tau(p_q-p) \bar{F}_V(p) dp 
\end{align*}

Where the second equality follows since $\mathbb{E}[Y|p,X]=\mathbb{P}(V>p|X)$ from Equation \ref{y_def}. Applying the first-order optimality condition, and  the Leibniz integral rule:
%$$ \diffp{}{{\pi(X)}}\mathbb{E}_{Y,P}[L^q_{\tau}(\pi(X),Y,P)|X]=0$$
\begin{align}
\diffp{}{{p_q}}\mathbb{E}_{Y,P}[L^q_{\tau}(p_q,Y,P)|X ] =& \int_{0}^{p_q} \tau \bar{F}_V(p) dp  - \int_{p_q}^{\infty} (1-\tau) \bar{F}_V(p)  dp  =0 \label{eq:quant_opt_cond} \\ 
 \implies & \tau  \int_{0}^{p_q} \bar{F}_V(p) dp  = (1-\tau) \left( \int_{0}^{\infty} \bar{F}_V(p) dp -  \int_{0}^{p_q} \bar{F}_V(p) dp \right)  \\
\implies & \int_{0}^{p_q} \bar{F}_V(p) dp  = (1-\tau) \int_{0}^{\infty} \bar{F}_V(p) dp \Halmos
\end{align}

% It follows that $$\tau = \frac{\int_{p_q}^{\infty} \bar{F}_V(p) dp}{\int_{0}^{\infty} \bar{F}_V(p) dp} \Halmos$$

% Since,
% $$ \int_{0}^{\infty} \bar{F}_V(p) dp =  \int_{0}^{p_q} \bar{F}_V(p) dp + \int_{p_q}^{\infty} \bar{F}_V(p) dp $$
% Then, 
% $$ \int_{L}^{\pi(X)} \tau \bar{F}_V(p) dp  = (1-\tau)[ \int_{L}^{\infty} \bar{F}_V(p) dp -  \int_{L}^{\pi(X)} \bar{F}_V(p) dp] $$
% $$ \int_{L}^{\pi(X)} \bar{F}_V(p) dp  = (1-\tau) \int_{L}^{\infty} \bar{F}_V(p) dp  $$

 \endproof

% \begin{align}
% \tau  \int_{0}^{\pi(X)} \bar{F}_V(p) dp  & = (1-\tau)  \int_{\pi(X)}^{\infty} \bar{F}_V(p)  dp  \label{eq:opt_cond_1}\\
% \tau  \int_{0}^{\pi(X)} \bar{F}_V(p) dp & = (1-\tau) \Big[ \int_{0}^{\infty} \bar{F}_V(p) dp -  \int_{0}^{\pi(X)} \bar{F}_V(p) dp \Big] \\
%  \int_{0}^{\pi(X)} \bar{F}_V(p) dp  &= (1-\tau) \int_{0}^{\infty} \bar{F}_V(p) dp \label{eq:opt_cond}  
% \end{align} 

It is also important to differentiate this from the standard quantile loss function applied to the (unobserved) valuation data.  If we applied quantile regression here, we would get an estimator of the fraction of the area under the valuation PDF, $\frac{\int_{0}^{p_q} f_V(p) dp}{\int_{0}^{\infty} f_V(p) dp}= \tau$, resulting in a policy where $\bar{{F}}_V(p_q) = 1 - \tau$. As a result, this pricing policy is clearly different. %Pricing at a quantile of the valuation function is proposed by \citet{allouah2021revenue} in the setting with a limited number of valuation samples and no contextual data.

%Using this lemma it is straightforward to show this corresponds to pricing at the $(1-\tau)^{th}$ percentile of the customers' valuations.
%Therefore, if the data is collected using a randomized controlled trial (with equal likelihood of each price being offered), minimizing this loss function is the same as pricing at the $(1-\tau)^{th}$ percentile of items which sold.

\subsection{Expected revenue bounds}
\label{sec:quantile_loss_bounds}

When we optimize the quantile pricing loss function, we can find bounds on the optimal revenue, relative to the best contextual pricing policy that has access to the valuation distribution.  
   % see if we can find a citation for that.

% \blu Say something about this being essentially optimal if we set $\tau$ to be small in the case where there is very little uncertainty in the the outcomes. \bla

% \begin{theorem}
% \label{quantile_revenue}
% \begin{equation}
% \frac{\mathcal{R}(\pi^q_{\tau}(X))}{\mathcal{R}(\pi^*(X))} \geq \min \left\{4(1-\tau)\tau,~ \min_{0 < f < 1}  \frac{(1-\tau)(f-1)e^{(1-\tau)(f-1)}}{f \ln(f) }  \right\}
% \end{equation}
% \end{theorem}

% \begin{restateable}{theorem}{theoremquant}
\begin{theorem}
\label{quantile_revenue}
\begin{equation*}
\frac{\mathcal{R}(p_q)}{\mathcal{R}(p^*)} \geq \min \left\{ \min_{\tau \leq z \leq 1} \frac{z\tau(\ln(z)+1)-z^2} {\tau-z}  , \min_{0 < f < 1}  \frac{(1-\tau)(f-1)e^{(1-\tau)(f-1)}}{f \ln(f) }  \right\}
\end{equation*} 
\end{theorem}

% \end{restateable}

Similar to the hinge pricing loss function, the proof for Theorem \ref{quantile_revenue} requires exploring two cases, one where the price chosen by minimizing the quantile pricing loss is above the optimal price and one where it is below.  We start by bounding the case where the price obtained by minimizing the quantile pricing loss function is greater than the optimal price.
 
\begin{restatable}{lemma}{quantgreater}
\label{lemma_quantile_revenue_case_2}
If there exists a pricing rule $p_q$ satisfying optimality conditions from Lemma \ref{lemma_opt_cond_quant} and $p_q > p^*$, then:
$$ \frac{\mathcal{R}(p_q)}{\mathcal{R}(p^*)} \geq  \min_{\tau \leq z \leq 1} \frac{z\tau(\ln(z)+1)-z^2} {\tau-z}.$$
\end{restatable}

This again follows nontrivially from the properties of log-concavity and is proven in Appendix \ref{sec:proofs}. To help parse this bound, it can be lower bounded by the following function with only a minor loss of accuracy.

\begin{restatable}{corollary}{colllowerbound}
 $$  \min_{\tau \leq z \leq 1} \frac{z\tau(\ln(z)+1)-z^2} {\tau-z} \geq 4(1-\tau)\tau$$
\end{restatable}

This bound helps understand the relationship with the parameter $\tau$ and clearly shows that increasing $\tau$ from 0 toward 0.5 improves the bound. Furthermore, it can be observed from Figure \ref{fig:rev_bounds_quantile}, which visualizes the revenue bound, this bound is well approximated by a quadratic. 

For the case where $p_q < p^*$, we can apply the bound from Lemma \ref{revenue_case_1} again. Lemma \ref{revenue_case_1} shows that a constant fraction of revenue can be achieved under a condition where the price prescribed is at least a fixed constant of the revenue achievable by a clairvoyant, $p_h \geq c \int_0^{\infty} \bar{F}_V(p) dp$. We show that the optimality condition for the quantile loss function implies this condition. %This condition is also useful for another loss function we introduce in the next section.

 \begin{restatable}{corollary}{collquantile}
 \label{corollary_quantile}
 If there exists a pricing rule $p_q$ satisfying optimality conditions from Lemma \ref{lemma_opt_cond_quant} and $p_q \leq p^*$
 $$  \frac{\mathcal{R}(p_q)}{\mathcal{R}(p^*)}  \geq \min_{0 < f < 1}  \frac{(1-\tau)(f-1)e^{(1-\tau)(f-1)}}{f \ln(f) }  $$
\end{restatable}

 \proof{Proof of Corollary \ref{corollary_quantile}:} From the quantile pricing optimality condition in Lemma \ref{lemma_opt_cond_quant},
\begin{equation*}
    \int_{0}^{p_q} \bar{F}_V(p) dp  = (1-\tau) \int_{0}^{\infty} \bar{F}_V(p) dp
\end{equation*}
  Since $p_q\geq \int_{0}^{p_q} \bar{F}_V(p) dp $, the conditions for Lemma \ref{revenue_case_1}
 are satisfied for the constant $1-\tau$.\Halmos 
\endproof

As a result, a version of Corollary \ref{minimum_at_c} also holds for the quantile pricing loss function, whereby the revenue ratio can be very simply lower bounded by $1-\tau$ for the range $0.5 \leq \tau <1$. As shown in Section \ref{sec:tightness_of_bound}, this bound is actually tight.

\begin{corollary}
$  \frac{\mathcal{R}(p_q)}{\mathcal{R}(p^*)}  \geq \min_{0 < f < 1}  \frac{(1-\tau)(f-1)e^{(1-\tau)(f-1)}}{f \ln(f) } = 1- \tau $ for $0.5 \leq \tau <1$
\end{corollary}

Theorem \ref{quantile_revenue} follows from combining the bounds in Lemma \ref{lemma_quantile_revenue_case_2} and Corollary \ref{corollary_quantile}.
Similar to the hinge pricing loss, the bound from Theorem \ref{quantile_revenue} is difficult to simplify further. However, since Theorem \ref{quantile_revenue} only has parameters $\tau$, $f$, $z$, it is possible to find the best revenue bound across all valuation distributions with simulation:

\begin{restatable}{corollary}{quadquantbound}
\label{simulated_quantile_bound}
$\max_{0 \leq \tau \leq 1} \frac{\mathcal{R}(p_q)}{\mathcal{R}(p^*)} \geq 0.749$, which occurs at $\tau^*=0.209$. 
\end{restatable}

  % To the best of our knowledge, these are the first bounds for the contextual posted-price setting we study. 

This gives practical guidance on how to choose $\tau$. In particular, $\tau$ with the best worst-case guarantee corresponds to setting the price to approximately the $79^{th}$ percentile of the items that sold. The complete relationship between the revenue bounds and $\tau$ is shown in Figure \ref{fig:rev_bounds_quantile}, which gives practitioners an idea of the worst-case revenue for different choices of $\tau$.

While this bound is marginally weaker than the bound for the hinge pricing loss function, it is applicable in the setting where only sale outcomes are observed. In this respect, the bound for the quantile pricing loss is perhaps more directly comparable with the 0.5  bound from \cite{chen2021model} than the hinge pricing loss.

%   \blu I don't think any of this below is really relevant\bla 

% In the proof for Theorem \ref{quantile_revenue}, there is a balance between setting the quantile too low, in which case an adversary can choose a distribution that has a high probability of selling at high prices, versus setting the quantile too high in which case the adversary can make the selling probability low at prices higher than the optimal. However, how quickly the selling probability can decrease with the price is limited by the log-concave distribution. 

% The result from Corollary \ref{simulated_quantile_bound} provides practical guidance to the practitioner on how to manage this balance in a robust manner. Similar to the hinge pricing loss function, if the valuation is deterministic (a point mass), a suitable choice of parameter for $\tau$ is close to zero, corresponding to pricing at close to the $100^{th}$ percentile of tickets sold. %We note that although the quantile pricing loss function has a better worst-case bound than the hinge pricing loss function, as we will see in Section \ref{sec:numerical_exp}, this doesn't always yield the highest revenue. 

\section{Tightness of bounds}
\label{sec:tightness_of_bound}

\begin{figure}[]
	\centering
	\subfloat[$p_h \leq p^*, ~c=0.8$]{\includegraphics[]{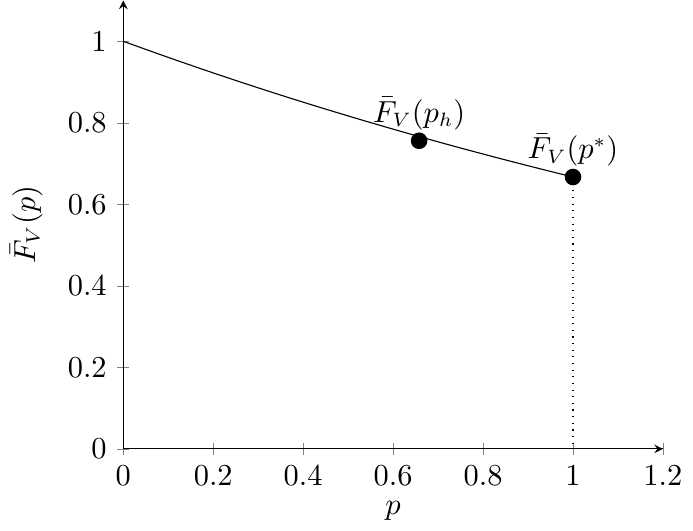}\label{hinge_flat_dist}}
	\subfloat[$ p_h > p^*, ~c=0.8$]{\includegraphics[]{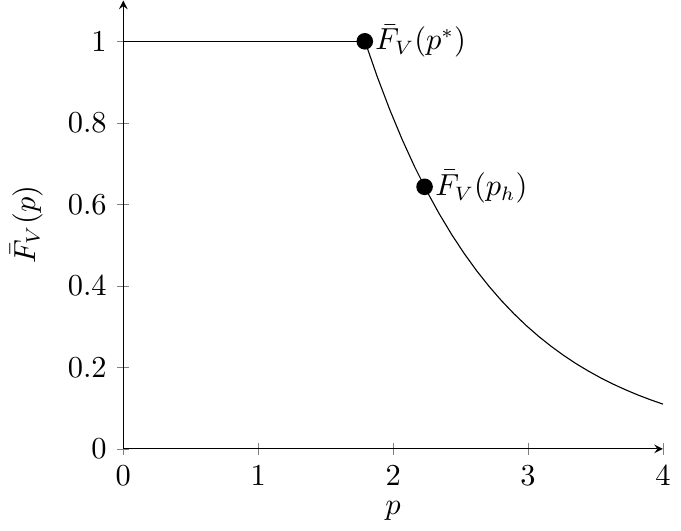}\label{hinge_exp_dist}}\\
 	\subfloat[$ p_q \leq p^*, ~\tau=0.5$]{\includegraphics[]{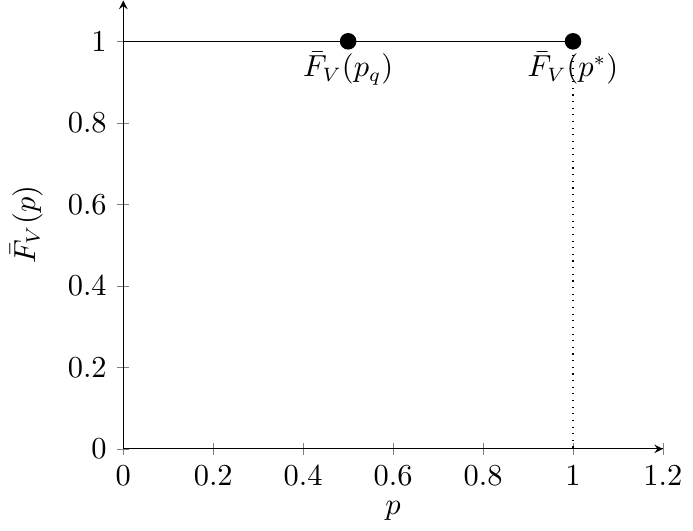}\label{flat_dist}}
	\subfloat[$ p_q > p^*, ~\tau=0.2$]{\includegraphics[]{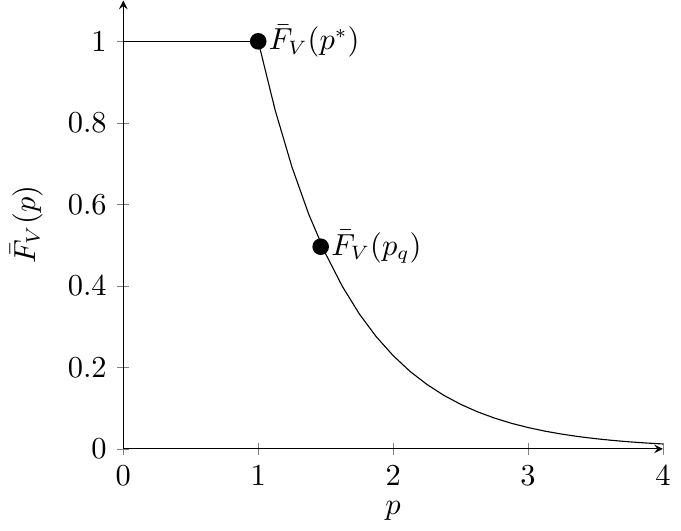}\label{quant_exp_dist}}
\caption{Worst-case valuation distributions from Lemma \ref{lemma_upper_bound}}
\label{fig:worst_case_dist}
\end{figure}

In this section, we examine the tightness of the bounds from Sections \ref{sec:exp_rev_bounds} and \ref{sec:quantile_loss_bounds}. This is achieved by finding valuation distributions that obtain the revenue given by the bounds. Since the worst-case distribution is different for different pricing rules, these distributions are parameterized by $c$ and $\tau$ respectively. Figure \ref{fig:worst_case_dist} illustrates the worst-case distributions.

\begin{restatable}{lemma}{upperbounds}
\label{lemma_upper_bound}
\begin{enumerate}  
\item For the hinge loss function and the case where $p_h \leq p^*$, let \begin{equation*}
\bar{F}_V(p)=  \begin{cases}
      e^{p\ln(g(c))}   &  \text{for} ~~ 0 \leq p \leq 1 \\ 
      0   & \text{for} ~~   p > 1  \\
    \end{cases} 
\end{equation*}
 where $g(c)=\argmin_{0 < f < 1} \dfrac{c(f-1)e^{c(f-1)}}{f \ln(f) }$. If $g(c) \geq e^{-1} $, $  \dfrac{\mathcal{R}(p_h)}{\mathcal{R}(p^*)} = \min_{0 < f < 1} \dfrac{c(f-1)e^{c(f-1)}}{f \ln(f) }$, i.e., Lemma \ref{revenue_case_1} is tight. Furthermore, $g(c)=1$ for $0<c<0.5$, which simplifies the complementary CDF to a simple step function.
\item For the hinge loss function and the case where $p_h \geq p^*$, let \begin{equation*}
\bar{F}_V(p)=  \begin{cases}
      1   &  \text{for} ~~ 0 \leq p \leq t(c) \\ 
      e^{t(c)-p}   & \text{for} ~~   p >  t(c)  \\ 
    \end{cases} 
\end{equation*}
where $t(c)= \dfrac{1}{2}\left(\sqrt{\dfrac{c-5}{c-1}} - 1 \right)$.  If $c \geq % 0.792$,
0.5 ,  \dfrac{\mathcal{R}(p_h)}{\mathcal{R}(p^*)} = \min_{z\leq -2c}  \dfrac{cz  e^{-z(\frac{1}{c}-1)-1}}{z + c} $, i.e., Lemma \ref{lemma_hinge_revenue_case_2} is tight. 
% $\bar{F}_V(p)=1$, for $p \leq M,~  \bar{F}_V(p)=e^{c(M-p)}$ for $ p > M $, then $  \lim_{M \rightarrow \infty} \frac{\mathcal{R}(p_h)}{\mathcal{R}(p^*)} = e^{-c}$
\item For the quantile loss function and the case where $p_q \leq p^*$, let \begin{equation*}
\bar{F}_V(p)=  \begin{cases}
      1  &  \text{for} ~~  0 \leq p \leq 1 \\ 
      0   & \text{for} ~~   p > 1  \\
    \end{cases} 
\end{equation*}
Then, $\dfrac{\mathcal{R}(p_q)}{\mathcal{R}(p^*)} = 1-\tau = \min_{0 < f < 1}  \dfrac{(1-\tau)(f-1)e^{(1-\tau)(f-1)}}{f \ln(f) } $ for $\tau \geq 0.5$.  i.e Lemma \ref{revenue_case_1} is tight in this range.
\item For the quantile loss function and the case where $p_q \geq p^*$, let \begin{equation*}
\bar{F}_V(p)=  \begin{cases}
      1  &  \text{for} ~~  0 \leq p \leq 1 \\ 
      e^{(p-1)(1-\frac{z(\tau)}{\tau})}   & \text{for} ~~   p > 1  \\
    \end{cases} 
\end{equation*}
where $z(\tau)=\argmin_{\tau \leq z \leq 1} \dfrac{z\tau(\ln(z)+1)-z^2} {\tau-z}$. If $\tau \leq 0.5 $, $  \dfrac{\mathcal{R}(p_h)}{\mathcal{R}(p^*)} = \min_{\tau \leq z \leq 1} \dfrac{z\tau(\ln(z)+1)-z^2} {\tau-z}$, i.e., Lemma \ref{lemma_quantile_revenue_case_2} is tight. 
\end{enumerate}
\end{restatable}

\begin{figure}[]
	\centering
	\includegraphics[width=0.8\textwidth]{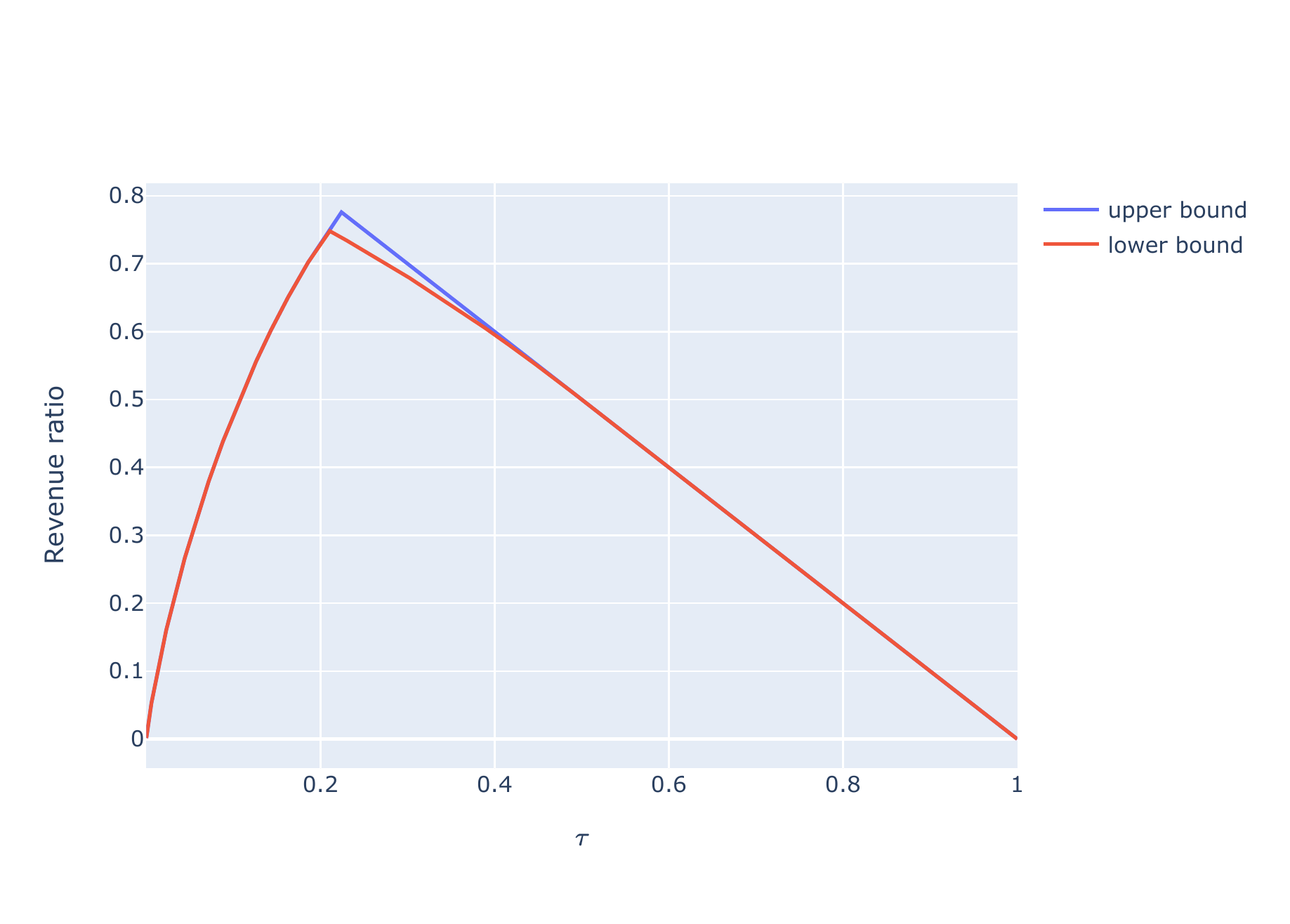}
	\caption{Revenue bounds from Theorem \ref{quantile_revenue} as a function of $\tau$ with upper bound from Lemma \ref{lemma_upper_bound}}
	\label{fig:rev_bounds_quantile}
\end{figure}

As a corollary of Lemma \ref{lemma_upper_bound}, the bounds for the hinge pricing loss function given in Theorem \ref{hinge_revenue} are tight, while those for the quantile pricing loss function given in Theorem \ref{quantile_revenue} are very close to being tight. This occurs because the revenue bounds from Lemma \ref{revenue_case_1} are satisfied by the distribution from Figure \ref{flat_dist} for the range $\tau  \geq 0.5$, but not for $\tau < 0.5$. As a result, it is unclear whether there exists an alternative valuation distribution that satisfies the bound for the range $0.209 < \tau < 0.5$, or whether it is possible to improve the bound for this range. The relative difference between the upper bound given by the distribution in Figure \ref{flat_dist} and the lower bound from Lemma \ref{revenue_case_1} is shown in Figure \ref{fig:rev_bounds_quantile}. We note that the difference is at most $0.027$, so the bound is very close to being tight. More practically, there is minimal effect on the optimal choice for a robust $\tau$. For the hinge pricing loss, there are no such issues, as valuation distributions can be found that satisfy bounds from Lemma \ref{revenue_case_1} and \ref{lemma_hinge_revenue_case_2}  and cover the range of $c$.

The worst-case distributions from Figure \ref{fig:rev_bounds_quantile} also provide practical insight into the difficulties in choosing a parameter $c$ or $\tau$. If $c$ is chosen to be small so that the price $p_h$ is a small fraction of the expected valuation, then the worst-case valuation distribution is similar to Figure \ref{hinge_flat_dist} but perhaps more easily understood from Figure \ref{flat_dist} (when c<0.5, this is the worst case bound for the hinge loss too). In this case, most customers have a valuation close to the expected value, which is approximately equal to the optimal price $p^*$. This results in a complementary CDF that is relatively flat (most customers will purchase at prices below the optimal price), followed by a sharp drop as no customers will purchase above the optimal price. From this example, it is clear why the hinge pricing loss gets close to a $c$ fraction of optimal revenue. However, as $c$ increases, the worst-case distribution changes to one where $p^* \leq p_h$. In this alternative regime where $c$ is larger, the worst-case valuation distribution is similar to Figure \ref{hinge_exp_dist} wherein the selling probability rapidly (exponentially) declines after the optimal price. However, since the $p_h$ is a fraction $c$ of the expected valuation, there must be some customers with a higher valuation. Combined with the log-concavity assumption, which limits how quickly the selling probability changes, this prevents $\bar{F}_V(p_h)$ from being too low, so it is still possible to achieve a fraction of the optimal revenue. The behavior for the quantile pricing loss is similar and shown in Figures \ref{flat_dist} and \ref{quant_exp_dist}.

% If a lower value of $c$ is chosen, (i.e much below the expected valuation), the adversarial valuation distribution will have $p_h < p^*$, and be characterized by many high-valued customers to obtain revenue from which the lower price $p_h$ doesn't fully exploit. Conversely, if a higher value of $c$ is chosen, the adversarial valuation distribution will have $p_h > p^*$, but be characterized by a long tail where the selling probability at $p_h$ is relatively low, relative to a significantly higher selling probability at the slightly lower price $p^*$.

%0.7759
% In particular, for the case when $p_q \geq p^*$, the bound from Lemma \ref{lemma_quantile_revenue_case_2} is tight, but for the case when $p_q \leq p^*$, there is potentially a small gap between the lower bound from Lemma \ref{revenue_quantile_case_1} and the empirical worst-case distribution.

\section{Cross-validation for parameter selection}
\label{sec:xval}

The parameters $c$ and $\tau$ can be chosen to robustly maximize revenue across all valuation distributions, as per Corollary \ref{simulated_hinge_bound} or \ref{simulated_quantile_bound}. However, this is often too conservative in practice.  We can utilize the pricing practitioner's partial knowledge of the valuation distribution or demand to set these parameters. We present a simple heuristic for choosing $c$ and $\tau$ for the specific valuation distribution encountered, based on ideas from cross-validation. To evaluate the effectiveness of the model parameters, we propose using a demand model estimated from the available data. Specifically, for each parameter value $\tau_k$ in a discretized set of size $K$, we find an optimal policy $\hat{\pi}_{\tau_k}$ by optimizing the corresponding hinge or quantile loss function. We then estimate the revenue obtained from this pricing policy using the estimated demand model and pick the parameter that corresponds to the highest estimated revenue. This approach allows us to utilize an accurate nonparametric demand model (such as a tree ensemble or neural network), without being concerned about the difficulty of optimizing such a non-convex model.
% \begin{algorithm}[!htbp]
% \caption{Cross validation for contextual pricing}
% \label{alg:cst}
% \begin{algorithmic}[1]
%     % \State Train an initial $f_0(x,p;\theta)$ on observational data.  \algorithmiccomment{Classifier initialization}
%     \State Estimate demand model $\hat{f}(X,P)$ from data
%     \For{each $k \in \{1,...,N\}$}
%         \State Find optimal policy using parameter $\tau_k,~\hat{\pi}_k = \argmin_{\pi \in \Pi} \frac{1}{n}\sum_{i=1}^n \argmin L_{\tau_{k}}(\pi(X_i),Y_i,P_i) $
%         \State Evaluate estimated reward $\hat{\mathcal{R}_k}=\frac{1}{n} \sum_{i=1}^n \hat{\pi}_k(X_i)\hat{f}(X_i,\hat{\pi}_k(X_i))$
%     \EndFor
%     \State Find maximum reward $k^*= \argmax_k \hat{\mathcal{R}_k}$, choose $\pi_{k^*}$ as policy
% \end{algorithmic}
% \end{algorithm}
\begin{algorithm}
\caption{Cross-validation for contextual pricing}\label{alg:cst}
Estimate demand model $\hat{f}(X,P)$ from data \;
\For{$k \in \{1,...,K\}$}
{
Find the optimal policy using parameter $\tau_k,~\hat{\pi}_k = \argmin_{\pi \in \Pi} \frac{1}{n}\sum_{i=1}^n L_{\tau_{k}}(\pi(X_i),Y_i,P_i) $\;
Evaluate estimated reward $\hat{\mathcal{R}_k}=\frac{1}{n} \sum_{i=1}^n \hat{\pi}_k(X_i)\hat{f}(X_i,\hat{\pi}_k(X_i))$\;
}
Find maximum reward $k^*= \argmax_k \hat{\mathcal{R}_k}$, choose $\pi_{k^*}$ as policy\;
\end{algorithm}
% We also show that the log-concave assumption is necessary. Without any regularity on the distribution of revenue, the revenue bounds are arbitrarily bad. For example consider a distribution where.
% Furthermore, the bounds are tight. Consider the case where the valuation distribution is
%We note, this is different from cross-validation in the 
\section{Linear policies and generalization bounds}
\label{sec:generalization_bounds}
An important class of pricing policies is the class of linear functions, where $\pi_{\theta}(X)=  \langle \theta, X \rangle $. This class of functions is valued for its simplicity and interpretability and is well-studied in the pricing literature  \cite[e.g.,][]{besbes2015surprising}.  Also, when using a linear policy class, the empirical risk minimization problem of finding the best policy by optimizing the hinge and quantile pricing loss functions is convex. This is not necessarily true for nonlinear policy classes. %Furthermore, because of its simplicity, it is less likely to overfit the data, allowing generalization bounds to be established.

The bounds introduced in previous sections apply when optimizing the expected loss function which is often not possible in practice. It is of interest to study how the solution will generalize when it is optimized using a finite sample of data to form generalization bounds. To achieve these bounds, we use regularized loss minimization, in which we jointly minimize the empirical risk and a regularization function that penalizes large values of $\theta$. 
\begin{equation*}
     \hat{\theta}_{\lambda} = \argmin_{||\theta||_2\leq B } \frac{1}{n} \sum_{i=1}^n L(\langle \theta, X_i \rangle,Y_i,P_i) + \lambda ||\theta||_2^2
\end{equation*}

Generalization bounds are well studied for regularized convex loss functions that are $\rho$-Lipshitz  \cite[e.g.,][]{shalev2014understanding}. In Lemma \ref{lemma_lipshitz} we show that the quantile and hinge pricing loss functions are $\rho$-Lipshitz. For these results, we will assume that the price range is bounded, $p \in [a,b]$, so that the historical probability of assigning a price $\phi(p|X)$ can be bounded away from 0 by a constant.

\begin{lemma}
\label{lemma_lipshitz} 
Assume there exists $d \leq \phi(p|X)$ for all $X \in \mathcal{X}$, $p \in [a,b]$. Then $L^q_{\tau}(\cdot ,Y,P)$ and $L^h_{\tau}(\cdot ,Y,P)$ are $\rho$-Lipshitz, with $\rho= \max\{\tau,1-\tau\}/ d$.
\end{lemma}

This simple result follows from taking the maximum absolute gradient of each loss function. This allows us to apply known generalization bounds to this setting. First, denote $\theta^*=\argmin_{\theta} \mathbb{E}[L(\langle \theta, X \rangle,Y,P)]$.

\begin{lemma}
\label{lemma_generalization_bound}
For either $L = L^h_{\tau}(\cdot ,Y,P)$ or $L =L^q_{\tau}(\cdot ,Y,P)$, let $\lambda= \sqrt{\frac{2\max\{\tau,1-\tau\}^2}{d^2B^2n}}$. For any training set of size $n$,
$$  \mathbb{E} [L( \langle \hat{\theta}_{\lambda}, X \rangle ,Y,P)] \leq \mathbb{E} [L(\langle \theta^*, X \rangle,Y,P)] + \frac{B\max\{\tau,1-\tau\}}{d} \sqrt{\frac{8}{n}}.$$
In particular, if $\epsilon > 0$, and $n \geq \frac{8 (\max\{\tau,1-\tau\})^2 B^2}{d^2 \epsilon^2}$, then $ \mathbb{E} [L( \langle \hat{\theta}_{\lambda}, X \rangle ,Y,P)] \leq \mathbb{E} [L(\langle \theta^*, X \rangle,Y,P)] + \epsilon $.
\end{lemma}

The result follows from Corollary 13.9 in \citet{shalev2014understanding}. An advantage of this bound is that it doesn't explicitly depend on the dimension of the covariate space $\mathcal{X}$, although there is a dependence on $B$ since we require $||\theta_{\lambda}||_2\leq B $. Therefore, it is still possible to get good bounds when the covariate space is high-dimensional yet sparse. This bound suggests that a large sample size $n$ is needed for small $d$, which occurs when there are some prices that have a very low probability of being assigned $\phi(p|X)$. This happens when the pricing policy is unbalanced. This unbalanced data issue is well known in the off-policy learning community and affects inverse propensity weighting methods because to account for observations at uncommon prices, large weights must be used, leading to high variance estimates. Approaches including normalization via re-weighting \citep{lunceford2004stratification, austin2015moving} and trimming of the weights to reduce the variance of the estimates \citep{elliott2008model,ionides2008truncated} have been shown to help mitigate this problem. It is also possible to apply generalization bounds to nonlinear function classes, although this may result in a non-convex revenue optimization problem, which may be challenging to solve.

% \blu Should we also mention that nice thing about linear functions is that there is no dependence on $m$ - see shalev swartz  \bla

% \blu Maybe need to be careful with the constant, put in some constants? \bla

\section{Numerical experiments}
\label{sec:numerical_exp}

We test the proposed loss functions using synthetic and real-world datasets. We benchmark against commonly used direct method approaches that first estimate demand using logistic regression \cite[\texttt{dm\_log}, e.g.,][]{chen2015statistical} and a tree ensemble model  \cite[\texttt{dm\_lgbm}, e.g.,][]{mivsic2017optimization}, then optimize estimated revenue to find a pricing policy. In particular, we adopt the logistic regression model implemented in \texttt{sci-kit learn} \citep{scikit-learn}, and the \texttt{lightgbm} gradient boosted tree package \citep{ke2017lightgbm} with default parameters in each case. We also benchmark against an Inverse Propensity Weighting approach (\texttt{kern\_ipw}) adapted to the continuous action setting \citep{kallus2018policy}, whereby revenue is estimated using a weighted average according to how far the historical price is from the  proposed policy as evaluated by a kernel. We use a Gaussian kernel with a 0.2 bandwidth, a value that worked well in our setting. We benchmark against the model-free assortment approach from Section \ref{sec:comp_chen} \cite[][\texttt{model\_free}]{chen2021model} and the $\epsilon$-insensitive regression approach from Section \ref{sec:airbnb_loss} \cite[][\texttt{eps\_insensitive}]{ye2018customized}. The hinge and quantile pricing loss functions (denoted \texttt{hinge} and \texttt{quant} respectively) are optimized using the cross-validation technique from Section \ref{sec:xval} to find the parameters ($\tau$ and $c$), with 10 rounds of cross-validation. We use the \texttt{lightgbm} model to evaluate the reward in this procedure. Likewise, to find the parameters $c_1$ and $c_2$ for \texttt{eps\_insensitive}, we use the same cross-validation procedure. In all experiments, we restrict the pricing policy to be a linear function of the covariates with no intercept term. In all algorithms, the optimal policy is found using the popular BFGS nonlinear optimization algorithm \citep{nocedal2006numerical}, implemented in \texttt{scipy} \citep{2020SciPy-NMeth}. This can  handle non-differentiable demand functions such as tree-based ensemble methods.

% \blu  should probably put in the IPW formula here, also change exp to hinge? Maybe also explain DM\bla

\subsection{Synthetic data}
\label{sec:synth_data}

Synthetic experiments are important in this setting due to the lack of publicly available pricing data with counterfactual outcomes on whether a customer would have purchased if a different price had been offered. As a result, estimating the revenue of different pricing policies from historical data is challenging. However, with synthetic data, the underlying probability distributions that govern customer behavior are known, so pricing policies can be evaluated.

In our synthetic experiments, we propose a number of different data-generating scenarios to test the algorithms under varied demand settings. In one set of experiments, the valuation distribution is uniformly distributed. The average valuation depends on the customer features through a function $g(X)$, such that $V \sim \text{Uniform}(g(X),g(X)+3)$, $X \sim \text{Uniform}(1,2)^{2}$, and where the historical pricing policy is uniform $P \sim \text{Uniform}(1,3)$. We study a \textit{linear} function $g(X)=\frac{1}{2} (X_{1,i}+X_{2,i})$
%$g(X)=\frac{1}{2} \sum_{i=1}^2 X_{ij} $, 
and a \textit{step} function, $g(X)=\frac{1}{2} \sum_{i=1}^2 \left(\mathbbm{1}\{X_{1j} \geq  \bar{X}_1 \}  + \mathbbm{1}\{X_{2j} \geq  \bar{X}_2 \} \right)$,
%$g(X)=\frac{1}{2} \sum_{i=1}^2 \mathbbm{1}\{X_{ij} \geq  \bar{X}_i \} $, 
where $\bar{X}_i=\frac{1}{n}\sum_{j=1}^n X_{ij}$. As described in Section \ref{sec:model}, we do not observe $V$, but only $Y$ as generated by Equation \ref{y_def}, depending on whether the valuation is greater than the price offered.
In another set of experiments, we generate valuations according to a shifted exponential distribution $V \sim \text{Exp}(g(X),\text{loc}= 5)$, with $X \sim \text{Uniform}(1,5)^{2}, P \sim \text{Uniform}(0,15)$, also with the linear and step dependence on $X$. %For each distribution, we study a number of different dependencies on $X$, both linear and nonlinear , including a \textit{linear} function $g(X)=\frac{1}{m} \sum_{i=1}^m X_{ij} $, a \textit{step} function, $g(X)=\frac{1}{m} \sum_{i=1}^m \mathbbm{1}\{X_{ij} \geq  \bar{X}_i \} $, where $\bar{X}_i=\frac{1}{n}\sum_{j=1}^n X_{ij}$ is the average value for each dimension, and an (inverted) \textit{absolute value} function $g(X)=\frac{1}{m}  (X_{i}^{(1)} - \sum_{i=1}^m  |X_{ij} - \bar{X}_i|)$, where $X_{i}^{(1)}=\max_j X_{ij}$ corresponds to the largest value for each dimension.  

\begin{figure}[]
	\centering
	\subfloat[uniform valuation, linear function]{\includegraphics[width=0.55\textwidth]{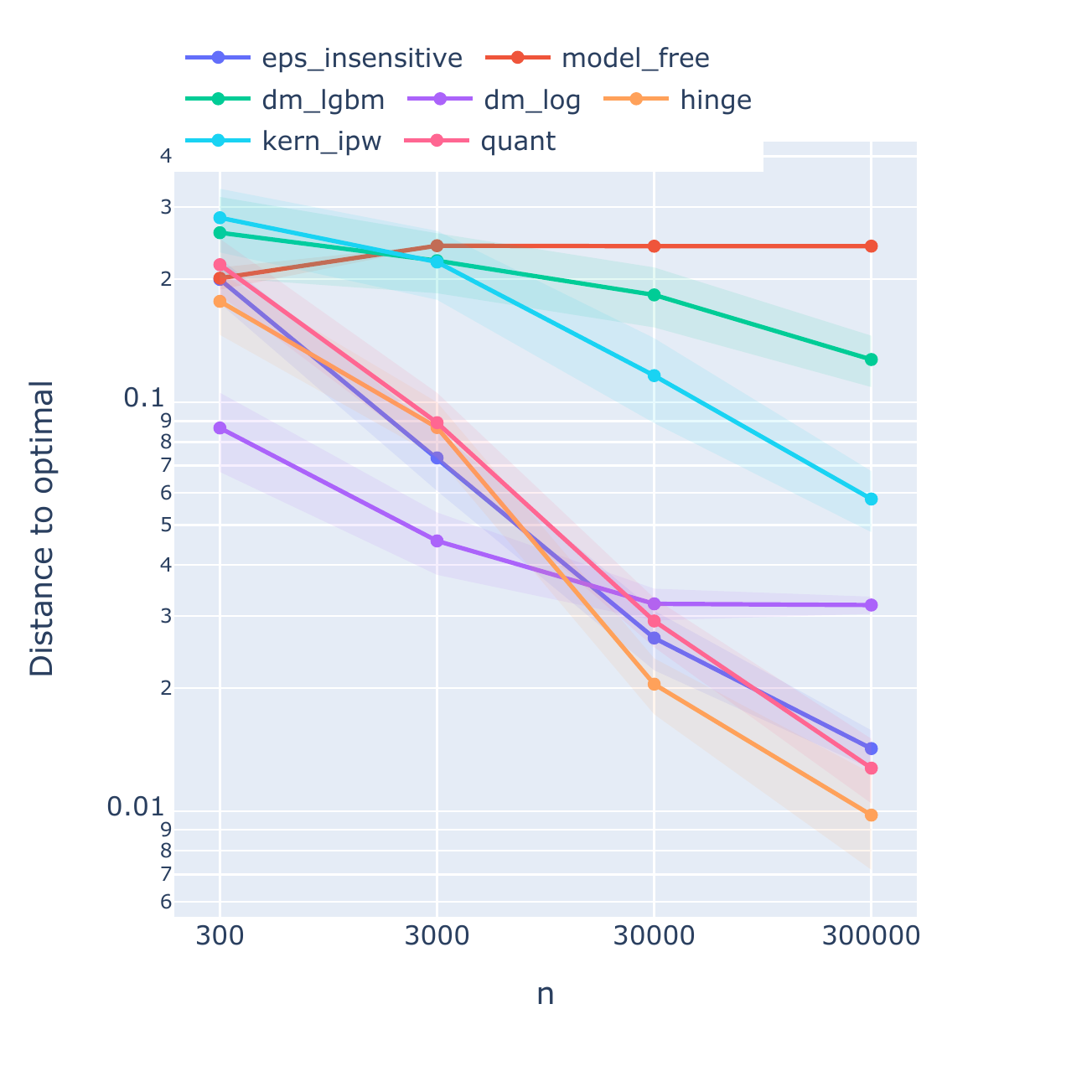}}
	\subfloat[uniform valuation, step function]{\includegraphics[width=0.55\textwidth]{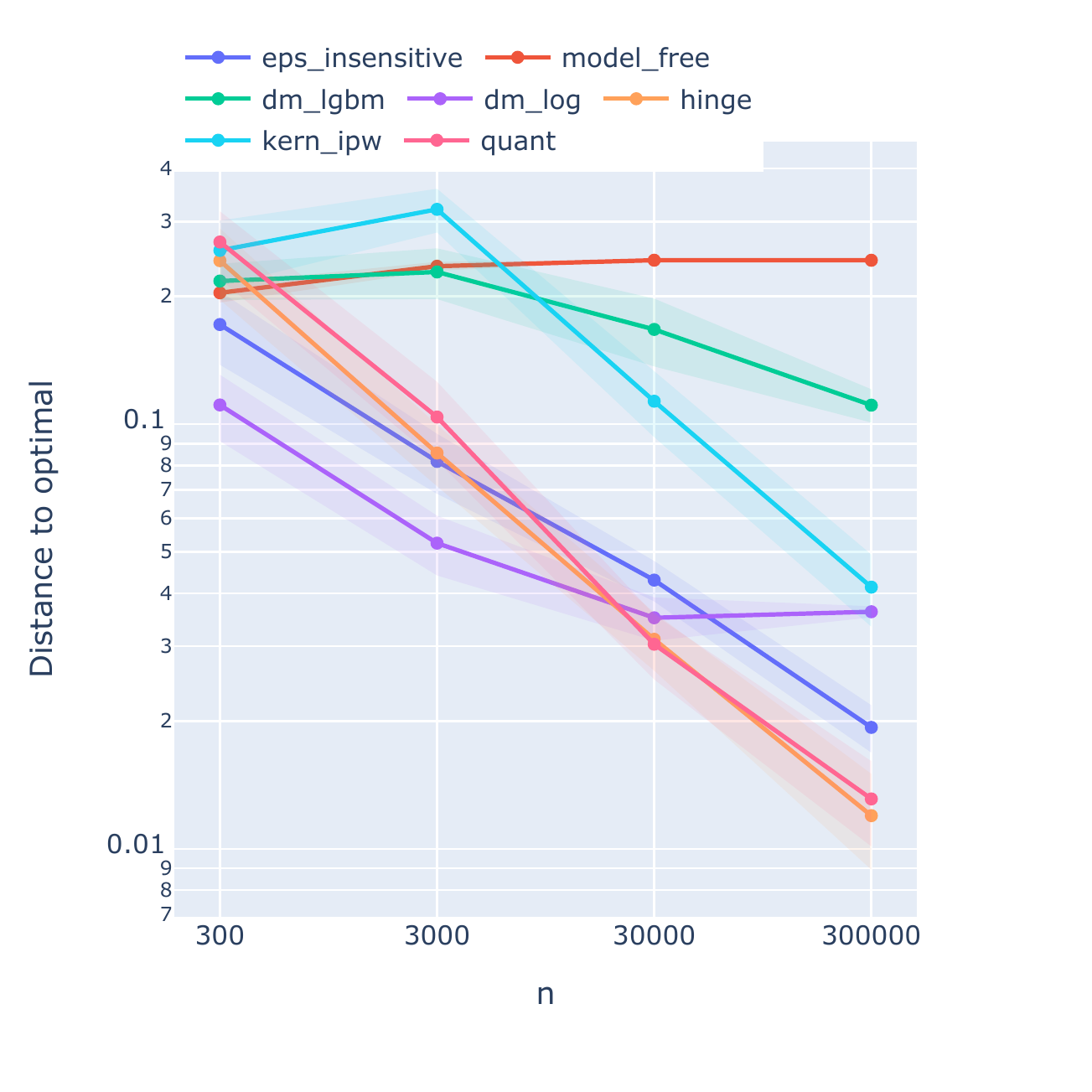}}\\
	\subfloat[exponential valuation, linear function]{\includegraphics[width=0.55\textwidth]{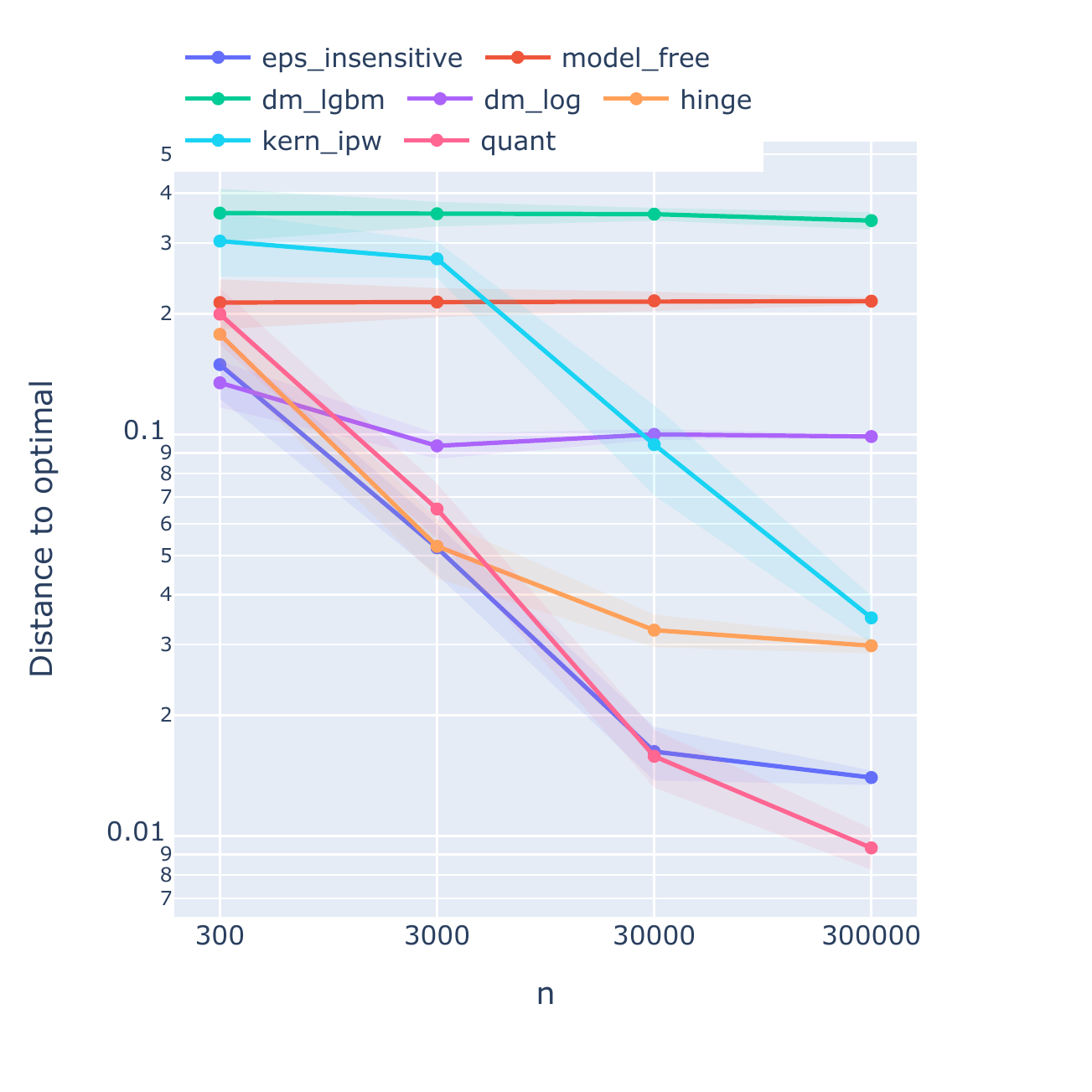}}
	\subfloat[exponential valuation, step function]{\includegraphics[width=0.55\textwidth]{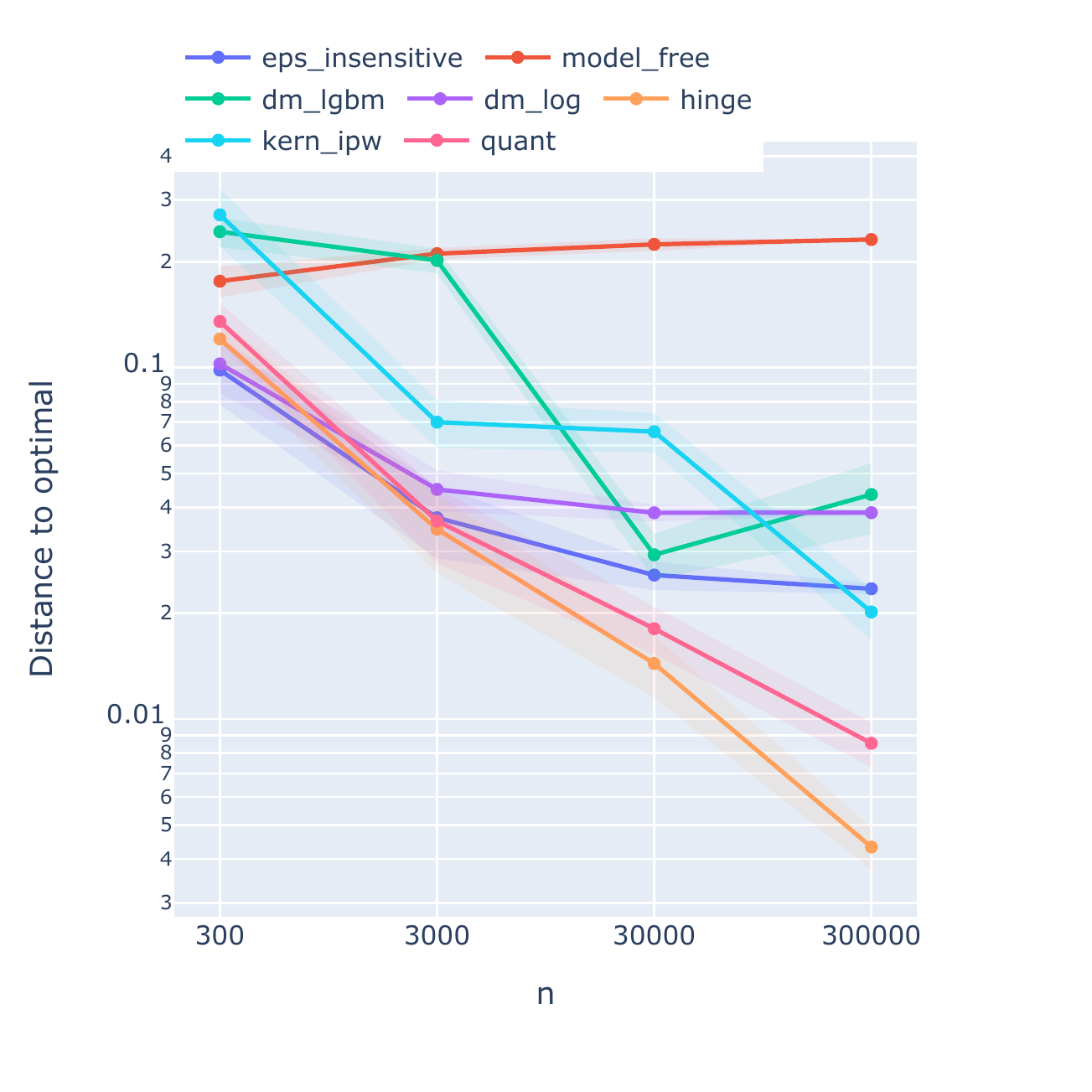}}
	\caption{Distance to optimality (valuation distribution, dependence on $X$)}
	\label{fig:m_2_plots}
\end{figure}

We compare the pricing policies generated to the optimal policy in each scenario, found by minimizing the valuation loss function (Equation \ref{eq:valuation_loss_fn}), although this valuation data isn't available to the other algorithms. We report the distance from the proposed pricing policy $\hat{\pi}$ to the optimal solution $\pi^*$, $\sum_{i=1}^{n}|\pi^*(X_i) - \hat{\pi}(X_i)|$. We vary the dataset size $n \in \{300,3000,30000,300000 \}$.  To initialize the BFGS optimizer, we use an initial solution $\theta_{0_i}= \frac{1}{2}$, for $i \in [2]$. We repeat each simulation 20 times and report the average with plus/minus one standard error, shown shaded in Figure \ref{fig:m_2_plots}. %\ref{fig:step_distance} and \ref{fig:abs_distance}. %These figures correspond to the linear, step, and absolute function covariate dependence of the customer valuation, respectively.

\subsubsection{Discussion:}

We observe that the hinge and quantile pricing loss functions are competitive with and in some cases, outperform state-of-the-art benchmarks. In general, for larger datasets or skewed valuation distributions, the convex pricing loss functions (\texttt{quant}, \texttt{hinge}, \texttt{eps\_insensitive}) tend to perform best. While there is little to differentiate the convex pricing loss functions when the historical pricing policy is uniform, we show in the next section that the \texttt{eps\_insensitive} approach from \cite{ye2018customized} can perform poorly when the historical pricing policy is imbalanced. 

For small datasets $(n=300,3000)$ with uniform valuation distributions, the logistic regression direct method performs well. This is potentially due to the low model complexity of logistic regression, which can have lower variance than more complex estimators such as gradient-boosted trees. While not convex, the estimated logistic regression revenue function is quasi-convex, and empirically the BFGS optimizer is often able to find a price close to the optimal. This is not the case for the boosted tree or kernel models, which are in general highly nonlinear and are unable to find good prices with small data samples. %This suggests that nonparametric demand models are a risky choice for pricing practitioners. 
We note that none of the approaches we benchmark against except \texttt{eps\_insensitive} result in a convex policy optimization problem, so it is possible that any approach could get stuck at a local minimum, an issue that the hinge and quantile pricing loss functions do not have.  %We note that the hinge and quantile pricing loss functions are also dependent on the boosted trees estimator through cross-validation, so it is possible that these approaches might perform better using logistic regression for these small data sizes.

For larger datasets  $(n=30000,300000)$, we observe that the kernel IPW and gradient-boosted tree methods both improve, while the logistic regression does not improve as much. This is likely due to the relatively larger bias of the logistic regression approach, which doesn't decrease with the sample size. Even in the linear setting, the logistic function is not able to exactly model the uniform or exponential complementary CDF. We empirically observe that it seems like the logistic function can better approximate the uniform CDF than the exponential CDF, which is more asymmetrical. Furthermore, as the number of samples increases, we empirically observe that the demand of the kernel IPW becomes smoother, so the optimizer is more likely to find a good solution.

 The convex pricing loss functions (\texttt{quant}, \texttt{hinge}, \texttt{eps\_insensitive}) often find the best prices for larger datasets. We hypothesize that this is because the gradient-boosted tree model used in the cross-validation becomes more accurate with more samples. However, we also observe that the hinge and quantile pricing models tend to choose better policies than the boosted tree model itself. This is likely because optimizing the boosted tree is very challenging due to non-convexity, whereas for each parameter instance in the cross-validation, the hinge and quantile regressions are convex. Furthermore, we emphasize that the hinge and quantile pricing loss algorithms have guarantees on expected revenue, whereas the algorithms we benchmark against do not.

% Interestingly, the policies generated by the boosted tree ensemble was inferior to the logistic regression in nearly all simulations, despite the boosted tree having a significantly higher predictive accuracy for larger datasets, empirically suggesting that the predictive performance on its own is not a good criterion for choosing a demand model to optimize. 

\subsection{Imbalanced historical pricing policies}

\begin{figure}[]
	\centering
	\subfloat[uniform valuation, linear function]{\includegraphics[width=0.55\textwidth]{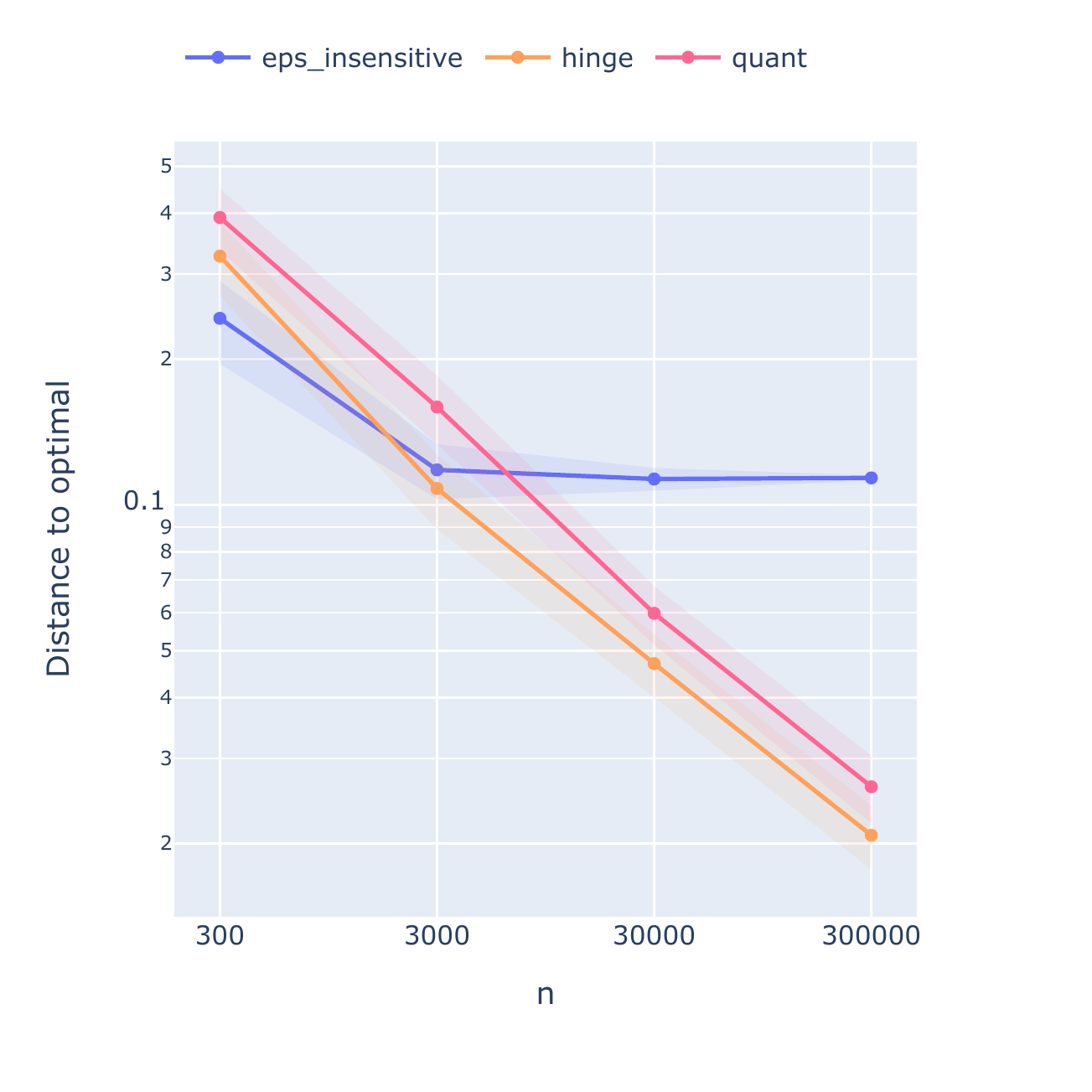}}
	\subfloat[uniform valuation, step function]{\includegraphics[width=0.55\textwidth]{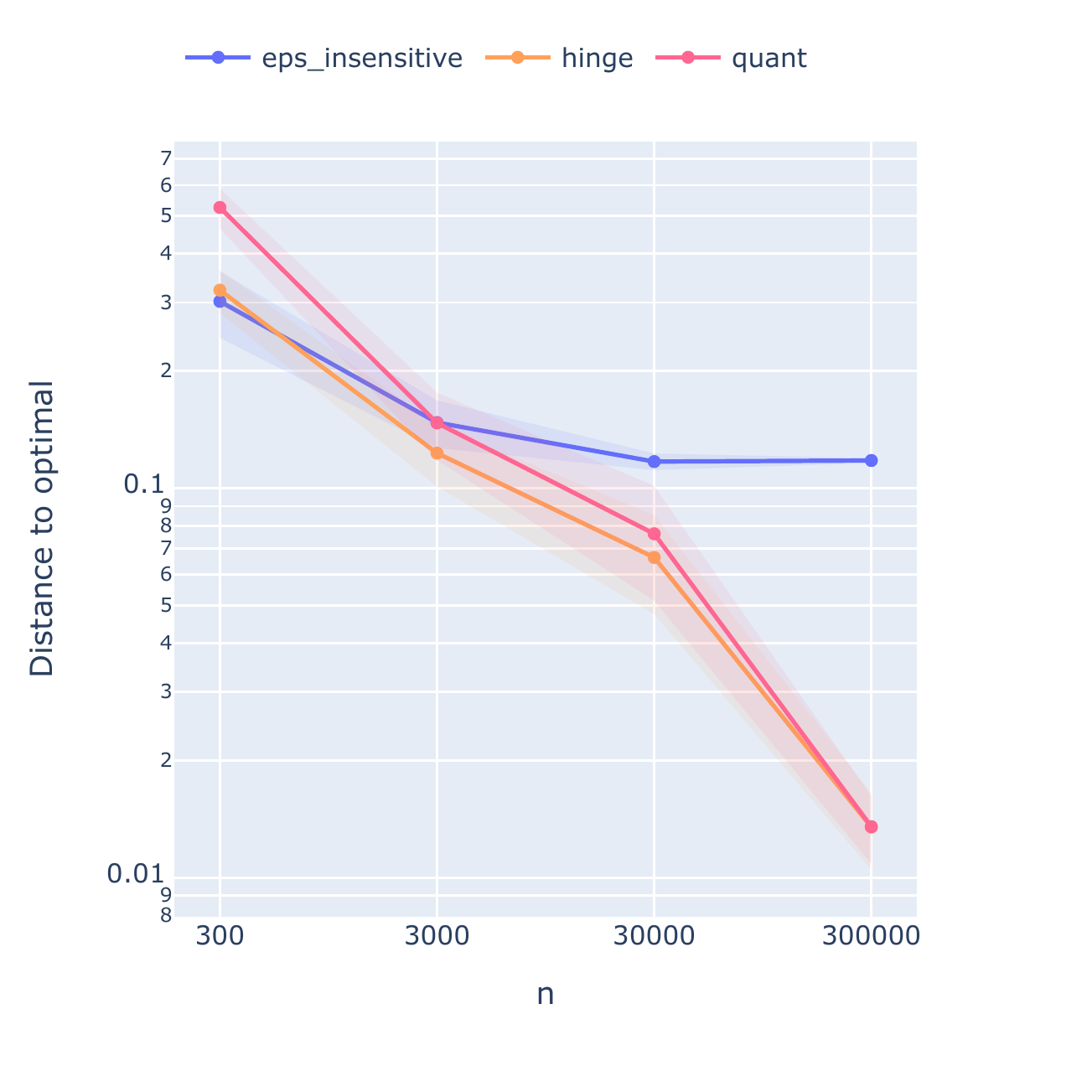}}\\
	\subfloat[exponential valuation, linear function]{\includegraphics[width=0.55\textwidth]{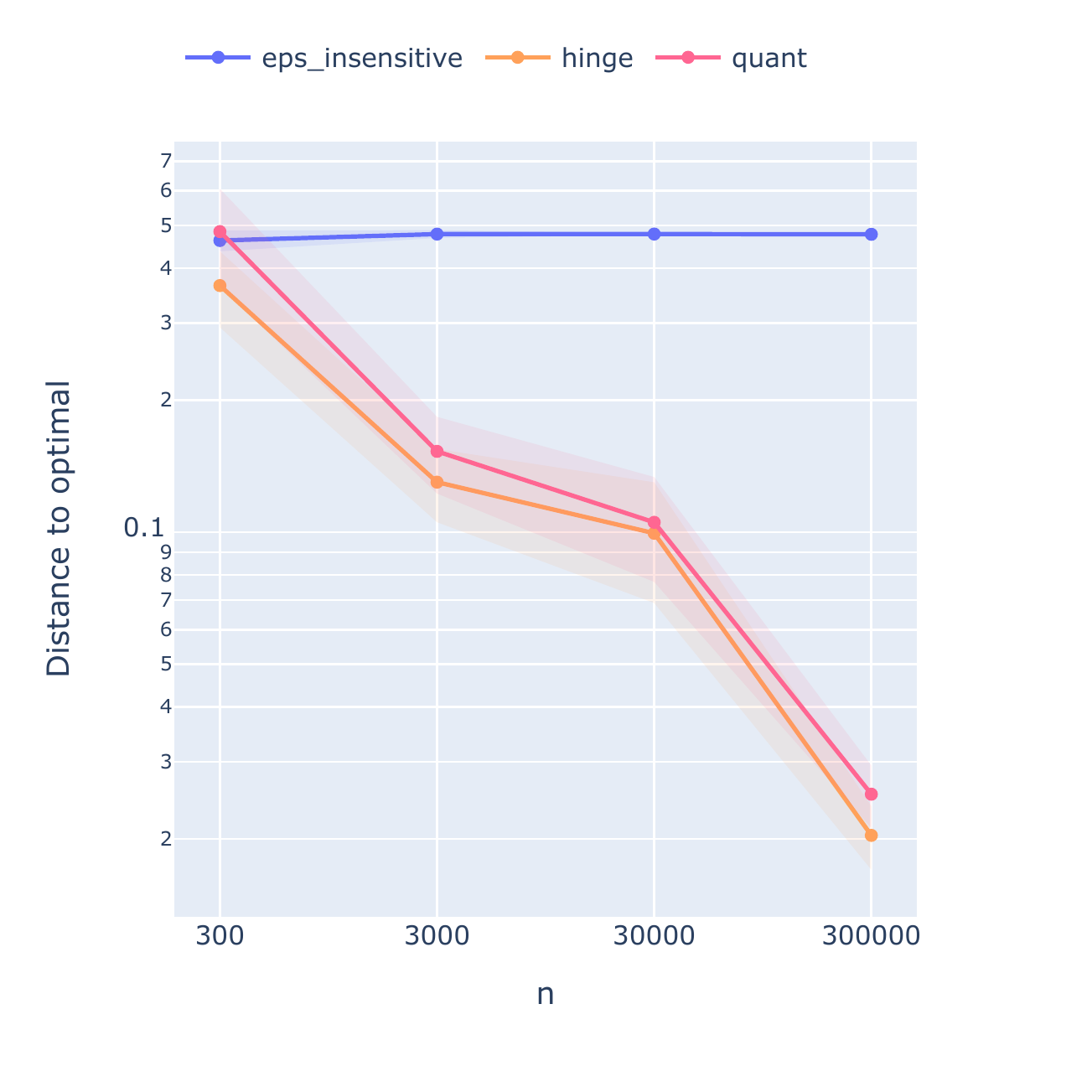} \label{fig:skew_a}} 
	\subfloat[exponential valuation, step function]{\includegraphics[width=0.55\textwidth]{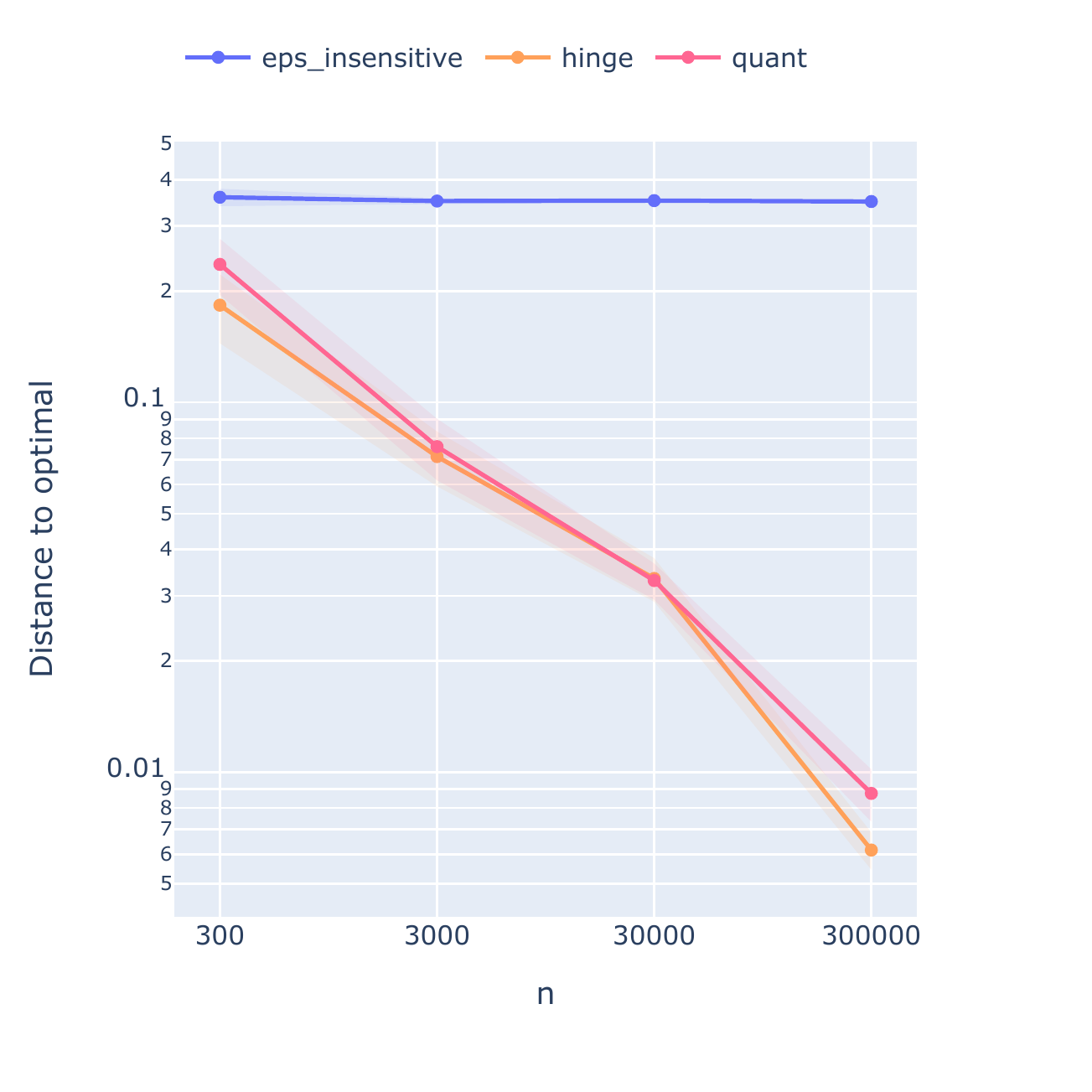}\label{fig:skew_b}} %\
	\caption{Distance to optimality (valuation distribution, dependence on $X$)}
	\label{fig:m_2_plots_skew}
\end{figure}

To better understand the difference between the \texttt{eps\_insensitive} approach from \cite{ye2018customized} and the hinge/quantile loss functions, we ran some additional experiments with an unbalanced historical pricing policy. Historical pricing policies that are not uniform are common in practice where firms are targeting prices to customers to maximize revenue. We repeated the experiments from Section \ref{sec:synth_data}, focusing on the convex pricing loss functions that performed similarly when the historical pricing policy is uniform. For the experiments with a uniform valuation distribution, we update the historical pricing policy to a policy that is skewed toward high prices, $P \sim \text{Triangular}(1,5,5)$, where the parameters are (lowest value, mode, highest value). For experiments with an exponential valuation distribution, we use $P \sim \text{Exp}(0.4)$, which is skewed towards lower prices. We observe the results in Figure \ref{fig:m_2_plots_skew}.

We observe that the \texttt{eps\_insensitive} approach from \cite{ye2018customized} performs significantly worse than the hinge or quantile pricing loss functions when the historical pricing policy is not balanced. In particular, in Figures \ref{fig:skew_a} and \ref{fig:skew_b} when the valuation and historical pricing policies are both exponential, \texttt{eps\_insensitive} is badly biased and the pricing policy does not improve with sample size. This difference is at least partially due to the propensity score adjustments that the hinge and quantile pricing policies use. 

\subsection{Case study: personalized prices for groceries}
\label{sec:dunnhumby}
We also test the proposed approaches on a real-world dataset involving personalized pricing for grocery products. The data we have tracks individual customers and whether they purchased strawberries from a chain of grocery stores at the posted price. In addition, demographic data is collected on the individual customers based on enrollment in a loyalty program, specifically, income, age, gender, marital status, home ownership, size of household, and whether the customer has children. Using this data, \citet{amram2020optimal} and \citet{biggs2021model} have studied the potential of offering personalized prices to customers based on these features to maximize revenue, using interpretable tree-based models. This data was originally collected by the analytics firm Dunnhumby, and we use the cleaned and processed version of the data from \citet{amram2020optimal}, who provide a detailed description of the data. The data size is  $97295$ rows, corresponding to unique customer trips to the supermarket, with $3.49\%$ of trips resulting in a sale of strawberries. Once the features described above have been one-hot encoded, it results in a dataset with dimension $m=34$. Unlike the setting studied in this paper, the historical pricing policy is not known. However, based on the available data, a distribution of historical prices $\phi(p)$ offered is estimated using a log-normal distribution. This pricing policy has no dependence on customer features $X$ since the grocery is not currently using personalized pricing.

% \blu Maybe put in the specific parameters of the dataset \bla

To evaluate the pricing policies, we use the approach used in \citet{biggs2021model}. Specifically, since the data doesn't show the counterfactual sales outcomes associated with different pricing policies, we used a model-based approach to estimate the contextual selling probabilities. These probabilities are estimated using a model trained on a held-out dataset, to avoid bias due to a finite data sample. To achieve this, the data is initially split into a prescription dataset and an evaluation dataset. All models used to prescribe prices (including estimating demand if necessary) are trained on the prescription set. A predictive model is trained on the evaluation dataset, which can estimate the revenue of any given pricing policy. Following \citet{biggs2021model}, we use a \texttt{lightgbm} boosted tree model, as the evaluator model as it had the highest out-of-sample accuracy of the methods we tested (81\% AUC, compared 71\% for logistic regression).
% \begin{figure}[]
% 	\centering
% 	{\includegraphics[width=0.85\textwidth]{fig/reward_dunnhumby.pdf}}
% 	\caption{Estimated per customer revenue for Dunnhumby Data with increasing training size}
% 	\label{fig:dunnhumby}
% \end{figure}
\begin{figure}[]
	\centering
	{\includegraphics[width=0.85\textwidth]{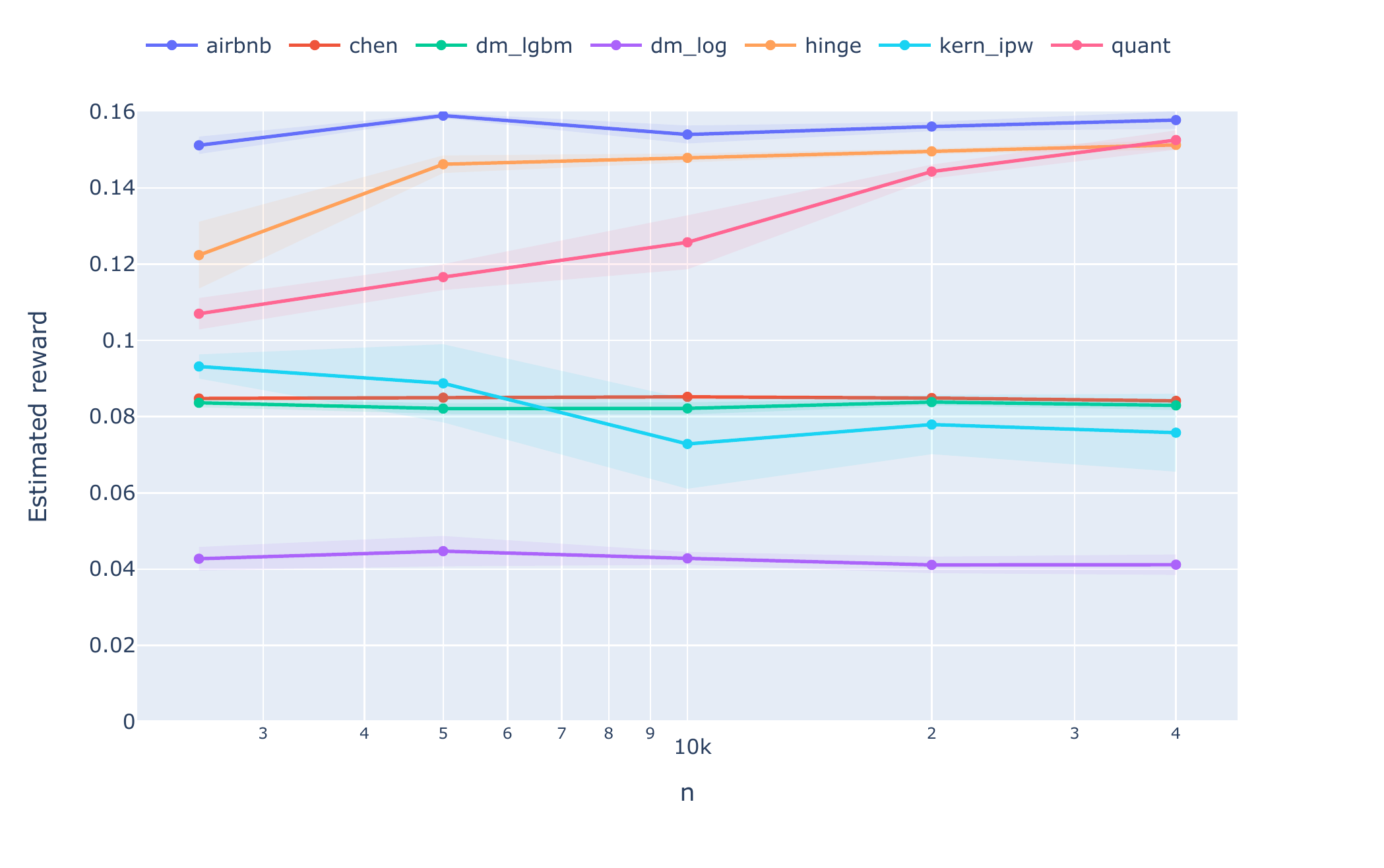}}
	\caption{Estimated per customer revenue for Dunnhumby Data with increasing training size}
	\label{fig:dunnhumby}
\end{figure}

In Figure \ref{fig:dunnhumby}, we show the estimated revenue of different policies, as evaluated by the out-of-sample \texttt{lightgbm} model. We repeat the experiment 10 times for each prescription dataset of size $n=2000,5000,10000, 20000,40000$. The remaining data is used as the evaluation set. The standard error of the average reward is shown in the shaded area. We omitted 2 trials where the \texttt{dm\_lgbm} and \texttt{kern\_IPW} did not converge to a solution.

We observe that the convex pricing loss functions perform better than the other benchmarks on this dataset. In particular,  \texttt{eps\_insensitive} pricing loss function performs the best across all dataset sizes, with the hinge and quantile pricing loss functions improving to comparable performance for larger datasets. This suggests there are practical advantages to the loss function from \cite{ye2018customized}, and further understanding the theoretical revenue guarantees for this function is a worthwhile avenue for future work. Interestingly, we observe that the \texttt{eps\_insensitive}, hinge, and quantile pricing loss functions outperform the \texttt{dm\_lgbm} method, even though the \texttt{lightgbm} model is used to evaluate the estimated revenue. As in the synthetic experiments, this is likely due to the difficulty of optimizing the highly nonlinear estimated revenue curve produced by the boosted tree and is also perhaps why there is no significant improvement as the data size increases. The kernel IPW function has a very high variance and is unreliable, while the logistic regression model does not perform well, possibly due to its inability to capture nonlinear relationships between covariates in the data.

\section{Conclusion}

We have proposed methods to address the problem of contextual pricing using observational posted-price data rather than the well-studied setting where willingness to pay (valuation) data is available. We have focused on pricing algorithms that can achieve bounds on expected revenue and can also be computed tractably. In particular, we have proposed two loss functions, the quantile and hinge pricing loss functions, which are convex and can be easily optimized. We have also shown how to choose the relevant parameters to optimize bounds on expected revenue according to an adversarially chosen valuation distribution, and how to heuristically choose parameter values when the seller has access to an estimated demand model that is accurate but challenging to optimize, such as a neural network or tree ensemble. Using both real-world and synthetic data, we have shown that the proposed loss functions are competitive against commonly used contextual pricing approaches, which are known to work well in practice, but do not have  theoretical expected revenue bounds and may be unpredictable due to non-convexity. 

%In terms of limitations and future work, it would be interesting to extend the setting to include inventory and multiple products, which are common in practice. We have focused on presenting expected revenue bounds in the case where the valuation distribution follows a log-concave distribution, which encompasses the most commonly used distributions in pricing. A possible avenue for future research is to consider regular valuation distributions (a relaxation of log-concavity). %Similarly, we have presented two convex loss functions to be used in this pricing setting, it is possible there are other convex loss functions which have stronger guarantees on revenue.

% References here (outcomment the appropriate case)

% CASE 1: BiBTeX used to constantly update the references
%   (while the paper is being written).
\bibliographystyle{informs2014} % outcomment this and next line in Case 1
\bibliography{bibliography.bib} % if more than one, comma separated

% CASE 2: BiBTeX used to generate mypaper.bbl (to be further fine tuned)
%\input{mypaper.bbl} % outcomment this line in Case 2

%If you don't use BiBTex, you can manually itemize references as shown below.
 \begin{APPENDICES}
\section{Proofs}
\label{sec:proofs}

\begin{repeatproposition}
\label{prop:no_opt_convex_surrogate}
There is no non-constant function $\mathbb{E}_{Y,P}[L(\cdot, Y,P)|V=v]$, left continuous in $v$ and with $L (\cdot, Y,P)$ convex with respect to its first argument, such that for any distribution $(Y,P,V) \sim D$ on $\{0,1\} \times \mathbb{R}_+ \times \mathbb{R}_+$ satisfying Equation \ref{y_def}, there exists a non-negative optimal price $p^* \in \argmax_{p} \mathcal{R}(p)$, satisfying $ \min_p \mathbb{E}_{Y,P} [ L (p, Y,P)] = \mathbb{E}_{Y,P} [ L (p^*, Y,P)]$.
\end{repeatproposition}

\proof{Proof of Proposition \ref{prop:no_opt_convex_surrogate}:}
If $L (p, (Y,P))$ is convex with respect to its first argument, then so is $\mathbb{E}_{Y,P}[L(p, (Y,P))|V]$ (\cite{boyd2004convex}, Section 3.2.1). Therefore $\mathbb{E}_{Y,P}[L(p, (Y,P))|V]$ satisfies the conditions for $L_c(p,V)$ from \cite{mohri2014learning}. \Halmos
% $\mathbb{P}(Y=1,P|V)=$
%Furthermore, $ \mathbb{E}_{Y,P} [ L_O (p, (Y,P))]= \mathbb{E}_V[ \mathbb{E}_{Y,P}[L_O(p, (Y,P))|V]]= \mathbb{E}_V[L_c(p,V)]$
% For any $L_O (p^*, (Y,P))$, Define $L_c(r,b) =  \int_P^b L_O(r,1,v) \phi(v) dv + \int_b^P L_O(r,0,v) \phi(v) dv $. Then the proof from \cite{mohri2014learning} applies. 
\endproof

% \blu Is there a proof which doesn't require right continuous condition? Need to justify/explain. Also potentially confusing that this is the first time we introduce joint distribution over V. Maybe should move this whole thing to the appendix. \bla

% \begingroup
% \def\thetheorem{\ref{revenue_case_1}}
% \begin{lemma}
% Consider a pricing policy which prescribes a price $p_h$ such that $p_h \geq c \int_0^{\infty} \bar{F}_V(p) dp$ and $p_h < p^*$, then:
% $$ \frac{\mathcal{R}(p_h)}{\mathcal{R}(p^*)} \geq  \min_{0 < f < 1} \frac{c(f-1)e^{c(f-1)}}{f \ln(f) }.$$
% \end{lemma}
% \addtocounter{lemma}{-1}
% \endgroup

\lemmaexpgreat*

\proof{Proof of Lemma \ref{revenue_case_1}:}

We begin by providing a lower bound on $p_h$ resulting from the optimality condition $p_h \geq c \int_0^{\infty} \bar{F}_V(p) dp$. This will lead to a revenue bound on $p_h \bar{F}_V(p)$.

\begin{equation}
\label{eq:up_int_id}
\int_{0}^{\infty} \bar{F}_V(p) dp \geq 
\int_{0}^{p^*} \bar{F}_V(p) dp  \geq  
\int_{0}^{p^*} \bar{F}_V(p^*)^{\frac{p}{p^*}} dp  =
\left[ \frac{p^*\bar{F}_V(p^*)^{\frac{p}{p^*}}}{\ln(\bar{F}_V(p^*))} \right]^{p^*}_0 =  \frac{p^*\left(\bar{F}_V(p^*)-1\right)}{\ln(\bar{F}_V(p^*))} 
\end{equation}

Where here the second inequality follows due to log-concavity,  $\bar{F}_V(p) \geq \bar{F}_V(0)^{\theta} \bar{F}_V(p^*)^{1-\theta}$ for $ 0< \theta<1$,  and since $\bar{F}_V(0)=1$. Furthermore, by substituting (\ref{eq:up_int_id}) into condition $p_h \geq c \int_0^{\infty} \bar{F}_V(p) dp$:

\begin{equation}
\frac{p_h}{p^*} \geq  \frac{c}{p^*}
 \int_{0}^{\infty} \bar{F}_V(p) dp \geq
\frac{c \left(\bar{F}_V(p^*)-1\right)}{\ln(\bar{F}_V(p^*))}  \label{price_ratio}
\end{equation}

Therefore:
\begin{equation}
 \frac{\mathcal{R}(p_h)}{\mathcal{R}(p^*)} = 
\frac{p_h\bar{F}_V(p_h)} {p^*\bar{F}_V(p^*)} \geq \frac{p_h\bar{F}_V(p^*)^{\frac{p_h}{p^*}} } {p^*\bar{F}_V(p^*)} \label{rev_monotone_func}
\end{equation}

Where the inequality follows from log-concavity. Furthermore, since $\frac{\mathcal{R}(p_h)}{\mathcal{R}(p^*)} \leq 1$, $\frac{p_h\bar{F}_V(p^*)^{\frac{p_h}{p^*}} } {p^*\bar{F}_V(p^*)} \leq 1$. We also have that $0\leq \bar{F}_V(p^*) \leq 1$ and $0 \leq \frac{p_h}{p^*} < 1$. Therefore, according to Lemma \ref{monotone_lemma}, which can be found in Appendix \ref{aux_proofs},  $\frac{p_h\bar{F}_V(p^*)^{\frac{p_h}{p^*}} } {p^*\bar{F}_V(p^*)}$ is a monotone increasing function within this range, which validates the substitution of (\ref{price_ratio}) into  (\ref{rev_monotone_func}):
\begin{align*}
 \frac{\mathcal{R}(p_h)}{\mathcal{R}(p^*)}  \geq \frac{p_h\bar{F}_V(p^*)^{\frac{p_h}{p^*}} } {p^*\bar{F}_V(p^*)} &\geq \frac{c \left(\bar{F}_V(p^*)-1\right)}{\ln(\bar{F}_V(p^*))}  \bar{F}_V(p^*)^{\frac{c \left(\bar{F}_V(p^*)-1\right)}{\ln(\bar{F}_V(p^*))}-1}  \\
 & \geq \frac{c \left(\bar{F}_V(p^*)-1\right)}{\bar{F}_V(p^*) \ln(\bar{F}_V(p^*))}  e ^{c(\bar{F}_V(p^*)-1) }  
\end{align*}
% \begin{equation*}
%  \frac{\mathcal{R}(p_h)}{\mathcal{R}(p^*)}  \geq \frac{p_h\bar{F}_V(p^*)^{\frac{p_h}{p^*}} } {p^*\bar{F}_V(p^*)} \geq \frac{c \left(\bar{F}_V(p^*)-1\right)}{\ln(\bar{F}_V(p^*))}  \bar{F}_V(p^*)^{\frac{c \left(\bar{F}_V(p^*)-1\right)}{\ln(\bar{F}_V(p^*))}-1}  
%   \geq \frac{c \left(\bar{F}_V(p^*)-1\right)}{\bar{F}_V(p^*) \ln(\bar{F}_V(p^*))}  e ^{c(\bar{F}_V(p^*)-1) }  
% \end{equation*}
Setting $f=\bar{F}_V(p^*)$ and minimizing this to find the worst-case distribution completes the proof. \Halmos
\endproof

% We note that the condition in Lemma \ref{revenue_case_1}, $p_h \geq c \int_0^{\infty} \bar{F}_V(p) dp$, is clearly satisfied by the optimality conditions for the hinge pricing loss function. We prove an expected revenue bound for the case where the prescribed price is greater than the optimal price $p_h \geq p^*$ in Lemma \ref{lemma_hinge_revenue_case_2}. 

\lemmaminc*
% \begin{repeatlemma}
%  \label{minimum_at_c}
% $  \min_{0 < f < 1}  \dfrac{c(f-1)e^{c(f-1)}}{f \ln(f) }= c$ for $0 \leq c \leq 0.5$
% \end{repeatlemma}

\proof{Proof of Lemma \ref{minimum_at_c}:} 
We achieve this by showing that the gradient of $\frac{c(f-1)e^{c(f-1)}}{f \ln(f) }$ is negative for all $0 < f < 1$ as long as $0 \leq c < 0.5$, so the solution lies on the boundary as $f \rightarrow 1$.

\begin{equation*}
\diffp{}{{f}} \dfrac{c(f-1)e^{c(f-1)}}{f \ln(f) } = \dfrac{ce^{cf-c}\left( \ln(f)(cf(f-1)+1)+(1-f) \right)}{f^2\ln(f)^2}
\end{equation*}

Clearly $\dfrac{ce^{cx-c}}{f^2\ln(f)^2}\geq 0$, so we will focus on showing $\ln(f)(cf(f-1)+1)+(1-f) \leq 0$. To achieve this, we will use the following bound, $\ln(f) \leq (f-1)-\frac{1}{2}(f-1)^2$ for $0 < f < 1$.  This follows from a Taylor expansion of $\ln(f)$ \citep{apostol1991calculus}.

\begin{proposition}\textbf{Taylor's theorem with mean-value form of the remainder:} Let $g: \mathbb{R} \rightarrow \mathbb{R}$ be a $k+1$ times differentiable with all derivatives continuous. Fix $x_0$ and let
\begin{align*}
    P_k(x)&=g(x_0)+g'(x_0)(x-x_0)+...+\frac{g^{(k)}(x_0)}{k!}(x-x_0)^k\\
    R_k(\xi,x)&=\frac{g^{(k+1)}(\xi)}{(k+1)!}(x-x_0)^{k+1}
\end{align*}
Then $g(x)=P_k(x)+R_k(\xi;x)$ for some $\xi$ between $x$ and $x_0$. Furthermore, if $R_k(\xi,x) \leq 0$ for all $\xi$ between $x$ and $x_0$, then $g(x)\leq P_k(x)$.
\end{proposition}

The second order Taylor expansion of $\ln(f)$ around 1 is $(f-1)-\frac{1}{2}(f-1)^2$. The remainder is $R_k(\xi,f)=\dfrac{2}{\xi^3 3!}(f-1)^3$. Clearly for all $0 < f < 1$ and $ f \leq \xi \leq 1$, $R_k(\xi,x) \leq 0$, so $\ln(f) \leq (f-1)-\frac{1}{2}(f-1)^2$ in this range.

Furthermore, since $cf(f-1)+1 \geq -0.25c+1 \geq 0$ for $0 < f < 1$ and $0 \leq c \leq 1$, we have that:
\begin{equation*}
\ln(f)(cf(f-1)+1)+(1-x) \leq ((f-1)-\frac{1}{2}(f-1)^2)(cf(f-1)+1)+(1-f) = \frac{1}{2} (f-1)^2 ((3-f)cf-1)
\end{equation*}

Because $(3-f)f$ is maximized at $f^*=1$ if $0 < f \leq 1$ and $(3-f^*)f^*=2$, it follows that for $c\leq 0.5$, $0 < f < 1$,  $((3-f)cf-1) < 0$. As a result, the gradient is negative for $0 < f < 1$, and the minimum occurs on the boundary $f \rightarrow 1$. Furthermore, 
\begin{equation*}
\lim_{f \rightarrow 1} \frac{c(f-1)e^{c(f-1)}}{f \ln(f) }=c
\end{equation*}

Which follows from applying L'Hopital's rule, which is states that $\lim_{f \rightarrow a} \frac{f(x)}{g(x)}=\lim_{f \rightarrow a} \frac{f'(x)}{g'(x)}$. As a result, $\lim_{f \rightarrow 1} \frac{f-1}{\ln(f)}=1$. Furthermore, clearly $\lim_{f \rightarrow 1} \frac{e^{c(f-1)}}{f}=1$, which proves the result. \Halmos
\endproof

\lemmahingeother*
% \begin{repeatlemma}
% \label{lemma_hinge_revenue_case_2}
% If there exists a pricing rule $p_h$ satisfying optimality conditions from Lemma \ref{lemma_opt_cond_hinge} and $p_h > p^*$, then:
% % \begin{equation} \frac{\mathcal{R}(p_h)}{\mathcal{R}(p^*)} \geq   
% %     \begin{dcases}
% %           \min_{z\leq -2c} \frac{cze^{-z(\frac{1}{c}-1)-1}}{z+c}   & \text{if} ~~  c \geq \frac{p_h  \bar{F}_V(p_h)'}{\bar{F}_V(p_h)(\ln(\bar{F}_V(p_h)) -1) + p_h \bar{F}_V(p_h)'}  \\
% %       (c + 1)e^{-c}   &  \text{if} ~~ c \leq \frac{p_h  \bar{F}_V(p_h)'}{\bar{F}_V(p_h)(\ln(\bar{F}_V(p_h)) -1) + p_h \bar{F}_V(p_h)'} 
% %     \end{dcases}  
% \begin{equation} 
% \frac{\mathcal{R}(p_h)}{\mathcal{R}(p^*)} \geq   
%           \min \left\{ \min_{z\leq -2c} \frac{cze^{-z(\frac{1}{c}-1)-1}}{z+c}, 
%       (c + 1)e^{-c} \right\} 
% \end{equation}
% \end{repeatlemma}

\proof{Proof of Lemma \ref{lemma_hinge_revenue_case_2}:} 

We start by proving some lower bounds on $\bar{F}_V(p_h)$, related to the gradient and position of $p_h$. This lower bound also acts as a lower bound on the revenue, as for the regime where $p^* \leq p_h$, we have that $\frac{\mathcal{R}(p_h)}{\mathcal{R}(p^*)} \geq \bar{F}_V(p_h)$. We will show that: 
\begin{equation}
\bar{F}_V(p_h) \geq e^{-sp_h(\frac{1}{c}-1)-1}
\end{equation}

Where $s$ is a gradient of the log of the complementary CDF at $p_h$: $$s = \diffp{\ln(\bar{F}_V(p))}{{p}} \Big\rvert_{p=p_h}= \frac{\bar{F}'_V(p_h)}{\bar{F}_V(p_h)}<0.$$

This also implies that $\bar{F}_V(p_h) \geq e^{-c}$. We begin by upper bounding the complementary CDF. The function $\ln(\bar{F}_V(p))$ is concave, so it lies below its tangent at $p_h$:

\begin{equation*}
    \ln(\bar{F}_V(p))  \leq \ln(\bar{F}_V(p_h)) + s(p - p_h) 
    \implies \bar{F}_V(p)  \leq \bar{F}_V(p_h) e^{s(p - p_h)}
\end{equation*}

 Furthermore, because $\bar{F}_V(p) \geq \bar{F}_V(p_h)^{\frac{p}{p_h}}$ for $0 \leq p \leq p_h$ and $\bar{F}_V(p) \leq \bar{F}_V(p_h)^{\frac{p}{p_h}}$ for $p_h \leq p < \infty$, it follows that $\bar{F}'_V(p_h) \leq \diffp{\bar{F}_V(p_h)^{\frac{p}{p_h}}}{{p}} \Big\rvert_{p=p_h} = \bar{F}_V(p_h) \frac{\ln(\bar{F}_V(p_h))}{p_h}$, which implies $s \leq \frac{\ln(\bar{F}_V(p_h))}{p_h}$. Furthermore, since $\bar{F}_V(p) \leq 1$, a refined upper bound is

\begin{equation}
\bar{F}_V(p) \leq \bar{F}_{UB}(p) :=   \begin{cases}
      1   &  \text{if} ~~ p \leq p_h - \frac{\ln(\bar{F}_V(p_h))}{s}  \\ 
      \bar{F}_V(p_h) e^{s(p - p_h)}   & \text{if} ~~  p \geq p_h - \frac{\ln(\bar{F}_V(p_h))}{s}    \\ 
    \end{cases} \label{cdf_upper_bound} 
\end{equation}

Where $p_h - \frac{\ln(\bar{F}_V(p_h))}{s}$ is where the upper bounds intersect. From the optimality condition in Lemma \ref{lemma_opt_cond_hinge}:
\begin{align*}
    p_h  = c \int_0^{\infty} \bar{F}_V(p) dp 
      & \leq c \left(\int_0^{p_h - \frac{\ln(\bar{F}_V(p_h))}{s}} 1 dp + \int_{p_h - \frac{\ln(\bar{F}_V(p_h))}{s}}^{\infty} \bar{F}_V(p_h) e^{s(p - p_h)} dp \right)\\
      & = c\left( p_h - \frac{\ln(\bar{F}_V(p_h))}{s} + \left[ \frac{\bar{F}_V(p_h)}{s} e^{s(p - p_h)}   \right]_{p_h - \frac{\ln(\bar{F}_V(p_h))}{s}}^{\infty} \right)\\
      & = c\left( p_h - \frac{\ln(\bar{F}_V(p_h))}{s} +  \frac{\bar{F}_V(p_h)}{s} e^{-\ln(\bar{F}_V(p_h))} \right) 
\end{align*}

With some rearranging, this implies:
\begin{equation}
\bar{F}_V(p_h) \geq e^{-sp_h(\frac{1}{c}-1)-1} \label{upper_bound_FV_1}
\end{equation}

Since $s \leq \frac{\ln(\bar{F}_V(p_h))}{p_h}$ for $0< c \leq 1$, this implies:
 \begin{equation}
 \bar{F}_V(p_h) \geq e^{-c} \label{fv_geq_ec}
 \end{equation}

 This also implies $\frac{\mathcal{R}(p_h)}{\mathcal{R}(p^*)} \geq e^{-1}$ when $c=1$. We now turn to two specific revenue bounds depending on the size of $c$. An upper bound on the optimal revenue can be found by optimizing the upper bound on the revenue function corresponding to (\ref{cdf_upper_bound}), $ p^* \bar{F}_V(p^*) \leq  \max_p p\bar{F}_{UB}(p)$, where $p^*_{UB}:=\argmax_p p\bar{F}_{UB}(p)$.
 
 For certain values of $c$, $p^*_{UB} \geq p_h - \frac{\ln(\bar{F}_V(p_h))}{s}$, corresponding to $p^*_{UB}$ lying on the second section of (\ref{cdf_upper_bound}). It follows that
 %This corresponds to $c \geq \frac{p_h  \bar{F}_V(p_h)'}{\bar{F}_V(p_h)(\ln(\bar{F}_V(p_h)) -1) + p_h \bar{F}_V(p_h)'}$. 
 %This corresponds to the case when $p^*_{UB} \geq p_h - \frac{\ln(\bar{F}_V(p_h))}{s}$ from (\ref{cdf_upper_bound}).  In this case, 
 $$\max_p p\bar{F}_{UB}(p) = \max_p p \bar{F}_V(p_h) e^{s(p - p_h)}$$
 From the optimality conditions, this has an maximum at $p_{UB}^*=\frac{-1}{s}$, so that $p_{UB}^*\bar{F}_{UB}(p_{UB}^*)= \dfrac{-\bar{F}_V(p_h)}{s} e^{-1 - sp_h}$. Furthermore, due to the position of $p_{UB}^*$ we can derive some additional useful bounds. Because $ p^*_{UB}\leq p_h$, this implies that: 
  \begin{equation}
  sp_h \leq -1 \label{gradient_bound_FV}
  \end{equation}
  
 Furthermore, because $p^*_{UB} \geq p_h - \frac{\ln(\bar{F}_V(p_h))}{s}$, this implies that:
 \begin{equation}
-1 \leq  s p_h - \ln(\bar{F}_V(p_h)) \label{another_bound_FV}
  \end{equation}

If we return to the expected revenue bounds and substitute $p_{UB}^*\bar{F}_{UB}(p_{UB}^*)$, we have 
\begin{equation*}
 \frac{\mathcal{R}(p_h)}{\mathcal{R}(p^*)} = \frac{p_h\bar{F}_V(p_h)} {p^*\bar{F}_V(p^*)} \geq \dfrac{p_h\bar{F}_V(p_h)}{\frac{-\bar{F}_V(p_h)}{s} e^{-1 - sp_h}} = - sp_h e^{1 + sp_h}% \geq - (\bar{F}_V(p_h)-1)\bar{F}_V(p_h) \geq (c+1)e^{-c}
\end{equation*}

It can be shown that $- sp_h e^{1 + sp_h}$ is monotone increasing in $sp_h$ for $sp_h \leq -1$, the upper bound from (\ref{gradient_bound_FV}). Therefore, by substituting the lower bound, $ s p_h \geq   \ln(\bar{F}_V(p_h)) -1 $ from (\ref{another_bound_FV}), we have that
\begin{equation}
  \frac{\mathcal{R}(p_h)}{\mathcal{R}(p^*)}  \geq - sp_h e^{1 + sp_h} \geq - (\ln(\bar{F}_V(p_h)) -1)\bar{F}_V(p_h) \geq (c+1)e^{-c} \label{eq:other_rev_bound}
\end{equation}

The second inequality follows from substituting the bound $\bar{F}_V(p_h) \geq e^{-c}$ from (\ref{fv_geq_ec}) and since $- (\ln(\bar{F}_V(p_h)) -1)\bar{F}_V(p_h)$ is monotone increasing in the range $0 \leq \bar{F}_V(p_h)\leq 1$.

%  Substituting this expression into the expected revenue ratio:
 
% \begin{equation*}
%  \frac{\mathcal{R}(p_h)}{\mathcal{R}(p^*)} =\frac{p_h\bar{F}_V(p_h)} {p^*\bar{F}_V(p^*)} \geq \dfrac{sp_h\bar{F}_V(p_h)}{sp_h - \ln(\bar{F}_V(p_h))}.    
% \end{equation*}

 Next we address the case when $p^*$ lies on the upper bound where $\bar{F}_{UB}(p^*)=1$, such that $p^*_{UB} \leq p_h - \frac{\ln(\bar{F}_V(p_h))}{s} $. This means that the inverse of the statement in (\ref{another_bound_FV}) is true, which through rearranging implies $\bar{F}_V(p_h) \geq e^{sp_h+1}$. Furthermore we have another lower bound on  $\bar{F}_V(p_h) $ from (\ref{upper_bound_FV_1}). Taking the maximum of these bounds gives us:  
 \begin{equation}
\bar{F}_V(p_h) \geq   \begin{cases}
      e^{sp_h+1}   &  \text{for} ~~ sp_h \geq -2c  \\ 
      e^{-sp_h(\frac{1}{c}-1)-1}   & \text{for} ~~   sp_h \leq -2c   \\ 
    \end{cases} \label{lower_bound_cdf} 
\end{equation}

 %This corresponds to the case $c \leq \frac{p_h  \bar{F}_V(p_h)'}{\bar{F}_V(p_h)(\ln(\bar{F}_V(p_h)) -1) + p_h \bar{F}_V(p_h)'}$.  
It also immediate that for this case, $\max_p p\bar{F}_{UB}(p) = p_h - \frac{\ln(\bar{F}_V(p_h))}{s}$. Substituting this expression into the expected revenue ratio:

\begin{equation*}
 \frac{\mathcal{R}(p_h)}{\mathcal{R}(p^*)} =\frac{p_h\bar{F}_V(p_h)} {p^*\bar{F}_V(p^*)} \geq \dfrac{sp_h\bar{F}_V(p_h)}{sp_h - \ln(\bar{F}_V(p_h))}.    
\end{equation*}

Furthermore, $\dfrac{sp_h\bar{F}_V(p_h)}{sp_h - \ln(\bar{F}_V(p_h))}$ is monotone increasing in $\bar{F}_V(p_h)$ for $\bar{F}_V(p_h) \geq e^{sp_h+1}$. Therefore, we can substitute for $\bar{F}_V(p_h)$:

 \begin{equation*}
\dfrac{sp_h\bar{F}_V(p_h)}{sp_h - \ln(\bar{F}_V(p_h))} \geq \dfrac{sp_h e^{sp_h+1}}{sp_h - (sp_h+1)} = - sp_h e^{sp_h+1}  ~~~\forall sp_h \geq -2c 
\end{equation*}

As mentioned previously, $- sp_h e^{sp_h+1}$ is monotone increasing for $sp_h \leq -1$, so it is minimized by $sp_h= -2c$ in the range $ -2c  \leq sp_h \leq -1$. The upper bound of $-1$ on $sp_h$ is from (\ref{gradient_bound_FV}), which also holds in this case.  This leads to a bound:
 \begin{equation}
 \frac{\mathcal{R}(p_h)}{\mathcal{R}(p^*)} \geq   2c e^{-2c+1}  ~~ \text{if}  ~ -2c  \leq sp_h \leq -1  \label{looser_LB}   
\end{equation}

For $sp_h \leq -2c $, by substituting the alternative bound for $\bar{F}_V(p_h)$ from (\ref{lower_bound_cdf}):
 \begin{equation*}
\dfrac{sp_h\bar{F}_V(p_h)}{sp_h - \ln(\bar{F}_V(p_h))} \geq \dfrac{sp_h  e^{-sp_h(\frac{1}{c}-1)-1}}{sp_h + sp_h(\frac{1}{c}-1)+1} = \dfrac{csp_h  e^{-sp_h(\frac{1}{c}-1)-1}}{sp_h + c} ~~~\forall sp_h \leq -2c  
\end{equation*}

Therefore by substituting for $z=sp_h$ and taking the minimum
\begin{equation}
 \frac{\mathcal{R}(p_h)}{\mathcal{R}(p^*)} \geq \min_{z\leq -2c}  \dfrac{cz  e^{-z(\frac{1}{c}-1)-1}}{z + c}  ~~ \text{if}  ~ sp_h \leq -2c   \label{tighter_LB}    
\end{equation}

Furthermore, it can be shown using simulation that $\min_{z\leq -2c}  \frac{cz  e^{-z(\frac{1}{c}-1)-1}}{z + c} \leq  2c e^{-2c+1} $ for all $0 < c < 1$, so the adversary will use the regime where $sp_h \leq -2c $ and the corresponding bound from (\ref{tighter_LB}). However, there are values of $c$ for which  $(c+1)e^{-c} \leq \min_{z\leq -2c}  \dfrac{cz  e^{-z(\frac{1}{c}-1)-1}}{z + c}$, hence leading to the bound: % is less than (\ref{looser_LB}) for 

$$\frac{\mathcal{R}(p_h)}{\mathcal{R}(p^*)} \geq   
          \min \left\{ \min_{z\leq -2c} \frac{cze^{-z(\frac{1}{c}-1)-1}}{z+c}, 
      (c + 1)e^{-c} \right\} \Halmos $$
      
% Since $p_h  > p^*$ and $\bar{F}_V(p^*) \leq 1$,
%   $\frac{\mathcal{R}(p_h)}{\mathcal{R}(p^*)} =\frac{p_h\bar{F}_V(p_h)} {p^*\bar{F}_V(p^*)} \geq e^{-c}    $.
\endproof

\minhingeclosedexpr*

\proof{Proof of Corollary \ref{min_hinge_closed_expr}:}

We begin by examining the first order optimality conditions for $\dfrac{cz  e^{-z(\frac{1}{c}-1)-1}}{z + c}$.

 \begin{align*}
\diffp{}{{z}}  \dfrac{cz  e^{-z(\frac{1}{c}-1)-1}}{z + c}&= \frac{ce^{-z(\frac{1}{c}-1)-1}}{z+c}-\dfrac{ce^{-z(\frac{1}{c}-1)-1}}{(z+c)^2}-\dfrac{cz(\frac{1}{c}-1) e^{-z(\frac{1}{c}-1)-1}}{z + c}\\
& =\dfrac{ce^{-z(\frac{1}{c}-1)-1}}{(z + c)^2}\left( c^2z+c^2+cz^2-cz-z^2\right)
  \end{align*}

Therefore, a root lies where $(c-1)z^2 +c(c-1)z + c^2=0$. Applying the quadratic formula to find this root

 \begin{align*}
 \dfrac{-c(c-1) 	\pm \sqrt{c^2(c-1)^2-4(c-1)c^2} }{2(c-1)}&= \dfrac{-c(1-c) 	\pm \sqrt{c^2(c-1)(c-5)} }{2(c-1)}\\
  &= \dfrac{c\left( 1-c \pm \sqrt{(c-1)(c-5)}\right) }{2(c-1)}
\\
  &= c\left( -\frac{1}{2} \pm  \dfrac{\sqrt{(c-1)(c-5)} }{2(c-1)}\right)
   % &=   \dfrac{c}{2}\left(\sqrt{\dfrac{c-5}{c-1}} -1 \right)
    \end{align*}
  \endproof

  We also know that $z \leq -2c$. Since $0<c<1$, $\sqrt{(c-1)(c-5)}\geq 0$ and  $c-1 \leq 0$, this constraint is only satisfied for  
    $$ c\left( -\frac{1}{2} +  \dfrac{\sqrt{(c-1)(c-5)} }{2(c-1)}\right).
   $$ This can be further simplified to:

   $$-\dfrac{c}{2}\left(\sqrt{\dfrac{c-5}{c-1}} +1 \right). \Halmos$$

\endproof

\quantgreater*
% \begin{repeatlemma}
% If there exists a pricing rule $p_q$ satisfying optimality conditions from Lemma \ref{lemma_opt_cond_quant} and $p_q > p^*$, then:
% $$ \frac{\mathcal{R}(p_q)}{\mathcal{R}(p^*)} \geq  \min_{\tau \leq z \leq 1} \frac{z\tau(\ln(z)+1)-z^2} {\tau-z}.$$
% \end{repeatlemma}

 \proof{Proof of Lemma \ref{lemma_quantile_revenue_case_2}:} 
% \begin{align*}
%     \int_{0}^{p_q} \bar{F}_V(p) dp  & = \int_{0}^{p^*} \bar{F}_V(p) dp + \int_{p^*}^{p_q} \bar{F}_V(p) dp    \\
%     & \geq  p^* \bar{F}_V(p^*) + \int_{p^*}^{p_q} \bar{F}_V(p^*)^{\frac{p_q-p}{p_q-p^*}}\bar{F}_V(p_q)^{\frac{p-p^*}{p_q-p^*}}  dp \\
%     % & \geq  p^* \bar{F}_V(p^*) + \bar{F}_V(p^*)^{\frac{p_q}{p_q-p^*}}\bar{F}_V(p_q)^{\frac{-p^*}{p_q-p^*}} \left[ \frac{(p_q-p^*)\left( \frac{\bar{F}_V(p_q)}{\bar{F}_V(p^*)} \right)^{\frac{p}{p_q-p^*}}} {\ln\left( \frac{\bar{F}_V(p_q)}{\bar{F}_V(p^*)} \right)} \right]^{p_q}_{p^*} \\
%     & =  p^* \bar{F}_V(p^*) + \frac{(\bar{F}_V(p_q)-\bar{F}_V(p^*)) (p_q-p^*) }{\ln(\bar{F}_V(p_q))-\ln(\bar{F}_V(p^*))}
% \end{align*}

From the optimality condition for Lemma \ref{lemma_opt_cond_quant} (specifically Equation \ref{eq:quant_opt_cond}), we have that 
 $\int_{0}^{p_q} \tau \bar{F}_V(p) dp  = \int_{p_q}^{\infty} (1-\tau) \bar{F}_V(p)  dp    $. By  lower bounding $\int_{0}^{p_q} \tau \bar{F}_V(p) dp $ and upper bounding $ \int_{p_q}^{\infty} (1-\tau) \bar{F}_V(p)  dp $ and doing some algebraic manipulation, we are able to bound $\frac{p_q}{p^*}$, which leads to bounds on the revenue ratio.
 
% We begin by providing an upper bound on $p_h$ resulting from the optimality condition $p_h \geq c \int_0^{\infty} \bar{F}_V(p) dp$. This will lead to a revenue bound on $p_h \bar{F}_V(p)$.

\begin{align*}
    \int_{0}^{p_q} \bar{F}_V(p) dp  = \int_{0}^{p^*} \bar{F}_V(p) dp + \int_{p^*}^{p_q} \bar{F}_V(p) dp    &
 \geq  p^* \bar{F}_V(p^*) + \int_{p^*}^{p_q} \bar{F}_V(p^*)^{\frac{p_q-p}{p_q-p^*}}\bar{F}_V(p_q)^{\frac{p-p^*}{p_q-p^*}}  dp \\
    % & \geq  p^* \bar{F}_V(p^*) + \bar{F}_V(p^*)^{\frac{p_q}{p_q-p^*}}\bar{F}_V(p_q)^{\frac{-p^*}{p_q-p^*}} \left[ \frac{(p_q-p^*)\left( \frac{\bar{F}_V(p_q)}{\bar{F}_V(p^*)} \right)^{\frac{p}{p_q-p^*}}} {\ln\left( \frac{\bar{F}_V(p_q)}{\bar{F}_V(p^*)} \right)} \right]^{p_q}_{p^*} \\
    & =  p^* \bar{F}_V(p^*) + \frac{(\bar{F}_V(p_q)-\bar{F}_V(p^*)) (p_q-p^*) }{\ln(\bar{F}_V(p_q))-\ln(\bar{F}_V(p^*))}
\end{align*}
% \begin{align}
%     \int_{p_q}^{\infty} \bar{F}_V(p) dp  & \leq \int_{p_q}^{\infty} \bar{F}_V(p^*)^{\frac{p_q-p}{p_q-p^*}}\bar{F}_V(p_q)^{\frac{p-p^*}{p_q-p^*}}  dp \\
%     % & =  \bar{F}_V(p^*)^{\frac{p_q}{p_q-p^*}}\bar{F}_V(p_q)^{\frac{-p^*}{p_q-p^*}} \left[ \frac{(p_q-p^*)\left( \frac{\bar{F}_V(p_q)}{\bar{F}_V(p^*)} \right)^{\frac{p}{p_q-p^*}}} {\ln\left( \frac{\bar{F}_V(p_q)}{\bar{F}_V(p^*)} \right)} \right]^{\infty}_{p_q} \\
%     & = \frac{- \bar{F}_V(p_q) (p_q-p^*) }{\ln(\bar{F}_V(p_q))-\ln(\bar{F}_V(p^*))}
% \end{align}
On the other hand, 
\begin{equation*}
    \int_{p_q}^{\infty} \bar{F}_V(p) dp   \leq \int_{p_q}^{\infty} \bar{F}_V(p^*)^{\frac{p_q-p}{p_q-p^*}}\bar{F}_V(p_q)^{\frac{p-p^*}{p_q-p^*}}  dp  = \frac{- \bar{F}_V(p_q) (p_q-p^*) }{\ln(\bar{F}_V(p_q))-\ln(\bar{F}_V(p^*))}
\end{equation*}
Therefore, from Lemma \ref{lemma_opt_cond_quant} we have that:
\begin{align}
   & \int_{0}^{p_q} \bar{F}_V(p) dp   = \frac{(1-\tau)}{\tau }  \int_{p_q}^{\infty} \bar{F}_V(p)  dp \nonumber \\
  \implies & ~~
p^* \bar{F}_V(p^*) + \frac{(\bar{F}_V(p_q)-\bar{F}_V(p^*)) (p_q-p^*) }{\ln(\bar{F}_V(p_q))-\ln(\bar{F}_V(p^*))}   \leq \frac{(1-\tau)}{\tau } \frac{- \bar{F}_V(p_q) (p_q-p^*) }{\ln(\bar{F}_V(p_q))-\ln(\bar{F}_V(p^*))} \nonumber \\
% \implies & ~~ p^* \bar{F}_V(p^*)  \leq \frac{ (\bar{F}_V(p^*) - \frac{1}{\tau}\bar{F}_V(p_q)) (p_q-p^*) }{\ln(\bar{F}_V(p_q))-\ln(\bar{F}_V(p^*))}\\
% \implies & ~~ p^* \left[ \bar{F}_V(p^*) \left( \ln \left( \frac{\bar{F}_V(p_q)}{\bar{F}_V(p^*)}\right) +1 \right) - \frac{1}{\tau} \bar{F}_V(p_q) \right]  \geq p_q (\bar{F}_V(p^*) -\frac{1}{\tau} \bar{F}_V(p_q))  \label{eq:step_a} \\
\implies &  ~
\frac{\left[ \bar{F}_V(p^*) \left( \ln \left( \frac{\bar{F}_V(p_q)}{\bar{F}_V(p^*)}\right) +1 \right) - \frac{1}{\tau} \bar{F}_V(p_q) \right] } {(\bar{F}_V(p^*) -\frac{1}{\tau} \bar{F}_V(p_q))}   \leq   
\frac{p_q} {p^*} \label{eq:step_b}\\
\implies &  ~
\frac{\bar{F}_V(p_q)\left[ \bar{F}_V(p^*) \left( \ln \left( \frac{\bar{F}_V(p_q)}{\bar{F}_V(p^*)}\right) +1 \right) - \frac{1}{\tau} \bar{F}_V(p_q) \right] } {\bar{F}_V(p^*)(\bar{F}_V(p^*) -\frac{1}{\tau} \bar{F}_V(p_q))}   \leq   
\frac{p_q\bar{F}_V(p_q)} {p^*\bar{F}_V(p^*)} =  \frac{\mathcal{R}(p_q)}{\mathcal{R}(p^*)}  \nonumber
\end{align}
Where (\ref{eq:step_b}) follows from $\ln(\bar{F}_V(p_q))-\ln(\bar{F}_V(p^*))<0$ since $p_q > p^*$,  and requires $\bar{F}_V(p_q)\geq \tau$, proven in Lemma \ref{f_greater_than_tau}, which can be found in Appendix \ref{aux_proofs}. Substituting $z= \frac{\bar{F}_V(p_q)}{\bar{F}_V(p^*)}$,

%\blu Think about if we need assumption $\bar{F}_V(p_q) \neq  \bar{F}_V(p^*)$, also be careful with $<, \leq$. \bla

\begin{align*}
    \frac{\bar{F}_V(p_q)\left[ \bar{F}_V(p^*) \left( \ln \left( \frac{\bar{F}_V(p_q)}{\bar{F}_V(p^*)}\right) +1 \right) - \frac{1}{\tau} \bar{F}_V(p_q) \right] } {\bar{F}_V(p^*)(\bar{F}_V(p^*) -\frac{1}{\tau} \bar{F}_V(p_q))} &= 
    \frac{z \bar{F}_V(p^*)\left[ \bar{F}_V(p^*) \left( \ln \left( z \right) +1 \right) - \frac{1}{\tau} \bar{F}_V(p^*) z \right] } {\bar{F}_V(p^*)(\bar{F}_V(p^*) -\frac{z}{\tau} \bar{F}_V(p^*))} \\
    & =     \frac{z \left[ \tau \left( \ln \left( z \right) +1 \right) - z \right] } {( \tau - z  )} \leq \frac{\mathcal{R}(p_q)}{\mathcal{R}(p^*)} 
\end{align*}
Where the requirement $\tau \leq z \leq 1$ follows from $\bar{F}_V(p^*) \geq \bar{F}_V(p_q)$, since $p_q > p^*$. \Halmos  \endproof

%\blu This requires differentiability, put in assumption \bla Differentiability is implied by log-concavity?
\colllowerbound*

We will use the bound $\ln(x)\leq x-1 $, which follows from a Taylor approximation. Since we have $\tau \leq z$, then it follows that $\frac{z\tau}{\tau-z}\leq 0$. Therefore, making this substitution,

\begin{equation*}
\frac{z  \tau \left( \ln \left( z \right) +1 \right) - z^2  } {( \tau - z  )} \geq \frac{z^2(\tau-1)}{ \tau-z}
\end{equation*}

Furthermore, we can find the minimizer of this lower bound.

\begin{align*}
\diffp{}{{z}}  \frac{z^2(\tau-1)}{ \tau-z} =& \frac{2z(\tau-1)}{\tau-z} + \frac{2z^2(\tau-1)}{(\tau-z)^2}=0 \\
\implies & 2z(\tau-z)+z =0\\
\implies & z = 2c
\end{align*}

Substituting this back into the bound,

\begin{equation*}
\min_z \frac{z^2(\tau-1)}{ \tau-z} = \frac{(2c)^2(\tau-1)}{ \tau-(2c)} =4\tau(1-\tau)
\end{equation*}

\upperbounds*

\proof{Proof of Lemma \ref{lemma_hinge_revenue_case_2}:} 

\textbf{Hinge loss}, $p_h \leq p^*$: First, we will show that $p^*=1$. The gradient of the expected revenue is positive $$  \diffp{\mathcal{R}(p)}{{p}} = \diffp{}{{p}}  p\bar{F}_V(p) = e^{pg(c)}(1+p\ln(g(c))) \geq 0 $$ for $p\ln(g(c))\geq -1$, or equivalently, $g(c)^p \geq e^{-1}$. Since $0<g(c)<1$, $g(c)^p$ is decreasing in $p$, so $g(c)^p \geq e^{-1}$ for all $0 \leq p \leq 1$. It follows that the maximum occurs at $p=1$.

Next we will show that $  \frac{\mathcal{R}(p_h)}{\mathcal{R}(p^*)} = \min_{0 < f < 1} \frac{c(f-1)e^{c(f-1)}}{f \ln(f) }$. Due to the optimality condition of hinge loss function, $$p_h= c \int_{0}^{\infty} \bar{F}_V(p) dp = c\int_{0}^{\infty} e^{p\ln(g(c))} dp   = \frac{c\left(g(c)-1\right)}{\ln(g(c))}.$$

Furthermore, $\bar{F}_V(p_h)=e^{p_h\ln(g(c))}$ and $\bar{F}_V(p^*)=g(c)$. If the derived expressions are substituted back into revenue expression:

$$ \frac{\mathcal{R}(p_h)}{\mathcal{R}(p^*)}= \frac{p_h\bar{F}_V(p_h)}{p^*\bar{F}_V(p^*)}= \frac{c(g(c)-1)e^{c(g(c)-1)}}{g(c) \ln(g(c))}. $$

Since $g(c)=\argmin_{0 < f < 1} \frac{c(f-1)e^{c(f-1)}}{f \ln(f) }$, this proves the result. Furthermore, that $g(c)=1$ for $0< c <0.5$ is proven in Corollary \ref{minimum_at_c}.

\textbf{Hinge loss}, $p_h \geq p^*$:
First, we will show that $p^*=t(c)$ and therefore that $\bar{F}_V(p^*)=1$ for the range of parameter values we consider. To achieve this, we will show that the revenue for the region $p\geq t(c)$ is less than $t(c)$.  The gradient of the expected revenue for $\bar{F}_V(p)=e^{t(c)-p} $ at $t(c)$ is $$  \diffp{\mathcal{R}(p)}{{p}}\Big\rvert_{p=t(c)}  = \diffp{}{{p}}  p\bar{F}_V(p)\big\rvert_{p=t(c)}  = e^{t(c)-p}(1-p)\big\rvert_{p=t(c)} =1-t(c). $$

It can be shown that $t(c) \geq 1$ for $0.5 \leq c <1$. To observe this, note that $t(0.5)= 1$ and that the gradient of $t(c)$ is positive in the range $0.5 \leq c <1$:

$$ \diffp{t(c)}{c} =\diffp{}{c}   \dfrac{1}{2}\left(\sqrt{\dfrac{c-5}{c-1}} - 1 \right) = \dfrac{1}{2}\left(\dfrac{1}{c-1}- \dfrac{c-5}{(c-1)^2}\right)\sqrt{\dfrac{c-1}{c-5}} = \dfrac{1}{2}\dfrac{4}{(c-1)^2}\sqrt{\dfrac{c-1}{c-5}}>0$$

It follows that the gradient of $\mathcal{R}(p)$ evaluated at $t(c)$ is less than or equal to zero. Since $pe^{t(c)-p} $ is unimodal, the maximum revenue is achieved at $p=t(c)$.

%As proven in Lemma \ref{lemma_hinge_revenue_case_2}, $t(c)=\frac{z^*}{c}-1$ where $z^*= \argmin_{z\leq -2c}  \dfrac{cz  e^{-z(\frac{1}{c}-1)-1}}{z + c}$. It follows that $t(c) \leq 1$ so the gradient at $t(c)$ is positive. Since $pe^{t(c)-p} $ is uni-modal, the maximum revenue is achieved at $p=t(c)$.

 Due to the optimality condition of hinge loss function, $$p_h= c \int_{0}^{\infty} \bar{F}_V(p) dp = c\int_{0}^{t(c)}1dp+ c\int_{t(c)}^{\infty} e^{t(c)-p} dp   = c(t(c)+1)$$ 

Therefore, $\bar{F}_V(p_h)=e^{t(c)(1-c)-c} $. Finally, the expected revenue ratio is:
$$\frac{\mathcal{R}(p_h)}{\mathcal{R}(p^*)}= \frac{p_h\bar{F}_V(p_h)}{p^*\bar{F}_V(p^*)}= \dfrac{c(t(c)+1)e^{t(c)(1-c)-c}}{t(c)}= \dfrac{cz^*  e^{-z^*(\frac{1}{c}-1)-1}}{z^* + c}$$

Where we have made the substitution that  $t(c)=\frac{-z^*}{c}-1$ in the last equality, and the result follows from the fact that $z^*= \argmin_{z\leq -2c}  \dfrac{cz  e^{-z(\frac{1}{c}-1)-1}}{z + c}$.

% \begin{lemma}
% \label{minimum_at_c}
% $  \min_{0 < f < 1}  \frac{c(f-1)e^{c(f-1)}}{f \ln(f) }= c$ for $0 \leq c \leq 0.5$
% \end{lemma}

\textbf{Quantile loss}, $p_q \leq p^*$: It is clear that the expected revenue bound of pricing with the quantile loss function with this valuation distribution is $1-\tau$. Furthermore, from Lemma \ref{minimum_at_c}, $\min_{0 < f < 1}  \frac{c(f-1)e^{c(f-1)}}{f \ln(f) }= c$ for $0 \leq c \leq 0.5$. Using $1-\tau=c$ completes the result.

\textbf{Quantile loss}, $p_h \geq p^*$: For this valuation distribution, we will show that $p^*=1$ and therefore that $\bar{F}_V(p^*)=1$. The gradient of $p\bar{F}_V(p_h) = e^{(p-1)(1-\frac{z(\tau)}{\tau})} $ at $p=1$ is
$$  \diffp{\mathcal{R}(p)}{{p}}\Big\rvert_{p=1}  = \diffp{}{{p}}  p\bar{F}_V(p)\big\rvert_{p=1}  = e^{(p-1)(1-\frac{z(\tau)}{\tau})}\left(1+p(1-\frac{z(\tau)}{\tau})\right)\Big\rvert_{p=1}  = 2-\frac{z(\tau)}{\tau} $$

Due to the unimodality of $pe^{(p-1)(1-\frac{z(\tau)}{\tau})}$, the maximum will occur at $p \leq 1$ if the gradient is negative, which occurs if $z(\tau) \geq 2\tau$. It can be verified that $z(\tau) \geq 2\tau$ for $0 < \tau \leq 0.5$ (we used numerical simulation). It follows that $p^*=1$.

 Due to the optimality condition of the quantile loss function, we have that
 \begin{align*}
 \tau \left(\int_{0}^{\infty} \bar{F}_V(p) dp \right) = \int_{p_q}^{\infty}\bar{F}_V(p) dp  
 \implies &\tau \left(\int_{0}^{1} 1 dp + \int_{1}^{\infty} e^{(p-1)(1-\frac{z(\tau)}{\tau})} dp\right) = \int_{p_q}^{\infty} e^{(p-1)(1-\frac{z(\tau)}{\tau})} dp \\
  \implies & \tau \left(\ 1 + \left[ \dfrac{e^{(p-1)(1-\frac{z(\tau)}{\tau})}}{1-\frac{z(\tau)}{\tau}} \right]_1^{\infty} \right) = \left[ \dfrac{e^{(p-1)(1-\frac{z(\tau)}{\tau})}}{1-\frac{z(\tau)}{\tau}} \right]_{p_q}^{\infty}  \\
\implies & \tau \left(1- \dfrac{1}{1-\frac{z(\tau)}{\tau}} \right) =  \dfrac{e^{(p_q-1)(1-\frac{z(\tau)}{\tau})}}{1-\frac{z(\tau)}{\tau}}  \\
\implies & z(\tau) =  e^{(p_q-1)(1-\frac{z(\tau)}{\tau})} \\
\implies & p_q =  \dfrac{\ln(z(\tau))}{1-\frac{z(\tau)}{\tau}} +1
 \end{align*}

Furthermore, by substitution of $p_q$ into $\bar{F}_V(p)$, it follows that $\bar{F}_V(p_q)= z(\tau)$. Finally, the expected revenue is:
 $$\frac{\mathcal{R}(p_h)}{\mathcal{R}(p^*)}= \frac{p_h\bar{F}_V(p_h)}{p^*\bar{F}_V(p^*)}= z(\tau)\left(\dfrac{\ln(z(\tau))}{1-\frac{z(\tau)}{\tau}} +1\right) = \dfrac{ z(\tau) \tau(\ln(z(\tau))+1)-z(\tau)^2} {\tau-z(\tau)}$$

Since $z(\tau)=\argmin_{\tau \leq z \leq 1} \dfrac{z\tau(\ln(z)+1)-z^2} {\tau-z}$, the result follows. \Halmos

\endproof

\section{Auxiliary proofs}
\label{aux_proofs}
\begin{lemma}
\label{monotone_lemma}
In the domain $0 \leq x \leq 1, 0 \leq b \leq 1 $ and $xb^{x-1} \leq 1$, $xb^{x-1}$ is monotone increasing. 
\end{lemma}

\proof{Proof of Lemma \ref{monotone_lemma}:} 
\begin{equation*}
\diffp{}{x} xb^{x-1} = \diffp{}{x} \frac{x}{b}e^{x \ln(b)} =  \frac{e^{x \ln(b)}}{b}+  \frac{x \ln(b)}{b} e^{x \ln(b)} = \frac{e^{x \ln(b)}}{b}\left[ 1+ x \ln(b) \right]
\end{equation*}

 $\frac{e^{x \ln(b)}}{b} \geq 0 $ for $b \geq 0$. Therefore the derivative is positive if $1+x \ln(b) > 0$, or equivalently, $x < \frac{-1}{\ln(b)}$, and negative for $x > \frac{-1}{\ln(b)}$, with the maximum occurring at $x=\frac{-1}{\ln(b)}$.  Therefore, the maximum value $xb^{x-1}$ can take is $\frac{-e^{-1}}{b\ln(b)}$. 
 
 It can be verified that $\frac{-e^{-1}}{b\ln(b)} \geq 1$; for example, by taking the derivative, the maximum occurs when $b=e^{-1}$. Finally, $xb^{x-1}=1$ at $x=1$. Therefore, $xb^{x-1}$ is monotone increasing in the range $0 \leq x < \frac{-1}{\ln(b)}$, and for $ \frac{-1}{\ln(b)} \leq x \leq 1$, $xb^{x-1} \geq 1$.  \Halmos

\endproof

\begin{lemma}
\label{f_greater_than_tau}
If $p^*\leq p_q$, then $\bar{F}_V(p_q)\geq \tau$.
\end{lemma}

\proof{Proof of Lemma \ref{f_greater_than_tau}:} 

Due to log-concavity: 
\begin{equation*}
\bar{F}_V(p) \geq \bar{F}_V(p_q)^{\frac{p}{p_q}} ~~ \forall p \leq p_q ~~~~ \text{and} ~~~ \bar{F}_V(p_q) \geq \bar{F}_V(p)^{\frac{p_q}{p}} ~ ~ \forall p \geq p_q
\end{equation*}

Therefore,
\begin{equation*}
    \int_{0}^{p_q} \bar{F}_V(p) dp \geq \int_{0}^{p_q} \bar{F}_V(p_q)^{\frac{p}{p_q}} = \left[ \frac{p_q \bar{F}_V(p_q)^{\frac{p}{p_q}}}{\ln(\bar{F}_V(p_q))} \right]_0^{p_q} = \frac{p_q (\bar{F}_V(p_q)-1)}{\ln(\bar{F}_V(p_q))}
\end{equation*}

\begin{equation*}
    \int_{p_q}^{\infty} \bar{F}_V(p) dp \leq \int_{0}^{p_q} \bar{F}_V(p_q)^{\frac{p}{p_q}} = \left[ \frac{p_q \bar{F}_V(p_q)^{\frac{p}{p_q}}}{\ln(\bar{F}_V(p_q))} \right]_{p_q}^{\infty} = \frac{-p_q \bar{F}_V(p_q)}{\ln(\bar{F}_V(p_q))}
\end{equation*}

From optimality condition in Lemma \ref{lemma_opt_cond_hinge}:
\begin{align*}
\int_{0}^{p_q} \bar{F}_V(p) dp ~  = ~ \frac{1-\tau}{\tau} \int_{p_q}^{\infty} \bar{F}_V(p) dp & \implies ~~
\frac{p_q (\bar{F}_V(p_q)-1)}{\ln(\bar{F}_V(p_q))}~   \leq  ~ - \frac{1-\tau}{\tau} \frac{p_q \bar{F}_V(p_q)}{\ln(\bar{F}_V(p_q))} \\
&\implies ~~ \frac{p_q (\frac{1}{\tau}\bar{F}_V(p_q)-1)}{\ln(\bar{F}_V(p_q))}  \leq 0
\end{align*}

Therefore $\bar{F}_V(p_q)\geq \tau$, since $\ln(\bar{F}_V(p_q)) < 0$.  \Halmos

\endproof

\begin{lemma}
\label{revenue_quantile_case_1}
Consider a pricing policy $p_q$ that satisfies the condition from Lemma \ref{lemma_opt_cond_quant} and for which $p_q < p^*$, then:
$$ \frac{\mathcal{R}(p_q)}{\mathcal{R}(p^*)} \geq \tau $$
\end{lemma}

\proof{Proof of Lemma \ref{revenue_quantile_case_1}:}

By rearranging Lemma \ref{lemma_opt_cond_quant}, we have that $(1-\tau)\int_{0}^{\infty} \bar{F}_V(p) dp = \int_{0}^{p_q} \bar{F}_V(p) dp $. By bounding each of these terms, we will find relationships between $p_q,p^*,\bar{F}_V(p_q)$ and $\bar{F}_V(p^*)$ that will allow us to bound the revenue.

From log concavity, it follows that $\bar{F}_V(p_q) \geq \bar{F}_V(p)^{\frac{p_q-p}{p^*-p}} \bar{F}_V(p^*)^{\frac{p^*-p_q}{p^*-p}}$ for $0 \leq p \leq p_q \leq p^*$. Rearranging, we can upper bound the complementary CDF according to $ \bar{F}_V(p) \leq \bar{F}_V(p_q)^{\frac{p^*-p}{p_q-p}} \bar{F}_V(p^*)^{\frac{p^*-p_q}{p_q-p}}$ for $0 \leq p \leq p_q \leq p^*$. However, we can refine this upper bound because $\bar{F}_V(p) \leq 1$. The point at which these bounds intersect is $p=p^*-\frac{(p^*-p_q)\ln(\bar{F}_V(p^*))}{\ln(\bar{F}_V(p_q))}$. Therefore,

\begin{equation*}
\bar{F}_V(p) \leq     \begin{dcases}
      1   &  \text{for} ~~ 0 \leq p \leq p^*-\frac{(p^*-p_q)\ln(\bar{F}_V(p^*))}{\ln(\bar{F}_V(p_q))}  \\
      \bar{F}_V(p_q)^{\frac{p^*-p}{p_q-p}} \bar{F}_V(p^*)^{\frac{p^*-p_q}{p_q-p}}   & \text{for} ~~  p^*-\frac{(p^*-p_q)\ln(\bar{F}_V(p^*))}{\ln(\bar{F}_V(p_q))}  \leq p \leq p_q \\
    \end{dcases} 
\end{equation*}

It follows that: 

\begin{align*}
\label{eq:otherway_quant}
 \int_{0}^{p_q} \bar{F}_V(p) dp &\leq   \int_{0}^{ p^*-\frac{(p^*-p_q)\ln(\bar{F}_V(p^*))}{\ln(\bar{F}_V(p_q))}} 1 dp +  \int_{p^*-\frac{(p^*-p_q)\ln(\bar{F}_V(p^*))}{\ln(\bar{F}_V(p_q))}}^{p_q} \bar{F}_V(p_q)^{\frac{p^*-p}{p_q-p}} \bar{F}_V(p^*)^{\frac{p^*-p_q}{p_q-p}}   dp \\
 &=  p^*-\frac{(p^*-p_q)\ln(\bar{F}_V(p^*))}{\ln(\bar{F}_V(p_q))} + \left[  \frac{(p_q-p)^2\bar{F}_V(p_q)^{\frac{p^*-p}{p_q-p}} \bar{F}_V(p^*)^{\frac{p^*-p_q}{p_q-p}}}{(p^*-p_q)(\ln(\bar{F}_V(p_q))-\ln(\bar{F}_V(p^*)))} \right]_{p^*-\frac{(p^*-p_q)\ln(\bar{F}_V(p^*))}{\ln(\bar{F}_V(p_q))} }^{p_q} \\
 & =  p^*-\frac{(p^*-p_q)\ln(\bar{F}_V(p^*))}{\ln(\bar{F}_V(p_q))} + \frac{(p^*-p_q)(\ln(\bar{F}_V(p^*))-\ln(\bar{F}_V(p_q)))}{\ln(\bar{F}_V(p_q))^2} \\
\end{align*}

Where the second equality follows from $\lim_{p \rightarrow p_q} \bar{F}_V(p_q)^{\frac{p^*-p}{p_q-p}} \bar{F}_V(p^*)^{\frac{p^*-p_q}{p_q-p}}  \rightarrow 0$. For a lower bound on the complementary CDF, we have $\bar{F}_V(p) \geq \bar{F}_V(p_q)^{\frac{p}{p_q}}$ for $0 \leq p \leq p_q$ and $\bar{F}_V(p) \geq \bar{F}_V(p_q)^{\frac{p-p_q}{p^*-p_q}} \bar{F}_V(p^*)^{\frac{p^*-p}{p^*-p_q}}$ for $p_q \leq p \leq p^*$. This follows from log-concavity and since $\bar{F}_V(0)=1$. It follows that:

\begin{align*}
\int_{0}^{\infty} \bar{F}_V(p) dp  &\geq 
\int_{0}^{p_q} \bar{F}_V(p) dp  + \int_{p_q}^{p^*} \bar{F}_V(p) dp \\
& \geq \int_{0}^{p_q} \bar{F}_V(p_q)^{\frac{p}{p_q}} dp +  \int_{p_q}^{p^*}  \bar{F}_V(p_q)^{\frac{p-p_q}{p^*-p_q}} \bar{F}_V(p^*)^{\frac{p^*-p}{p^*-p_q}} dp \\
&= \left[ \frac{ p_q \bar{F}_V(p_q)^{\frac{p}{p_q}} }{\ln(\bar{F}_V(p_q))} \right]_0^{p_q}+ \left[ \frac{(\bar{F}_V(p_q)^{\frac{p-p_q}{p^*-p_q}} \bar{F}_V(p^*)^{\frac{p^*-p}{p^*-p_q}} )(p^*-p_q)}{\ln(\bar{F}_V(p^*))-\ln(\bar{F}_V(p_q))}\right]_{p_q}^{p^*} \\
&= \frac{ p_q\left(\bar{F}_V(p_q)-1\right)}{\ln(\bar{F}_V(p_q))} + \frac{(\bar{F}_V(p^*)-\bar{F}_V(p_q))(p^*-p_q)}{\ln(\bar{F}_V(p^*))-\ln(\bar{F}_V(p_q))}
\end{align*}

% \begin{equation}
% \label{eq:otherway_quant}
%  \int_{0}^{p_q} \bar{F}_V(p) dp =  \frac{ p_q \left(\bar{F}_V(p_q)-1\right)}{\ln(\bar{F}_V(p_q))} 
% \end{equation}

Where here the second inequality follows due to log-concavity,  $\bar{F}_V(p) \geq \bar{F}_V(0)^{\theta} \bar{F}_V(p^*)^{1-\theta}$ for $ 0< \theta<1$,  and since $\bar{F}_V(0)=1$. Furthermore, by substituting (\ref{eq:up_int_id}) into condition $p_h \geq c \int_0^{\infty} \bar{F}_V(p) dp$:

\begin{equation}
\frac{p_h}{p^*} \geq  \frac{c}{p^*}
 \int_{0}^{\infty} \bar{F}_V(p) dp \geq
\frac{c \left(\bar{F}_V(p^*)-1\right)}{\ln(\bar{F}_V(p^*))}  \label{price_ratio}
\end{equation}

Therefore:
\begin{equation}
 \frac{\mathcal{R}(p_h)}{\mathcal{R}(p^*)} = 
\frac{p_h\bar{F}_V(p_h)} {p^*\bar{F}_V(p^*)} \geq \frac{p_h\bar{F}_V(p^*)^{\frac{p_h}{p^*}} } {p^*\bar{F}_V(p^*)}  \Halmos\label{rev_monotone_func}
\end{equation}
\endproof

       % appendix for experimental results
\end{APPENDICES}

%%%%%%%%%%%%%%%%%
\end{document}